\documentclass{article} 
\usepackage{iclr2027_conference,times}

\usepackage{amsmath,amsfonts,bm}

\def\eqref#1{equation~\ref{#1}}
\def\Eqref#1{Equation~\ref{#1}}

\def\1{\bm{1}}

\DeclareMathAlphabet{\mathsfit}{\encodingdefault}{\sfdefault}{m}{sl}
\SetMathAlphabet{\mathsfit}{bold}{\encodingdefault}{\sfdefault}{bx}{n}

\usepackage{hyperref}
\usepackage{url}

\usepackage{booktabs}
\usepackage{multirow}
\usepackage{graphicx}
\usepackage{amsmath}
\usepackage{array}
\usepackage[T1]{fontenc}
\usepackage{wrapfig}
\usepackage{needspace}

\title{Your Temporal Link Predictor Is Blind to Who Is Active: A Missing Factor That Transfers Across Models}

\author{\normalfont
\begin{tabular}[t]{@{}l@{\hspace{1.5em}}l@{\hspace{1.5em}}l@{}}
\textbf{Ji Zhang}$^{*\dagger}$ & \textbf{Zixin Liu}$^{*}$ & \textbf{Yiran Ding}$^{*}$ \\
Harvard University & Wuhan University & University of California, Berkeley \\
{\small\texttt{jizhang@g.harvard.edu}} & {\small\texttt{liuzixin@whu.edu.cn}} & {\small\texttt{yiran\_ding@berkeley.edu}} \\[0.7em]
\textbf{Jiayi Wang} & \textbf{Yilu Du} & \textbf{Weijia Xuan} \\
Harvard University & Wuhan University & Tsinghua University
\end{tabular}
}

\iclrfinalcopy 
\begin{document}

\maketitle
\lhead{Preprint.} 
\begingroup
\renewcommand{\thefootnote}{}
\footnotetext{$^{*}$Equal contribution. \quad $^{\dagger}$Corresponding author.}
\endgroup

\begin{abstract}
An interaction has two parts: someone decides to act, and then chooses whom to act on. Temporal link prediction has concentrated on the second, and we show that it is blind to the first by construction: a standard negative keeps the real source and swaps the destination, and we prove that this cancels the source's activity exactly from the optimal score, so no model trained and evaluated this way is ever rewarded for learning it. Under the harder historical and inductive negatives, whose sources differ, the same factor becomes the dominant signal. We model it with \textbf{Source Node Activity Modeling (SNAM)}, a self-exciting event intensity fitted by an exact point-process likelihood to decayed interaction counts the history states already contain; it has \textbf{fewer than 20 parameters}. On their own, never looking at the destination, \textbf{these parameters beat DyGFormer and TPNet} on four of five datasets under historical negatives. Added to the frozen scores of TPNet, TGN, DyGFormer and DSRD, four models of different design, \textbf{without retraining anything}, they raise AP on almost every backbone--dataset pair in both settings, by up to 25 points. Our full model ranks first overall against eleven baselines on 13 datasets and three protocols, and on million-event streams trains an epoch 9--100$\times$ faster than TPNet and DyGFormer. We conclude that source activity is a blind spot of temporal link prediction, and a cheap, transferable one to close. Code is available at {\def\UrlBreaks{\do\/\do-}\url{https://github.com/Erutaner/Your-Temporal-Link-Predictor-Is-Blind-to-Who-Is-Active}}.
\end{abstract}

\section{Introduction}
\begin{wrapfigure}[15]{r}{0.48\textwidth}
    \vspace{-\intextsep}
    \centering
    \includegraphics[width=\linewidth]{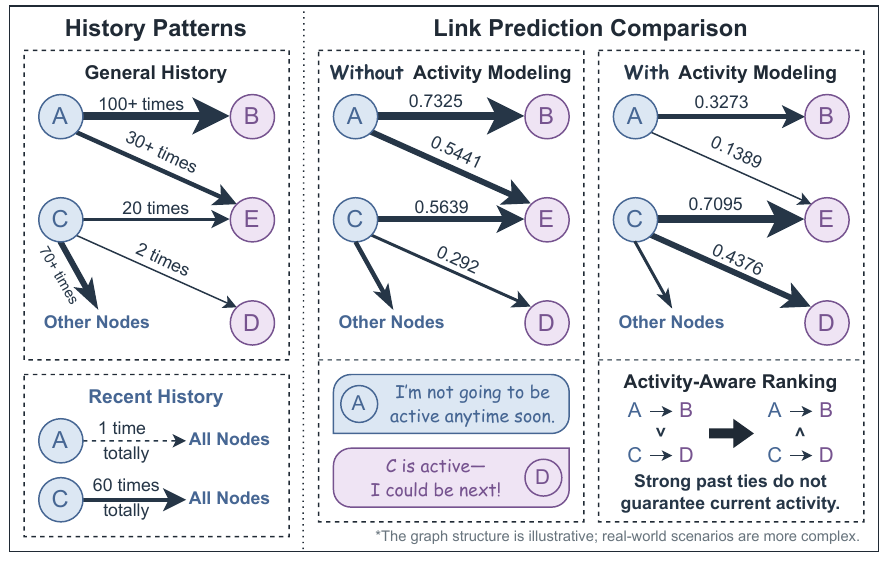}
    \caption{Illustration of source activity in temporal link prediction.}
    \label{fig:scenario}
\end{wrapfigure}
Predicting whether an interaction is about to happen requires two judgements: whether the two parties are compatible, and whether the initiator is active right now. Figure~\ref{fig:scenario} gives a payment network as an example. Account A transferred funds to B many times in the past but has been inactive for an extended period; C has rarely transacted with D but has recently initiated frequent transfers. A model that relies mainly on the strength of historical relationships ranks A$\rightarrow$B first and underestimates C$\rightarrow$D. Comparing candidate transactions across source accounts therefore requires assessing both pairwise compatibility and the likelihood that the source will initiate a transaction at the current time.

This is the setting of temporal link prediction. Existing studies learn interaction patterns through node memory, temporal neighborhood aggregation, and historical sequence encoding \citep{rossi2020temporal,xu2020inductive,yu2023dyglib,liu2025salom,xiong2025tetgn}, and enhance relation modeling with temporal structure and candidate-specific information \citep{lu2024tpnet,yi2025craft,tieu2025lstep}. These designs characterize which destinations are compatible with a source; a link discrimination objective never asks whether that source is about to act. This rate of acting is the conditional intensity of point processes \citep{hawkes1971spectra,zhou2025advances}, here of the interactions a node initiates; a few temporal graph models use it as a training signal \citep{trivedi2019dyrep,wen2022trend}, but source activity has never entered the link score as a separate factor, for a reason: under a link objective alone it cancels between a positive and its random negative, so nothing rewards learning it; only negatives with other sources expose its weight. Benchmark studies agree: simple activity heuristics rival learned models \citep{daniluk2023global,cornell2025heuristics}. We therefore formulate temporal link prediction as two factors, the source's likelihood of acting now and the candidate's compatibility, learned separately and combined.


For the first factor we introduce Source Node Activity Modeling (SNAM). Using time-decayed interaction counts at $K$ fixed timescales, SNAM estimates the source's conditional intensity with only $K+1$ parameters shared by all nodes. Its point-process likelihood fits this intensity to both the times at which a node acts and the intervals in which it does not, the latter never seen by a link objective. Because SNAM depends only on historical interaction counts and its own parameters, its score can be added to other link prediction models: with the same time unit and timescales on the same dataset, transfer requires only selecting the activity coefficient $\beta$ on the backbone's validation split, without retraining either model.


The second factor completes the model. Multi-Timescale Historical Representation (MTHR) maintains every node's decayed states and interaction counts at the same $K$ timescales, computed once per block of events, and feeds both factors. Candidate Relation Modeling (CRM) scores the candidate's compatibility from node-pair historical features and a candidate-conditioned encoding of the source's recent interactions. The final score adds $\beta$ times SNAM's log intensity to the relation score; this is where source activity enters comparisons across sources.

Historical and inductive negative sampling allow comparisons between candidate interactions involving different source nodes, aligning with the cross-source prediction needs of the motivating scenario; the related theoretical analysis is provided in Appendix~\ref{app:activity_motivation}. We evaluate our model on 13 datasets. The full model achieves the best overall average rank across three negative sampling strategies, with its main advantages under historical and inductive negative sampling. Under historical negative sampling, source activity scoring alone outperforms DyGFormer and TPNet on four of five datasets, with further improvements when combined with relation scoring. Cross-model transfer improves AP across all 20 backbone--dataset combinations, with gains of up to 25 percentage points.


We make three contributions. (1) \textbf{Diagnosis.} A standard negative keeps the source and swaps the destination; we prove that the source's activity cancels exactly from the optimal score, so the link objective never rewards it, which calls for learning it separately. (2) \textbf{Method.} SNAM does so with a point-process likelihood and $K+1$ parameters. Because it depends only on interaction counts, its log intensity can be added to other backbones' scores with $\beta$ chosen on validation and nothing retrained; the full model builds both factors on shared block-wise multi-timescale states. (3) \textbf{Results.} On 13 datasets against eleven baselines, SNAM alone beats DyGFormer and TPNet on four of five datasets under historical negatives and lifts four other backbones without retraining; the full model ranks first overall and, on long streams, is faster than TPNet and DyGFormer in training and inference.

\section{Related Work}
\textbf{Historical Representation and Relation Modeling.} Existing methods encode temporal histories through walks, memory updates, and neighborhood aggregation \citep{wang2021causal,kumar2019predicting,rossi2020temporal,xu2020inductive,liu2025salom,xiong2025tetgn}, using sequence encoders, simplified networks, or state space models \citep{yu2023dyglib,cong2023graphmixer,li2025dygmamba,tieu2025lstep}. Others capture structural relationships through common neighbors and temporal paths \citep{wang2021causal,luo2022nat,souza2022provably,lu2024tpnet,liu2025salom}, or temporal patterns through time encoding and frequency analysis \citep{chung2025timeencoder,tian2024freedyg,tieu2025lstep}, building on broader work on time representations \citep{kazemi2019time2vec,xu2019functional}. DyGFormer and CRAFT capture source--candidate correlations through neighbor co-occurrence and cross-attention, respectively; CRAFT also uses destination recency in scoring \citep{yu2023dyglib,yi2025craft}. We combine reusable historical aggregates with candidate-conditioned recent interaction encoding.


\textbf{Interaction Rates and Source Node Activity.} Temporal point processes model event times conditioned on history \citep{zhou2025advances}, using self-excitation, recurrent networks, and attention \citep{hawkes1971spectra,mei2017neural,zuo2020transformer}. Alternative formulations parameterize cumulative intensity or inter-event time distributions \citep{omi2019fully,shchur2020intensity}. Graph methods define such intensities on node pairs from node representations and neighborhoods \citep{trivedi2017knowevolve,zuo2018embedding,trivedi2019dyrep,wen2022trend}; a node's own rate is folded into the pair function. TREND \citep{wen2022trend} comes closest, adding a node-level term, but only as an auxiliary loss for the representation, not part of the link score. We model the source's rate on its own, with $K+1$ parameters over decayed interaction counts, so it enters the link score as a separate factor and transfers to other backbones on the same stream without retraining.

\textbf{Evaluation Protocols and Activity Information.} Temporal graph benchmarks show that data splits, negative sampling, and evaluation procedures affect model comparisons \citep{poursafaei2022towards,huang2023tgb,gastinger2024tgb2,yi2025tgbseq,dizaji2026tgrab}; standardized evaluation is also emphasized in point-process benchmarking \citep{xue2024easytpp}. Studies of global temporal dynamics, recency, and popularity further show that useful predictive information can be extracted without complex relation encoders \citep{daniluk2023global,cornell2025heuristics}. We examine the complementarity between source intensity and relation scoring under different negative sampling protocols, distinguishing cross-source discrimination from candidate ranking for a fixed source.

\section{Preliminaries}
\textbf{Definition 1 (Temporal Graph).} A temporal graph is represented as a sequence of interactions $\mathcal{E}=[e_1,e_2,\ldots]$ ordered by non-decreasing timestamps, where $e_i=(u_i,v_i,t_i,\phi_i)$ denotes an interaction record. Here, $u_i,v_i\in\mathcal{N}$ are the source and destination nodes, $t_i$ is the timestamp satisfying $t_i\leq t_{i+1}$, and $\phi_i\in\mathbb{R}^{F}$ is an optional interaction feature vector. Repeated interactions between the same pair of nodes are allowed.

\textbf{Definition 2 (Interaction History).} Given a temporal graph $\mathcal{E}$ and a query time $t$, the interaction history consists of all interaction records strictly preceding $t$:
\[
\mathcal{H}_{t^-}=\{e_i\in\mathcal{E}:t_i<t\}.
\]

\textbf{Definition 3 (Temporal Link Prediction).} Given the interaction history $\mathcal{H}_{t^-}$ and a node pair $(u,v)$, temporal link prediction aims to predict whether an interaction between $u$ and $v$ occurs at the query time $t$.

\textbf{Definition 4 (Time Block).} Given $m$ interaction records ordered by non-decreasing timestamps, we partition them into $C\leq m$ contiguous, non-overlapping blocks $\{\mathcal{E}_c\}_{c=0}^{C-1}$. Under an equal partition, the number of interactions in block $c$ is
\[
B_c=q+\mathbf{1}(c<r),\qquad q=\left\lfloor\frac{m}{C}\right\rfloor,\quad r=m-Cq,
\]
where $\mathbf{1}(\cdot)$ denotes the indicator function. The start time of block $c$ is defined as $a_c=\min_{e_i\in\mathcal{E}_c}t_i$.
\section{Methodology}
Our model comprises three main modules: Multi-Timescale Historical Representation (MTHR), Candidate Relation Modeling (CRM), and Source Node Activity Modeling (SNAM). MTHR aggregates interaction histories across multiple timescales. CRM integrates these representations with candidate-specific historical information to quantify node-pair compatibility, while SNAM estimates the conditional intensity of source-initiated interactions from historical activity statistics. The two branches model relational compatibility and source node activity, respectively, and their outputs are integrated into the final prediction score.
\subsection{Multi-Timescale Historical Representation}
We aggregate historical interactions across multiple timescales and combine block-start caching with query-time updates to construct reusable node embeddings and provide historical statistics for relation and activity modeling.
\paragraph{Block-Start Historical Aggregation.}
Given $K$ timescales with decay rates $\omega_1\geq\cdots\geq\omega_K>0$, we define the exponential kernel $d_k(t,s)=\exp[-\omega_k(t-s)]$ for $t\geq s$. Let $a_c$ denote the start of block $c$ and $p_u(e)$ the other endpoint of an interaction $e$ incident to node $u$. We construct the time-decayed adjacency matrix from interactions preceding the block start:
\begin{equation}
A_c^k[u,w]
=
\sum_{\substack{e\ni u,\ p_u(e)=w\\t_e<a_c}}
d_k(a_c,t_e).
\label{eq:mthr_adjacency}
\end{equation}
Each interaction contributes to both endpoints' neighborhoods, and repeated interactions are accumulated with their respective decay weights. Given a base node embedding table $E\in\mathbb R^{|\mathcal N|\times d}$, we compute
\begin{equation}
\begin{aligned}
X_{1,c}^k&=A_c^kE,\qquad n_c^k=A_c^k\mathbf1,\\
m_{c,u}&=\sum_{k=1}^{K}g_{1,k}P_1
\frac{X_{1,c,u}^k}{\sqrt{1+n_c^k[u]}},\\
M_c[u,:]&=m_{c,u}^{\top},\qquad X_{2,c}^k=A_c^kM_c.
\end{aligned}
\label{eq:mthr_block_aggregation}
\end{equation}
Here, $P_1$ and $g_{1,k}$ are a learnable linear transformation and timescale coefficient, respectively, and $\mathbf1$ is the all-ones vector. The vector $X_{1,c,u}^k$ is a weighted sum of the embeddings of node $u$'s historical interaction partners, and $n_c^k[u]$ is the corresponding time-decayed interaction count. Normalization and cross-timescale fusion yield the intermediate representation $m_{c,u}$. The second layer further aggregates neighbors' historical representations to obtain $X_{2,c,u}^k$, incorporating neighborhood information within two hops. These states are cached at the block start and reused across queries within the block.

\paragraph{Query-Time Historical Updates.}
For interaction content $\psi_u(e)$ held fixed during aggregation and updates, let $H_{u,\psi}^k(t)=\sum_{e\ni u,t_e<t}d_k(t,t_e)\psi_u(e)$ denote its historical aggregate. The exponential kernel enables an exact update using the block-start state and newly observed interactions:
\begin{equation}
H_{u,\psi}^k(t)
=
d_k(t,a_c)H_{u,\psi}^k(a_c)
+
\sum_{\substack{e\ni u\\a_c\leq t_e<t}}
d_k(t,t_e)\psi_u(e).
\label{eq:mthr_history_update}
\end{equation}
Substituting $\psi_u(e)=E_{p_u(e)}$, $m_{c,p_u(e)}$, $1$, and $\bar\phi_e$ yields the query-time first-layer aggregate $\widehat x_{1,u}^k(t;c)$, second-layer aggregate $\widehat x_{2,u}^k(t;c)$, activity count $n_u^k(t)$, and edge-feature aggregate $x_{f,u}^k(t)$, respectively. Here, $\bar\phi_e$ denotes optional edge features standardized using training-set statistics. This update avoids repeatedly processing pre-block interactions; the intermediate neighbor representations used in the second layer remain fixed at the block start.

\paragraph{Reusable Node Representations.}
We divide the first-layer state by $\sqrt{1+n_u^k(t)}$ and the second-layer and edge-feature states by $1+n_u^k(t)$, obtaining normalized representations $\widetilde x_{1,u}^k$, $\widetilde x_{2,u}^k$, and $\widetilde x_{f,u}^k$, respectively. We then fuse information across timescales:
\begin{equation}
\begin{aligned}
r_u(t)={}&
\sum_{k=1}^{K}
\left(g_{1,k}P_1\widetilde x_{1,u}^k
+g_{2,k}P_2\widetilde x_{2,u}^k
+g_{f,k}W_f\widetilde x_{f,u}^k\right)
+\frac{b_{\mathrm{cold}}}{1+n_u^K(t)},\\
h_u(t)={}&\tanh\!\left(Wr_u(t)+b_W\right).
\end{aligned}
\label{eq:mthr_node_representation}
\end{equation}
Here, $P_2,W\in\mathbb R^{d\times d}$ and $W_f\in\mathbb R^{d\times F}$ are learnable transformations, $g_{2,k},g_{f,k}$ are timescale coefficients, and $b_W,b_{\mathrm{cold}}\in\mathbb R^d$ are learnable vectors. The parameters $P_1,g_{1,k}$ are shared with~\eqref{eq:mthr_block_aggregation}. The vector $b_{\mathrm{cold}}$ provides a shared baseline for nodes with sparse interaction histories, with its contribution decreasing as the activity count $n_u^K(t)$ at the slowest timescale increases.

\subsection{Candidate Relation Modeling}

Building on the node embeddings, we combine node-pair historical features with candidate-conditioned encoding of recent interactions to capture associations between source nodes and candidate targets. All computations use only the history preceding the time $t$ of the candidate interaction $(u,v,t)$.

\paragraph{Node-Pair Historical Features.}
We extract three types of features: direct interactions, historical representation matching, and structural associations. First, we define the time-decayed node-pair count $n_{uv}^k(t)=\sum_{e:\{u_e,v_e\}=\{u,v\},\,t_e<t}d_k(t,t_e)$. We use $\log(1+n_{uv}^k(t))$ together with the endpoint counts $\log(1+n_u^k(t))$ and $\log(1+n_v^k(t))$ to assess historical associations in the context of both nodes' activity levels. This allows the model to account for a candidate target's historical activity when assessing its likelihood of receiving an interaction at the query time.

We next apply low-rank bilinear matching to the normalized neighborhood representations:
\begin{equation}
q_{ij}^k=
\left\langle
L_{ij}^k\widetilde x_{i,u}^k,\,
R_{ij}^k\widetilde x_{j,v}^k
\right\rangle,
\qquad (i,j)\in\{(1,1),(1,2),(2,1)\}.
\label{eq:crm_matching}
\end{equation}
Here, $L_{ij}^k,R_{ij}^k\in\mathbb R^{r\times d}$ are learnable projections, and $r$ is the matching dimension. We additionally use $q_{\mathrm{static}}=\langle E_u,E_v\rangle/\sqrt d$ to measure base embedding compatibility and $q_{ff}^k=\langle C_f^k\widetilde x_{f,u}^k,C_f^k\widetilde x_{f,v}^k\rangle$ to measure historical edge-feature compatibility, where $C_f^k\in\mathbb R^{r\times F}$ is a shared learnable projection. We apply $\operatorname{slog}(z)=\operatorname{sign}(z)\log(1+|z|)$ to $q_{ij}^k$ and $q_{ff}^k$ to compress their numerical ranges.

To incorporate structural information, we extract structural associations using fixed random projections. Let $R\in\mathbb R^{|\mathcal N|\times D}$ be the projection matrix and $B_t^k$ the time-decayed adjacency matrix constructed from all interactions preceding $t$. We define the block-start structural representation $S_{1,c}^k=A_c^kR$ and compute
\begin{equation}
\begin{aligned}
s_1^k(t)&=B_t^kR,\qquad s_2^k(t;c)=B_t^kS_{1,c}^k,\\
\kappa_{ij}^k(u,v,t)
&=\chi\!\left(\langle s_{i,u}^k,s_{j,v}^k\rangle\right),
\qquad (i,j)\in\{1,2\}^2,
\end{aligned}
\label{eq:crm_structural_features}
\end{equation}
where $\chi(z)=0.1\log(1+\max(z,0))$. We concatenate the counts, matching scores, and structural association features across timescales into $\eta_{uv}(t)$, which serves as the node-pair input to the relation scoring network.

\paragraph{Candidate-Conditioned Recent Interaction Encoding.}
To capture associations between individual historical interactions and the current candidate, we select the most recent interactions of source node $u$ preceding $t$, up to a maximum of $M$ records, denoted by $\mathcal R_u(t)$. For each interaction $e$, let $p=p_u(e)$. We construct an encoding from its temporal weights $a_e^k(t)=d_k(t,t_e)$, structural associations $b_{e,v}^k=\chi(\langle S_{1,c,p}^k,S_{1,c,v}^k\rangle)$, representation matching score $j_{e,v}=\langle W_{\mathrm{sim}}m_{c,p},m_{c,v}\rangle/\sqrt d$, and standardized edge features:
\begin{equation}
\begin{aligned}
\zeta_{e,v}(t)
&=\left[(a_e^k(t))_{k=1}^{K};(b_{e,v}^k)_{k=1}^{K};j_{e,v};\bar\phi_e\right],\\
c_{uv}(t)
&=\sum_{e\in\mathcal R_u(t)}\rho\!\left(\zeta_{e,v}(t)\right).
\end{aligned}
\label{eq:crm_recent_encoding}
\end{equation}
Here, $W_{\mathrm{sim}}\in\mathbb R^{d\times d}$ is a learnable transformation, and $\rho$ is a shared multilayer perceptron. By incorporating candidate information before aggregation, this encoding allows the same history to produce different representations for different candidates. Finally, $c_{uv}(t)$, $\eta_{uv}(t)$, and $h_u(t),h_v(t)$ are jointly provided to the relation scoring network.

\subsection{Source Node Activity Modeling}

Let $\lambda_u(t)$ denote the conditional intensity of interactions initiated by source node $u$ at time $t$, representing its instantaneous event rate given the interaction history. We reuse the previously defined time-decayed interaction counts and define
\begin{equation}
\lambda_u(t)=\mu+\sum_{k=1}^{K}w_k n_u^k(t),\qquad
w_k=\omega_k\exp(\theta_k),\qquad
\mu=\omega_K\exp(\theta_\mu).
\label{eq:snam_intensity}
\end{equation}
Here, $\mu$ is the baseline event rate, $w_k$ controls the contribution of historical activity at the $k$-th timescale to the current intensity, $\theta_\mu,\theta_1,\ldots,\theta_K$ are learnable parameters, and $\omega_k$ is a fixed decay rate.

The module contains only $K+1$ learnable parameters shared across all source nodes, and the computation of $\lambda_u(t)$ does not depend on any other trainable parameters. This design yields a small parameter footprint and enables cross-model parameter transfer under consistent historical statistics and timescale configurations.

\subsection{Score Integration and Optimization}
We model candidate relation compatibility and source node activity separately and combine their outputs at the scoring stage. During training, we compute a separate loss for each branch and optimize their sum.

\paragraph{Scoring Function.}
We feed the node embeddings $h_u(t),h_v(t)$, node-pair historical features $\eta_{uv}(t)$, and recent interaction encoding $c_{uv}(t)$ into a relation scoring network $F_\Theta$ parameterized by $\Theta$, yielding two real-valued outputs:
\begin{equation}
\begin{aligned}
z(u,v,t)&=F_\Theta\!\left(h_u(t),h_v(t),\eta_{uv}(t),c_{uv}(t)\right)\in\mathbb R^2,\\
g(u,v,t)&=\log\!\left(\exp(z_0)+\exp(z_1)\right),\qquad
f(u,v,t)=\max(z_0,z_1).
\end{aligned}
\label{eq:score_relation}
\end{equation}
Here, $z_0,z_1$ are two scoring channels. We use the smooth aggregate $g$ during training and the base score $f$ during evaluation. The network combines nonlinear and linear output paths, with its full architecture provided in the Appendix~\ref{app:relation_scoring_architecture}. We then incorporate the source activity estimated by SNAM to obtain the final score:
\begin{equation}
s_\beta(u,v,t)=f(u,v,t)+\beta\log\lambda_u(t),
\label{eq:score_integration}
\end{equation}
where $\beta\geq0$ controls the contribution of activity information. This combination accounts for the source node's current interaction rate alongside relational compatibility.

\paragraph{Loss Functions and Optimization.}
Let training block $c$ contain $B_c$ positive interactions $\mathcal E_c^+=\{(u_i,v_i,t_i)\}_{i=1}^{B_c}$. For each positive interaction, we sample a negative target $v_i^-$, yielding $\mathcal E_c^-=\{(u_i,v_i^-,t_i)\}_{i=1}^{B_c}$. We train the relation branch using the binary logistic loss:
\begin{equation}
\mathcal L_{\mathrm{link},c}
=-\frac{1}{B_c}\sum_{i=1}^{B_c}
\left[
\log\sigma\!\left(g(u_i,v_i,t_i)\right)
+\log\sigma\!\left(-g(u_i,v_i^-,t_i)\right)
\right],
\label{eq:loss_link}
\end{equation}
where $\sigma(x)=1/(1+\exp(-x))$. Because $v_i^-$ inherits the source and timestamp of its positive, the optimal discriminator under this loss depends only on the relative preference among destinations; the source's activity cancels exactly and nothing rewards learning it (Appendix~\ref{app:activity_motivation}, ~\eqref{eq:app-a6}). We therefore introduce a point-process negative log-likelihood for the activity branch:
\begin{equation}
\mathcal L_{\mathrm{act},c}
=\frac{1}{B_c}
\left[
\sum_{u\in\mathcal U_{\mathrm{train}}}
\int_{a_c}^{b_c}\lambda_u(\tau)\,\mathrm d\tau
-\sum_{i=1}^{B_c}\log\lambda_{u_i}(t_i^-)
\right].
\label{eq:loss_activity}
\end{equation}
Here, $\mathcal U_{\mathrm{train}}$ is the set of training source nodes, $[a_c,b_c)$ is the observation window associated with block $c$, and $b_c$ is the start time of the next block. The notation $t_i^-$ indicates that the intensity uses only the history preceding the current interaction. The event term encourages higher intensities at observed interaction times, while the integral term penalizes excessive cumulative intensity over the observation window. The exponential kernel admits a closed-form integral (\eqref{eq:app-a24} in Appendix~\ref{app:activity_likelihood}), eliminating the need for negative sampling or temporal discretization in this objective.

The overall training objective is
\begin{equation}
\mathcal L_c(\Theta,\vartheta)
=\mathcal L_{\mathrm{link},c}(\Theta)
+\mathcal L_{\mathrm{act},c}(\vartheta),
\qquad
\vartheta=(\theta_\mu,\theta_1,\ldots,\theta_K).
\label{eq:loss_total}
\end{equation}
Here, $\Theta$ collectively denotes the parameters of the relation branch and its historical encoder, while $\vartheta$ contains the SNAM parameters. The two losses act on disjoint parameter sets.

\paragraph{SNAM: Beyond an Integrated Component.}
SNAM learns source activity rates directly from interaction histories without relying on the internal representations of a specific relation encoder. Its parameters and training objective are independent of the other modules. This decoupled design enables SNAM to serve as a plug-in activity modeling component for existing link prediction models: replacing $f$ in ~\eqref{eq:score_integration} with the base model's scoring function incorporates source activity information without retraining the base model. Furthermore, under consistent historical statistics, time units, and decay-scale configurations, trained SNAM parameters can be directly reused across models without model-specific retraining.

SNAM can also be trained independently and use $\log\lambda_u(t)$ as its sole prediction score, allowing us to assess the discriminative power of source activity alone. We experimentally evaluate its cross-model reuse, effectiveness across datasets, and standalone predictive performance.

\begin{figure*}[t]
\vspace{-\intextsep}
\centering
\begin{minipage}[t]{0.31\textwidth}
\vspace{0pt}
\centering
\includegraphics[width=\linewidth]{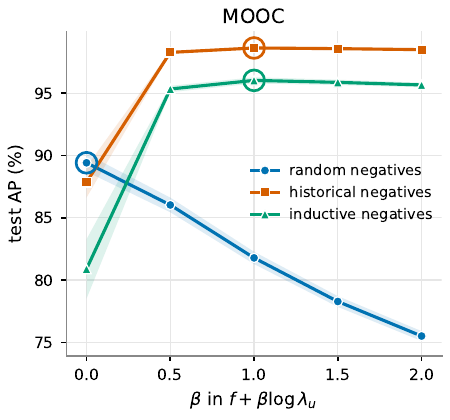}
\caption{$\beta$ sensitivity under different negative sampling protocols on MOOC.}
\label{fig:beta_mooc}
\end{minipage}\hfill
\begin{minipage}[t]{0.665\textwidth}
\vspace{0pt}
\centering
\includegraphics[width=\linewidth]{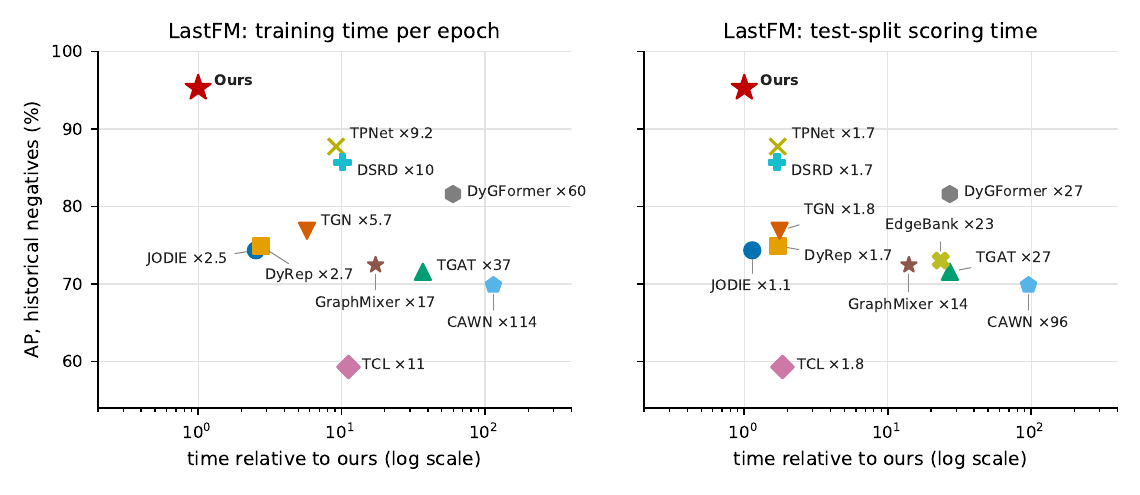}
\caption{Model accuracy and runtime (log scale). Left: time per training epoch; right: full test-set scoring time.}
\label{fig:speed_lastfm}
\end{minipage}
\end{figure*}

\section{Experiments}

\subsection{Experimental Settings}
\textbf{Datasets and Baselines.} We evaluate on 13 temporal link prediction benchmarks: Wikipedia, Reddit, MOOC, LastFM, Enron, Social Evo., UCI, Flights, Can. Parl., US Legis., UN Trade, UN Vote, and Contact \citep{poursafaei2022towards}. We compare against 11 methods: JODIE \citep{kumar2019predicting}, TGN \citep{rossi2020temporal}, CAWN \citep{wang2021causal}, EdgeBank \citep{poursafaei2022towards}, GraphMixer \citep{cong2023graphmixer}, NAT \citep{luo2022nat}, PINT \citep{souza2022provably}, DyGFormer \citep{yu2023dyglib}, TPNet \citep{lu2024tpnet}, GRN \citep{chang2026grn}, and DSRD \citep{chang2026dsrd}. Dataset statistics and baseline sources are in Tables~\ref{tab:datasets} and~\ref{tab:baseline_sources}.

\textbf{Task Settings.} Following DyGLib, we split each dataset chronologically into approximately 70\%/15\%/15\% for training/validation/testing and evaluate transductive and inductive link prediction, the latter involving nodes unseen during training. Training uses random negatives; evaluation uses random, historical, and inductive negatives with AP and AUC-ROC metrics. Detailed protocols are in Appendices~\ref{app:data_splits_visibility} and~\ref{app:negative_sampling_protocols}.

\textbf{Model Configuration.} Node embedding, low-rank matching, and random projection dimensions are 128, 16, and 256, respectively. We retain at most 20 recent interactions and 16 timescales, whose range is determined by the training interaction span and smallest positive inter-event interval. Dataset-specific timescales, block partitions, and other configurations are in Tables~\ref{tab:datasets} and~\ref{tab:training_settings}.

\textbf{Implementation and Optimization.} We use AdamW with a base learning rate of $10^{-3}$, and $10^{-2}$ and $5\times10^{-2}$ for additional output projections and SNAM parameters, respectively. Training runs for at most 500 epochs with early stopping; checkpoints and the activity coefficient $\beta$ are selected by validation performance. Each run uses one NVIDIA RTX 5090 GPU. Main experiments report means and standard deviations over five seeds; other run counts are specified in table captions.

\subsection{Overall Effectiveness and Efficiency}

\begin{table}[t]
\vspace{-\intextsep}
\caption{Transductive link prediction, AP (\%), mean{\scriptsize$\pm$std} over seeds (five; DyGFormer$^\dagger$ three, LastFM one) under random (rnd), historical (hist) and inductive (ind) negative sampling. \textbf{Bold}: best; \underline{underline}: second best. $\dagger$: retrained by us with the corrected released code (released configuration).}
\label{tab:trans_ap}
\centering
\resizebox{\textwidth}{!}{%
\renewcommand{\arraystretch}{0.85}\setlength{\tabcolsep}{3pt}%
\begin{tabular}{llccccccccccccc}
\toprule
NSS & Dataset & JODIE & TGN & CAWN & EdgeBank & GraphMixer & NAT & PINT & DyGFormer & TPNet & GRN$^\dagger$ & DSRD$^\dagger$ & Ours & Ours+$\lambda$ \\
\midrule
\multirow{14}{*}{\rotatebox{90}{rnd}} & Wikipedia & 96.50{\scriptsize$\pm$0.14} & 98.45{\scriptsize$\pm$0.06} & 98.76{\scriptsize$\pm$0.03} & 90.37{\scriptsize$\pm$0.00} & 97.25{\scriptsize$\pm$0.03} & 98.03{\scriptsize$\pm$0.07} & 98.45{\scriptsize$\pm$0.04} & \underline{99.05{\scriptsize$\pm$0.01}}$^\dagger$ & \textbf{99.32{\scriptsize$\pm$0.03}} & 94.50{\scriptsize$\pm$0.23}$^\dagger$ & 97.30{\scriptsize$\pm$0.02}$^\dagger$ & 98.95{\scriptsize$\pm$0.02} & 98.95{\scriptsize$\pm$0.02} \\
 & Reddit & 98.31{\scriptsize$\pm$0.14} & 98.63{\scriptsize$\pm$0.06} & 99.11{\scriptsize$\pm$0.01} & 94.86{\scriptsize$\pm$0.00} & 97.31{\scriptsize$\pm$0.01} & 99.13{\scriptsize$\pm$0.10} & 99.15{\scriptsize$\pm$0.02} & 99.23{\scriptsize$\pm$0.00}$^\dagger$ & \textbf{99.27{\scriptsize$\pm$0.00}} & 93.01{\scriptsize$\pm$0.21}$^\dagger$ & \underline{99.26{\scriptsize$\pm$0.01}}$^\dagger$ & 99.15{\scriptsize$\pm$0.00} & 99.15{\scriptsize$\pm$0.00} \\
 & MOOC & 80.23{\scriptsize$\pm$2.44} & 89.15{\scriptsize$\pm$1.60} & 80.15{\scriptsize$\pm$0.25} & 57.97{\scriptsize$\pm$0.00} & 82.78{\scriptsize$\pm$0.15} & 85.88{\scriptsize$\pm$0.55} & 88.08{\scriptsize$\pm$0.86} & 87.52{\scriptsize$\pm$0.49} & \textbf{96.39{\scriptsize$\pm$0.09}} & 75.37{\scriptsize$\pm$0.66}$^\dagger$ & 83.01{\scriptsize$\pm$0.08}$^\dagger$ & \underline{89.42{\scriptsize$\pm$0.71}} & \underline{89.42{\scriptsize$\pm$0.71}} \\
 & LastFM & 70.85{\scriptsize$\pm$2.13} & 77.07{\scriptsize$\pm$3.97} & 86.99{\scriptsize$\pm$0.06} & 79.29{\scriptsize$\pm$0.00} & 75.61{\scriptsize$\pm$0.24} & 88.02{\scriptsize$\pm$1.94} & 89.66{\scriptsize$\pm$1.81} & \underline{92.99}$^\dagger$ & \textbf{94.50{\scriptsize$\pm$0.08}} & 75.89{\scriptsize$\pm$0.33}$^\dagger$ & 92.30{\scriptsize$\pm$0.28}$^\dagger$ & 92.15{\scriptsize$\pm$1.13} & 92.15{\scriptsize$\pm$1.13} \\
 & Enron & 84.77{\scriptsize$\pm$0.30} & 86.53{\scriptsize$\pm$1.11} & 89.56{\scriptsize$\pm$0.09} & 83.53{\scriptsize$\pm$0.00} & 82.25{\scriptsize$\pm$0.16} & 90.60{\scriptsize$\pm$0.66} & 92.20{\scriptsize$\pm$0.15} & 92.45{\scriptsize$\pm$0.10}$^\dagger$ & \underline{92.90{\scriptsize$\pm$0.17}} & 78.77{\scriptsize$\pm$0.59}$^\dagger$ & 91.13{\scriptsize$\pm$0.12}$^\dagger$ & \textbf{93.15{\scriptsize$\pm$0.18}} & \textbf{93.15{\scriptsize$\pm$0.18}} \\
 & Social Evo. & 89.89{\scriptsize$\pm$0.55} & 93.57{\scriptsize$\pm$0.17} & 84.96{\scriptsize$\pm$0.09} & 74.95{\scriptsize$\pm$0.00} & 93.37{\scriptsize$\pm$0.07} & 88.92{\scriptsize$\pm$3.45} & 94.42{\scriptsize$\pm$0.03} & \textbf{94.73{\scriptsize$\pm$0.01}} & \textbf{94.73{\scriptsize$\pm$0.02}} & 82.46{\scriptsize$\pm$0.87}$^\dagger$ & 92.51{\scriptsize$\pm$0.03}$^\dagger$ & \underline{94.50{\scriptsize$\pm$0.03}} & \underline{94.50{\scriptsize$\pm$0.03}} \\
 & UCI & 89.43{\scriptsize$\pm$1.09} & 92.34{\scriptsize$\pm$1.04} & 95.18{\scriptsize$\pm$0.06} & 76.20{\scriptsize$\pm$0.00} & 93.25{\scriptsize$\pm$0.57} & 93.40{\scriptsize$\pm$0.26} & \underline{96.45{\scriptsize$\pm$0.11}} & 95.72{\scriptsize$\pm$0.17}$^\dagger$ & \textbf{97.35{\scriptsize$\pm$0.04}} & 87.33{\scriptsize$\pm$0.43}$^\dagger$ & 93.96{\scriptsize$\pm$0.70}$^\dagger$ & 95.84{\scriptsize$\pm$0.33} & 95.84{\scriptsize$\pm$0.33} \\
 & Flights & 95.60{\scriptsize$\pm$1.73} & 97.95{\scriptsize$\pm$0.14} & 98.51{\scriptsize$\pm$0.01} & 89.35{\scriptsize$\pm$0.00} & 90.99{\scriptsize$\pm$0.05} & 98.57{\scriptsize$\pm$0.12} & 98.80{\scriptsize$\pm$0.02} & \underline{98.91{\scriptsize$\pm$0.01}} & \textbf{98.93{\scriptsize$\pm$0.02}} & 84.24{\scriptsize$\pm$2.37}$^\dagger$ & 90.77{\scriptsize$\pm$0.02}$^\dagger$ & 98.89{\scriptsize$\pm$0.01} & 98.89{\scriptsize$\pm$0.01} \\
 & Can. Parl. & 69.26{\scriptsize$\pm$0.31} & 70.88{\scriptsize$\pm$2.34} & 69.82{\scriptsize$\pm$2.34} & 64.55{\scriptsize$\pm$0.00} & 77.04{\scriptsize$\pm$0.46} & \underline{79.72{\scriptsize$\pm$1.76}} & 68.36{\scriptsize$\pm$1.43} & 61.99{\scriptsize$\pm$1.66}$^\dagger$ & \textbf{80.06{\scriptsize$\pm$0.28}}$^\dagger$ & 68.93{\scriptsize$\pm$1.56}$^\dagger$ & 71.43{\scriptsize$\pm$0.11}$^\dagger$ & 76.95{\scriptsize$\pm$2.54} & 76.95{\scriptsize$\pm$2.58} \\
 & US Legis. & 75.05{\scriptsize$\pm$1.52} & \underline{75.99{\scriptsize$\pm$0.58}} & 70.58{\scriptsize$\pm$0.48} & 58.39{\scriptsize$\pm$0.00} & 70.74{\scriptsize$\pm$1.02} & \textbf{78.71{\scriptsize$\pm$0.87}} & 74.85{\scriptsize$\pm$0.97} & 71.11{\scriptsize$\pm$0.59} & 68.73{\scriptsize$\pm$0.80}$^\dagger$ & 66.24{\scriptsize$\pm$1.20}$^\dagger$ & 65.54{\scriptsize$\pm$2.14}$^\dagger$ & 68.58{\scriptsize$\pm$0.79} & 68.48{\scriptsize$\pm$0.90} \\
 & UN Trade & 64.94{\scriptsize$\pm$0.31} & 65.03{\scriptsize$\pm$1.37} & 65.39{\scriptsize$\pm$0.12} & 60.41{\scriptsize$\pm$0.00} & 62.61{\scriptsize$\pm$0.27} & \textbf{73.95{\scriptsize$\pm$1.16}} & \underline{70.20{\scriptsize$\pm$0.58}} & 66.46{\scriptsize$\pm$1.29} & 67.16{\scriptsize$\pm$0.25}$^\dagger$ & 56.49{\scriptsize$\pm$0.78}$^\dagger$ & 59.24{\scriptsize$\pm$0.15}$^\dagger$ & 68.40{\scriptsize$\pm$0.13} & 68.40{\scriptsize$\pm$0.13} \\
 & UN Vote & 63.91{\scriptsize$\pm$0.81} & 65.72{\scriptsize$\pm$2.17} & 52.84{\scriptsize$\pm$0.10} & 58.49{\scriptsize$\pm$0.00} & 52.11{\scriptsize$\pm$0.16} & \textbf{70.45{\scriptsize$\pm$0.68}} & \underline{66.25{\scriptsize$\pm$0.78}} & 55.55{\scriptsize$\pm$0.42} & 58.08{\scriptsize$\pm$0.06}$^\dagger$ & 53.94{\scriptsize$\pm$0.80}$^\dagger$ & 51.33{\scriptsize$\pm$0.84}$^\dagger$ & 59.55{\scriptsize$\pm$0.16} & 59.55{\scriptsize$\pm$0.16} \\
 & Contact & 95.31{\scriptsize$\pm$1.33} & 96.89{\scriptsize$\pm$0.56} & 90.26{\scriptsize$\pm$0.28} & 92.58{\scriptsize$\pm$0.00} & 91.92{\scriptsize$\pm$0.03} & 97.39{\scriptsize$\pm$0.22} & \underline{98.64{\scriptsize$\pm$0.02}} & 98.29{\scriptsize$\pm$0.01} & \textbf{98.66{\scriptsize$\pm$0.01}} & 81.53{\scriptsize$\pm$0.92}$^\dagger$ & 91.76{\scriptsize$\pm$0.08}$^\dagger$ & 98.17{\scriptsize$\pm$0.02} & 98.17{\scriptsize$\pm$0.02} \\
\cmidrule{2-15}
 & Avg. Rank & 8.92 & 6.88 & 8.38 & 11.38 & 9.31 & 5.46 & 4.58 & 4.73 & \textbf{2.58} & 11.92 & 8.31 & \underline{4.27} & \underline{4.27} \\
\midrule
\multirow{14}{*}{\rotatebox{90}{hist}} & Wikipedia & 83.01{\scriptsize$\pm$0.66} & 86.86{\scriptsize$\pm$0.33} & 71.21{\scriptsize$\pm$1.67} & 73.35{\scriptsize$\pm$0.00} & 90.90{\scriptsize$\pm$0.10} & 68.36{\scriptsize$\pm$4.47} & \underline{91.33{\scriptsize$\pm$0.76}} & 76.01{\scriptsize$\pm$0.51}$^\dagger$ & 81.55{\scriptsize$\pm$4.10} & 91.03{\scriptsize$\pm$0.21}$^\dagger$ & 90.71{\scriptsize$\pm$0.19}$^\dagger$ & 88.90{\scriptsize$\pm$0.81} & \textbf{93.28{\scriptsize$\pm$0.08}} \\
 & Reddit & 80.03{\scriptsize$\pm$0.36} & 81.22{\scriptsize$\pm$0.61} & 80.82{\scriptsize$\pm$0.45} & 73.59{\scriptsize$\pm$0.00} & 78.44{\scriptsize$\pm$0.18} & 80.58{\scriptsize$\pm$0.91} & 83.93{\scriptsize$\pm$1.11} & 81.32{\scriptsize$\pm$1.35}$^\dagger$ & 81.02{\scriptsize$\pm$1.31} & 73.76{\scriptsize$\pm$0.41}$^\dagger$ & \underline{88.00{\scriptsize$\pm$0.49}}$^\dagger$ & 81.72{\scriptsize$\pm$1.76} & \textbf{90.01{\scriptsize$\pm$0.25}} \\
 & MOOC & 78.94{\scriptsize$\pm$1.25} & 87.06{\scriptsize$\pm$1.93} & 74.05{\scriptsize$\pm$0.95} & 60.71{\scriptsize$\pm$0.00} & 77.77{\scriptsize$\pm$0.92} & 85.42{\scriptsize$\pm$3.26} & 84.80{\scriptsize$\pm$1.57} & 85.85{\scriptsize$\pm$0.66} & \underline{92.69{\scriptsize$\pm$0.95}} & 73.28{\scriptsize$\pm$1.06}$^\dagger$ & 78.85{\scriptsize$\pm$1.49}$^\dagger$ & 87.86{\scriptsize$\pm$1.24} & \textbf{98.64{\scriptsize$\pm$0.12}} \\
 & LastFM & 74.35{\scriptsize$\pm$3.81} & 76.87{\scriptsize$\pm$4.64} & 69.86{\scriptsize$\pm$0.43} & 73.03{\scriptsize$\pm$0.00} & 72.47{\scriptsize$\pm$0.49} & 78.75{\scriptsize$\pm$1.63} & 84.95{\scriptsize$\pm$0.43} & 81.61$^\dagger$ & \underline{87.74{\scriptsize$\pm$0.50}} & 73.12{\scriptsize$\pm$0.23}$^\dagger$ & 85.71{\scriptsize$\pm$1.34}$^\dagger$ & 84.64{\scriptsize$\pm$0.94} & \textbf{95.30{\scriptsize$\pm$0.09}} \\
 & Enron & 69.85{\scriptsize$\pm$2.70} & 73.91{\scriptsize$\pm$1.76} & 64.73{\scriptsize$\pm$0.36} & 76.53{\scriptsize$\pm$0.00} & 77.98{\scriptsize$\pm$0.92} & 72.90{\scriptsize$\pm$1.38} & 85.41{\scriptsize$\pm$0.84} & 76.60{\scriptsize$\pm$0.54}$^\dagger$ & 80.79{\scriptsize$\pm$1.68} & 73.40{\scriptsize$\pm$0.66}$^\dagger$ & \underline{87.07{\scriptsize$\pm$0.42}}$^\dagger$ & 84.09{\scriptsize$\pm$1.95} & \textbf{90.42{\scriptsize$\pm$0.64}} \\
 & Social Evo. & 87.44{\scriptsize$\pm$6.78} & 94.45{\scriptsize$\pm$0.56} & 85.53{\scriptsize$\pm$0.38} & 80.57{\scriptsize$\pm$0.00} & 94.93{\scriptsize$\pm$0.31} & 91.29{\scriptsize$\pm$4.89} & 96.73{\scriptsize$\pm$0.29} & \underline{97.38{\scriptsize$\pm$0.14}} & 96.80{\scriptsize$\pm$0.41} & 82.68{\scriptsize$\pm$1.23}$^\dagger$ & 93.63{\scriptsize$\pm$0.55}$^\dagger$ & 97.02{\scriptsize$\pm$0.30} & \textbf{99.22{\scriptsize$\pm$0.02}} \\
 & UCI & 75.24{\scriptsize$\pm$5.80} & 80.43{\scriptsize$\pm$2.12} & 65.30{\scriptsize$\pm$0.43} & 65.50{\scriptsize$\pm$0.00} & 84.11{\scriptsize$\pm$1.35} & 75.38{\scriptsize$\pm$1.28} & \underline{93.96{\scriptsize$\pm$0.20}} & 81.59{\scriptsize$\pm$0.61}$^\dagger$ & 86.34{\scriptsize$\pm$0.80} & 81.43{\scriptsize$\pm$0.77}$^\dagger$ & 86.81{\scriptsize$\pm$0.84}$^\dagger$ & 91.16{\scriptsize$\pm$0.55} & \textbf{96.00{\scriptsize$\pm$0.15}} \\
 & Flights & 66.48{\scriptsize$\pm$2.59} & 66.70{\scriptsize$\pm$1.64} & 64.72{\scriptsize$\pm$0.97} & 70.53{\scriptsize$\pm$0.00} & 71.47{\scriptsize$\pm$0.26} & 64.74{\scriptsize$\pm$1.83} & 66.82{\scriptsize$\pm$1.44} & 66.59{\scriptsize$\pm$0.49} & 69.10{\scriptsize$\pm$1.27} & 61.46{\scriptsize$\pm$3.72}$^\dagger$ & \underline{71.73{\scriptsize$\pm$0.06}}$^\dagger$ & 70.70{\scriptsize$\pm$1.03} & \textbf{83.00{\scriptsize$\pm$0.33}} \\
 & Can. Parl. & 51.79{\scriptsize$\pm$0.63} & 68.42{\scriptsize$\pm$3.07} & 66.53{\scriptsize$\pm$2.77} & 63.84{\scriptsize$\pm$0.00} & 74.34{\scriptsize$\pm$0.87} & \textbf{77.72{\scriptsize$\pm$1.78}} & 63.84{\scriptsize$\pm$5.36} & 62.37{\scriptsize$\pm$1.21}$^\dagger$ & 72.07{\scriptsize$\pm$2.70}$^\dagger$ & 68.61{\scriptsize$\pm$3.39}$^\dagger$ & 60.00{\scriptsize$\pm$2.15}$^\dagger$ & \underline{76.85{\scriptsize$\pm$4.10}} & \underline{76.85{\scriptsize$\pm$4.10}} \\
 & US Legis. & 51.71{\scriptsize$\pm$5.76} & 74.00{\scriptsize$\pm$7.57} & 68.82{\scriptsize$\pm$8.23} & 63.22{\scriptsize$\pm$0.00} & 81.65{\scriptsize$\pm$1.02} & \textbf{91.12{\scriptsize$\pm$1.97}} & 73.58{\scriptsize$\pm$4.16} & \underline{85.30{\scriptsize$\pm$3.88}} & 75.37{\scriptsize$\pm$1.77}$^\dagger$ & 67.84{\scriptsize$\pm$2.52}$^\dagger$ & 55.87{\scriptsize$\pm$5.61}$^\dagger$ & 71.67{\scriptsize$\pm$9.07} & 75.01{\scriptsize$\pm$5.11} \\
 & UN Trade & 61.39{\scriptsize$\pm$1.83} & 58.44{\scriptsize$\pm$5.51} & 55.71{\scriptsize$\pm$0.38} & \underline{81.32{\scriptsize$\pm$0.00}} & 57.05{\scriptsize$\pm$1.22} & 78.65{\scriptsize$\pm$1.16} & 70.07{\scriptsize$\pm$2.28} & 64.41{\scriptsize$\pm$1.40} & 61.97{\scriptsize$\pm$0.59}$^\dagger$ & 55.98{\scriptsize$\pm$1.08}$^\dagger$ & 55.82{\scriptsize$\pm$1.87}$^\dagger$ & 71.55{\scriptsize$\pm$4.36} & \textbf{91.39{\scriptsize$\pm$0.93}} \\
 & UN Vote & 70.02{\scriptsize$\pm$0.81} & 69.37{\scriptsize$\pm$3.93} & 51.26{\scriptsize$\pm$0.04} & \textbf{84.89{\scriptsize$\pm$0.00}} & 51.20{\scriptsize$\pm$1.60} & 71.39{\scriptsize$\pm$2.68} & 71.79{\scriptsize$\pm$2.53} & 60.84{\scriptsize$\pm$1.58} & 64.78{\scriptsize$\pm$1.22}$^\dagger$ & 60.44{\scriptsize$\pm$2.95}$^\dagger$ & 51.69{\scriptsize$\pm$1.57}$^\dagger$ & 70.28{\scriptsize$\pm$3.08} & \underline{75.66{\scriptsize$\pm$1.03}} \\
 & Contact & 95.31{\scriptsize$\pm$2.13} & 93.05{\scriptsize$\pm$2.35} & 84.16{\scriptsize$\pm$0.49} & 88.81{\scriptsize$\pm$0.00} & 93.36{\scriptsize$\pm$0.41} & 96.84{\scriptsize$\pm$0.57} & 97.61{\scriptsize$\pm$0.18} & 97.57{\scriptsize$\pm$0.06} & \underline{98.02{\scriptsize$\pm$0.15}} & 74.77{\scriptsize$\pm$1.65}$^\dagger$ & 92.94{\scriptsize$\pm$0.44}$^\dagger$ & 96.54{\scriptsize$\pm$0.39} & \textbf{98.79{\scriptsize$\pm$0.03}} \\
\cmidrule{2-15}
 & Avg. Rank & 9.62 & 7.38 & 11.31 & 9.27 & 7.38 & 6.92 & 4.65 & 6.38 & 5.08 & 10.00 & 7.08 & \underline{4.42} & \textbf{1.50} \\
\midrule
\multirow{14}{*}{\rotatebox{90}{ind}} & Wikipedia & 75.65{\scriptsize$\pm$0.79} & 85.62{\scriptsize$\pm$0.44} & 74.06{\scriptsize$\pm$2.62} & 80.63{\scriptsize$\pm$0.00} & \underline{88.59{\scriptsize$\pm$0.17}} & 53.85{\scriptsize$\pm$6.25} & 82.39{\scriptsize$\pm$2.70} & 69.34{\scriptsize$\pm$0.32}$^\dagger$ & 79.35{\scriptsize$\pm$5.52} & 87.38{\scriptsize$\pm$0.57}$^\dagger$ & 87.91{\scriptsize$\pm$0.39}$^\dagger$ & 86.00{\scriptsize$\pm$1.46} & \textbf{91.16{\scriptsize$\pm$0.16}} \\
 & Reddit & 86.98{\scriptsize$\pm$0.16} & 88.10{\scriptsize$\pm$0.24} & \underline{91.67{\scriptsize$\pm$0.24}} & 85.48{\scriptsize$\pm$0.00} & 85.26{\scriptsize$\pm$0.11} & 81.71{\scriptsize$\pm$1.03} & 87.59{\scriptsize$\pm$0.65} & 91.51{\scriptsize$\pm$0.51}$^\dagger$ & 88.19{\scriptsize$\pm$0.33} & 79.63{\scriptsize$\pm$0.43}$^\dagger$ & \textbf{93.21{\scriptsize$\pm$0.16}}$^\dagger$ & 90.32{\scriptsize$\pm$0.43} & 91.60{\scriptsize$\pm$0.26} \\
 & MOOC & 65.23{\scriptsize$\pm$2.19} & 77.50{\scriptsize$\pm$2.91} & 73.51{\scriptsize$\pm$0.94} & 49.43{\scriptsize$\pm$0.00} & 74.27{\scriptsize$\pm$0.92} & 77.28{\scriptsize$\pm$1.94} & 75.70{\scriptsize$\pm$1.43} & 81.24{\scriptsize$\pm$0.69} & \underline{88.18{\scriptsize$\pm$0.97}} & 63.96{\scriptsize$\pm$0.31}$^\dagger$ & 75.06{\scriptsize$\pm$0.85}$^\dagger$ & 80.89{\scriptsize$\pm$2.39} & \textbf{96.03{\scriptsize$\pm$0.25}} \\
 & LastFM & 62.67{\scriptsize$\pm$4.49} & 65.95{\scriptsize$\pm$5.98} & 67.48{\scriptsize$\pm$0.77} & 75.49{\scriptsize$\pm$0.00} & 68.12{\scriptsize$\pm$0.33} & 64.02{\scriptsize$\pm$1.48} & 72.24{\scriptsize$\pm$0.92} & 73.56$^\dagger$ & 77.99{\scriptsize$\pm$1.30} & 68.62{\scriptsize$\pm$0.43}$^\dagger$ & \underline{83.77{\scriptsize$\pm$1.26}}$^\dagger$ & 74.03{\scriptsize$\pm$0.79} & \textbf{89.50{\scriptsize$\pm$0.75}} \\
 & Enron & 68.96{\scriptsize$\pm$0.98} & 70.89{\scriptsize$\pm$2.72} & 75.15{\scriptsize$\pm$0.58} & 73.89{\scriptsize$\pm$0.00} & 75.01{\scriptsize$\pm$0.79} & 70.34{\scriptsize$\pm$1.73} & 80.23{\scriptsize$\pm$1.37} & 77.46{\scriptsize$\pm$0.21}$^\dagger$ & 75.36{\scriptsize$\pm$1.81} & 70.26{\scriptsize$\pm$0.95}$^\dagger$ & \underline{85.43{\scriptsize$\pm$0.29}}$^\dagger$ & 85.30{\scriptsize$\pm$0.67} & \textbf{89.00{\scriptsize$\pm$0.24}} \\
 & Social Evo. & 89.82{\scriptsize$\pm$4.11} & 95.13{\scriptsize$\pm$0.56} & 88.32{\scriptsize$\pm$0.27} & 83.69{\scriptsize$\pm$0.00} & 94.72{\scriptsize$\pm$0.33} & 92.12{\scriptsize$\pm$4.53} & 97.15{\scriptsize$\pm$0.25} & \underline{97.68{\scriptsize$\pm$0.10}} & 97.35{\scriptsize$\pm$0.32} & 82.72{\scriptsize$\pm$1.02}$^\dagger$ & 93.45{\scriptsize$\pm$0.52}$^\dagger$ & 97.56{\scriptsize$\pm$0.25} & \textbf{99.35{\scriptsize$\pm$0.02}} \\
 & UCI & 65.99{\scriptsize$\pm$1.40} & 70.94{\scriptsize$\pm$0.71} & 64.61{\scriptsize$\pm$0.48} & 57.43{\scriptsize$\pm$0.00} & 80.10{\scriptsize$\pm$0.51} & 57.55{\scriptsize$\pm$0.81} & \underline{83.61{\scriptsize$\pm$0.23}} & 72.05{\scriptsize$\pm$0.62}$^\dagger$ & 77.26{\scriptsize$\pm$1.57} & 75.44{\scriptsize$\pm$0.51}$^\dagger$ & 80.74{\scriptsize$\pm$0.29}$^\dagger$ & 83.39{\scriptsize$\pm$0.26} & \textbf{88.21{\scriptsize$\pm$0.16}} \\
 & Flights & 69.07{\scriptsize$\pm$4.02} & 71.09{\scriptsize$\pm$2.72} & 69.18{\scriptsize$\pm$1.52} & \underline{81.08{\scriptsize$\pm$0.00}} & 74.87{\scriptsize$\pm$0.21} & 59.16{\scriptsize$\pm$1.48} & 61.99{\scriptsize$\pm$1.70} & 70.92{\scriptsize$\pm$1.78} & 64.78{\scriptsize$\pm$1.50} & 65.48{\scriptsize$\pm$5.23}$^\dagger$ & 75.19{\scriptsize$\pm$0.17}$^\dagger$ & 71.59{\scriptsize$\pm$1.57} & \textbf{86.54{\scriptsize$\pm$0.19}} \\
 & Can. Parl. & 48.42{\scriptsize$\pm$0.66} & 65.34{\scriptsize$\pm$2.87} & 67.75{\scriptsize$\pm$1.00} & 62.16{\scriptsize$\pm$0.00} & 69.48{\scriptsize$\pm$0.63} & \textbf{78.03{\scriptsize$\pm$1.27}} & 59.15{\scriptsize$\pm$6.20} & 60.93{\scriptsize$\pm$0.93}$^\dagger$ & 71.34{\scriptsize$\pm$2.10}$^\dagger$ & 61.35{\scriptsize$\pm$5.64}$^\dagger$ & 54.74{\scriptsize$\pm$2.00}$^\dagger$ & 74.46{\scriptsize$\pm$2.55} & \underline{74.74{\scriptsize$\pm$2.78}} \\
 & US Legis. & 50.27{\scriptsize$\pm$5.13} & 67.57{\scriptsize$\pm$6.47} & 65.81{\scriptsize$\pm$8.52} & 64.74{\scriptsize$\pm$0.00} & 79.63{\scriptsize$\pm$0.84} & \textbf{88.40{\scriptsize$\pm$2.34}} & 67.78{\scriptsize$\pm$4.29} & \underline{81.25{\scriptsize$\pm$3.62}} & 75.87{\scriptsize$\pm$1.90}$^\dagger$ & 64.48{\scriptsize$\pm$2.39}$^\dagger$ & 55.43{\scriptsize$\pm$7.77}$^\dagger$ & 72.82{\scriptsize$\pm$9.66} & 77.62{\scriptsize$\pm$3.89} \\
 & UN Trade & 60.42{\scriptsize$\pm$1.48} & 61.04{\scriptsize$\pm$6.01} & 62.54{\scriptsize$\pm$0.67} & 72.97{\scriptsize$\pm$0.00} & 60.15{\scriptsize$\pm$1.29} & \underline{78.38{\scriptsize$\pm$2.24}} & 69.72{\scriptsize$\pm$1.84} & 55.79{\scriptsize$\pm$1.02} & 58.24{\scriptsize$\pm$0.64}$^\dagger$ & 52.62{\scriptsize$\pm$0.70}$^\dagger$ & 58.00{\scriptsize$\pm$1.61}$^\dagger$ & 67.30{\scriptsize$\pm$4.71} & \textbf{87.79{\scriptsize$\pm$0.70}} \\
 & UN Vote & 67.79{\scriptsize$\pm$1.46} & 67.63{\scriptsize$\pm$2.67} & 52.19{\scriptsize$\pm$0.34} & 66.30{\scriptsize$\pm$0.00} & 51.60{\scriptsize$\pm$0.73} & \textbf{72.53{\scriptsize$\pm$1.94}} & 68.37{\scriptsize$\pm$1.92} & 51.91{\scriptsize$\pm$0.84} & 55.91{\scriptsize$\pm$0.93}$^\dagger$ & 56.20{\scriptsize$\pm$2.88}$^\dagger$ & 50.64{\scriptsize$\pm$1.38}$^\dagger$ & 64.55{\scriptsize$\pm$3.36} & \underline{70.05{\scriptsize$\pm$1.39}} \\
 & Contact & 93.43{\scriptsize$\pm$1.78} & 90.18{\scriptsize$\pm$3.28} & 89.31{\scriptsize$\pm$0.27} & 85.20{\scriptsize$\pm$0.00} & 90.87{\scriptsize$\pm$0.35} & 91.13{\scriptsize$\pm$1.48} & 94.05{\scriptsize$\pm$0.52} & 94.75{\scriptsize$\pm$0.28} & 95.84{\scriptsize$\pm$0.32} & 71.35{\scriptsize$\pm$2.02}$^\dagger$ & 90.19{\scriptsize$\pm$0.58}$^\dagger$ & \underline{96.37{\scriptsize$\pm$0.26}} & \textbf{97.76{\scriptsize$\pm$0.07}} \\
\cmidrule{2-15}
 & Avg. Rank & 9.92 & 7.46 & 8.62 & 8.46 & 7.08 & 7.69 & 6.31 & 6.62 & 6.00 & 10.23 & 6.77 & \underline{4.31} & \textbf{1.54} \\
\midrule
\multicolumn{2}{l}{Overall Avg. Rank} & 9.49 & 7.24 & 9.44 & 9.71 & 7.92 & 6.69 & 5.18 & 5.91 & 4.55 & 10.72 & 7.38 & \underline{4.33} & \textbf{2.44} \\
\bottomrule
\end{tabular}}
\end{table}

\textbf{Predictive Performance.}
Table~\ref{tab:trans_ap} reports transductive AP under three negative sampling strategies. Our full model (Ours $+\lambda$) achieves the best overall average rank of 2.44, primarily benefiting from historical and inductive negatives. SNAM improves the respective average ranks from 4.42 and 4.31 to 1.50 and 1.54, surpassing all baselines and supporting the complementarity of activity and relation scoring. Without SNAM, our model still outranks the strongest baseline, TPNet (4.33 vs.\ 4.55). Under random negatives, both configurations perform similarly without a clear advantage over the strongest baseline, consistent with the theoretical conditions discussed in Appendix~\ref{app:activity_motivation}.
Corrections to several baseline implementations (Appendix~\ref{app:baseline_corrections}) yield both performance increases and decreases.
Additional results in Tables~\ref{tab:ind_ap}, \ref{tab:trans_auc} and~\ref{tab:ind_auc}, including inductive link prediction, show substantial gains under historical and inductive negatives on nearly all datasets.

\textbf{Computational Efficiency.}
On LastFM (Figure~\ref{fig:speed_lastfm}), our model achieves the highest historical-negative AP and lowest training and test-scoring times among the compared methods. TPNet and DyGFormer require approximately $9.2\times$ and $60\times$ our per-epoch training time, and $1.7\times$ and $27\times$ our full-test-set scoring time, respectively. These results support the efficiency of shared historical aggregation and within-block reuse, with larger-dataset advantages further demonstrated in Figure~\ref{fig:speed_appendix} and Appendix~\ref{app:efficiency_results}.
Timings use the same dedicated RTX 5090 and report medians over three runs; test scoring is timed under random negative sampling.

\subsection{SNAM: Does the Activity-Based Approach Really Work?}
We test whether independently modeling source activity provides useful information beyond relation scoring.

\Needspace{14\baselineskip}

\begin{wraptable}{r}{0.59\textwidth}
\vspace{-\intextsep}
\raggedleft
\caption{Source-activity transfer across backbones:
transductive test AP (\%), historical negative sampling.}
\label{tab:transfer}

\fontsize{6}{7}\selectfont
\renewcommand{\arraystretch}{0.95}
\setlength{\tabcolsep}{1.8pt}

\begin{tabular}{@{}llccccc@{}}
\toprule
backbone & score & Wikipedia & Reddit & UCI & Enron & LastFM \\
\midrule

\multirow{2}{*}{TPNet}
& alone
& 83.44{\tiny$\pm$0.93}
& 80.22{\tiny$\pm$0.21}
& 86.59{\tiny$\pm$1.15}
& 81.74{\tiny$\pm$0.56}
& 87.26{\tiny$\pm$0.90} \\
& $+\beta\log\lambda_u$
& \textbf{92.99{\tiny$\pm$0.06}}
& \textbf{90.79{\tiny$\pm$0.17}}
& \textbf{94.98{\tiny$\pm$0.12}}
& \textbf{91.38{\tiny$\pm$0.20}}
& \textbf{96.00{\tiny$\pm$0.02}} \\
\midrule

\multirow{2}{*}{TGN$^\dagger$}
& alone
& 86.79{\tiny$\pm$0.74}
& 80.23{\tiny$\pm$0.42}
& 80.10{\tiny$\pm$1.06}
& 73.89{\tiny$\pm$1.37}
& 77.31{\tiny$\pm$5.13} \\
& $+\beta\log\lambda_u$
& \textbf{92.52{\tiny$\pm$0.06}}
& \textbf{89.11{\tiny$\pm$0.09}}
& \textbf{93.35{\tiny$\pm$0.18}}
& \textbf{87.87{\tiny$\pm$0.53}}
& \textbf{93.70{\tiny$\pm$0.24}} \\
\midrule

\multirow{2}{*}{DyGFormer$^\dagger$}
& alone
& 76.01{\tiny$\pm$0.51}
& 81.32{\tiny$\pm$1.36}
& 81.59{\tiny$\pm$0.61}
& 76.60{\tiny$\pm$0.54}
& 81.61 \\
& $+\beta\log\lambda_u$
& \textbf{92.29{\tiny$\pm$0.05}}
& \textbf{90.30{\tiny$\pm$0.22}}
& \textbf{93.80{\tiny$\pm$0.16}}
& \textbf{89.04{\tiny$\pm$0.35}}
& \textbf{95.10} \\
\midrule

\multirow{2}{*}{DSRD$^\dagger$}
& alone
& 90.80{\tiny$\pm$0.19}
& 88.08{\tiny$\pm$0.63}
& 87.08{\tiny$\pm$0.39}
& 86.79{\tiny$\pm$0.14}
& 85.88{\tiny$\pm$1.52} \\
& $+\beta\log\lambda_u$
& \textbf{93.73{\tiny$\pm$0.05}}
& \textbf{92.88{\tiny$\pm$0.14}}
& \textbf{95.44{\tiny$\pm$0.09}}
& \textbf{93.14{\tiny$\pm$0.38}}
& \textbf{96.88{\tiny$\pm$0.01}} \\
\midrule

\multicolumn{2}{@{}l}{Ours ($f+\beta\log\lambda_u$)}
& 93.28{\tiny$\pm$0.08}
& 90.01{\tiny$\pm$0.25}
& 96.00{\tiny$\pm$0.15}
& 90.42{\tiny$\pm$0.64}
& 95.30{\tiny$\pm$0.09} \\
\bottomrule
\end{tabular}
\par
\end{wraptable}

\textbf{Cross-Model Transfer.}
Table~\ref{tab:transfer} applies SNAM parameters learned on each dataset to TPNet, TGN, DyGFormer, and DSRD, an unpublished 2026 preprint \citep{chang2026dsrd}. Both SNAM and the backbone remain frozen; only $\beta$ is selected on each backbone's validation split. Under historical negatives, AP improves across all 20 backbone--dataset combinations by up to 16.39 percentage points. Thus, SNAM complements different relation encoders without retraining either component. Transfer here means parameter reuse across models on the same dataset. Tables~\ref{tab:transfer_appendix} and~\ref{tab:transfer_tpnet_streams} provide additional results, and Appendix~\ref{app:activity_motivation} explains the rationale.

\begin{table*}[t]
\vspace{-\intextsep}
\centering
\begin{minipage}[t]{0.40\textwidth}
\caption{Effect of the number of time scales $K$ on transductive AP (\%).}
\label{tab:a1_scales}
\centering
\scriptsize
\setlength{\tabcolsep}{3pt}
\renewcommand{\arraystretch}{0.90}
\begin{tabular}{lcccc}
\toprule
Dataset & $K$ & rnd & hist & ind \\
\midrule
\multirow{4}{*}{\textit{UCI}}
& 1 & 83.62{\tiny$\pm$0.65} & 79.48{\tiny$\pm$0.41} & 84.60{\tiny$\pm$0.72} \\
& 8 & \textbf{96.18}{\tiny$\pm$0.14} & 95.43{\tiny$\pm$0.32} & 87.93{\tiny$\pm$0.41} \\
& 15$^\star$ & 95.81{\tiny$\pm$0.41} & 96.04{\tiny$\pm$0.17} & 88.18{\tiny$\pm$0.04} \\
& 28 & 95.86{\tiny$\pm$0.57} & \textbf{96.12}{\tiny$\pm$0.10} & \textbf{88.25}{\tiny$\pm$0.03} \\
\midrule
\multirow{4}{*}{\textit{Reddit}}
& 1 & 89.32{\tiny$\pm$0.05} & 77.50{\tiny$\pm$0.30} & 82.47{\tiny$\pm$0.11} \\
& 11 & \textbf{99.15}{\tiny$\pm$0.01} & 90.14{\tiny$\pm$0.13} & 91.78{\tiny$\pm$0.05} \\
& 16$^\star$ & \textbf{99.15}{\tiny$\pm$0.00} & \textbf{90.16}{\tiny$\pm$0.17} & \textbf{91.79}{\tiny$\pm$0.13} \\
& 32 & \textbf{99.15}{\tiny$\pm$0.01} & 90.01{\tiny$\pm$0.07} & 91.45{\tiny$\pm$0.17} \\
\bottomrule
\end{tabular}
\end{minipage}\hfill
\begin{minipage}[t]{0.58\textwidth}
\caption{Score component ablation under historical negative sampling: transductive AP (\%). $\dagger$: retrained by us with the corrected released code. \textbf{Bold}: best.}
\label{tab:i1_decomposition}
\centering
\scriptsize
\setlength{\tabcolsep}{2pt}
\renewcommand{\arraystretch}{1.167}
\begin{tabular}{lccccc}
\toprule
& Wikipedia & Reddit & UCI & Enron & LastFM \\
\midrule
DyGFormer$^\dagger$ & 76.01{\tiny$\pm$0.51} & 81.32{\tiny$\pm$1.35} & 81.59{\tiny$\pm$0.61} & 76.60{\tiny$\pm$0.54} & 81.61 \\
TPNet & 81.55{\tiny$\pm$4.10} & 81.02{\tiny$\pm$1.31} & 86.34{\tiny$\pm$0.80} & 80.79{\tiny$\pm$1.68} & 87.74{\tiny$\pm$0.50} \\
\midrule
$f$ only & 88.71{\tiny$\pm$0.80} & 82.74{\tiny$\pm$1.44} & 91.47{\tiny$\pm$0.52} & 84.04{\tiny$\pm$2.43} & 84.60{\tiny$\pm$1.20} \\
$\log\lambda_u$ only & 90.67{\tiny$\pm$0.01} & 80.65{\tiny$\pm$0.01} & 91.77{\tiny$\pm$0.01} & 83.59{\tiny$\pm$0.00} & 93.19{\tiny$\pm$0.00} \\
$f+\beta\log\lambda_u$ & \textbf{93.25{\tiny$\pm$0.05}} & \textbf{90.16{\tiny$\pm$0.17}} & \textbf{96.04{\tiny$\pm$0.17}} & \textbf{90.26{\tiny$\pm$0.77}} & \textbf{95.36{\tiny$\pm$0.03}} \\
\bottomrule
\end{tabular}
\end{minipage}
\end{table*}

\begin{wraptable}{R}{0.55\textwidth}
\vspace{-\intextsep}
\raggedleft
\caption{Effect of source-activity terms on transductive
test AP (\%) under inductive negative sampling.}
\label{tab:a3_transductive_inductive}

\fontsize{6}{7}\selectfont
\setlength{\tabcolsep}{1.2pt}
\renewcommand{\arraystretch}{0.95}

\begin{tabular}{@{}lccccc@{}}
\toprule
term $b$ & Wikipedia & Reddit & UCI & Enron & LastFM \\
\midrule

none ($f$ alone, $\beta=0$)
& 86.19{\tiny$\pm$1.21}
& 90.62{\tiny$\pm$0.27}
& 83.35{\tiny$\pm$0.29}
& 85.32{\tiny$\pm$0.85}
& 73.96{\tiny$\pm$0.95} \\

$\log\lambda_u$ (SNAM)$^\star$
& 91.23{\tiny$\pm$0.09}
& 91.79{\tiny$\pm$0.13}
& 88.18{\tiny$\pm$0.04}
& \textbf{89.04}{\tiny$\pm$0.18}
& 89.82{\tiny$\pm$0.71} \\
\midrule

$\log(1+n_u^{K})$
& 88.31{\tiny$\pm$0.23}
& 91.75{\tiny$\pm$0.16}
& 83.35{\tiny$\pm$0.29}
& 86.50{\tiny$\pm$0.78}
& 82.27{\tiny$\pm$0.85} \\

$\log(1+n_u^{K/2})$
& 89.44{\tiny$\pm$0.44}
& 91.26{\tiny$\pm$0.19}
& 86.85{\tiny$\pm$0.48}
& 88.93{\tiny$\pm$0.39}
& 85.23{\tiny$\pm$0.89} \\

$-\log(1+\Delta t_u)$
& 90.39{\tiny$\pm$0.32}
& 91.44{\tiny$\pm$0.24}
& \textbf{88.66}{\tiny$\pm$0.07}
& 87.47{\tiny$\pm$0.47}
& \textbf{89.93}{\tiny$\pm$0.70} \\

MLP on $\log(1+n_u)$
& \textbf{91.34}{\tiny$\pm$0.11}
& \textbf{91.86}{\tiny$\pm$0.13}
& 88.33{\tiny$\pm$0.06}
& 87.84{\tiny$\pm$0.43}
& \textbf{89.93}{\tiny$\pm$0.73} \\

\bottomrule
\end{tabular}
\par
\end{wraptable}
\textbf{Independence and Orthogonality.}
Table~\ref{tab:i1_decomposition} isolates SNAM: scoring with $\log\lambda_u(t)$ alone, using only $K+1$ learnable parameters, outperforms DyGFormer and TPNet on four of five datasets under historical negatives. Alternative activity models also generally improve upon relation-only scoring under inductive negatives (Table~\ref{tab:a3_transductive_inductive}), showing that the benefit extends beyond SNAM's implementation. Together with cross-model transfer, these results support source activity as an independently useful modeling perspective complementary to existing relation architectures.

\subsection{Parameter Sensitivity Analysis}
\textbf{Activity Term Coefficient $\beta$.} On MOOC (Figure~\ref{fig:beta_mooc}), transductive test AP peaks at $\beta=1.0$ under historical and inductive negatives, with no similar improvement under random negatives. SNAM leaves candidate rankings for a fixed source and timestamp unchanged, so nothing is lost on what random negatives test; the decline is an artifact of pooling sources in batch-level evaluation, whereas the two protocols that actually compare sources are the ones that gain.
 Appendix~\ref{app:activity_motivation} provides the formal analysis, and Figure~\ref{fig:beta_appendix} shows the remaining datasets.

\textbf{Number of Timescales $K$.} Table~\ref{tab:a1_scales} shows that moving from one timescale to several brings large gains, whereas adding more beyond about ten changes little. A moderate number of timescales therefore captures most of the benefit of multi-timescale historical modeling.

\subsection{Ablation Study}
Table~\ref{tab:a4_trunk_channels} retrains the model without the candidate-conditioned recent interaction encoding, the random-projection structural features, the pair and endpoint counts, or all three. Dropping the encoding or the projections changes the full score by under one point on every stream; dropping the counts costs up to two, and all three at most four once the activity term is in place. The advantage therefore lies in the multi-timescale backbone and the activity term, not in the engineered input features.

\section{Conclusion and Future Work}
Random destination negatives preserve the source and time, cancelling their shared distribution from the optimal score and leaving absolute source activity unconstrained. Historical and inductive negatives allow different sources, making activity informative. SNAM learns source intensity with $K+1$ parameters, outperforming DyGFormer and TPNet on four of five datasets under historical negatives. Added to four frozen backbones, it improves AP on almost every backbone--dataset pair by up to 25 percentage points. The full model ranks first overall against eleven baselines on 13 datasets and outpaces TPNet and DyGFormer on long streams.


The cancellation argument is not specific to source activity: any factor that a negative sampler holds fixed between a positive and its negatives drops out of the optimal score, so the sampler decides what a model can learn and what a metric can see. Negatives drawn at the positive's timestamp cancel global activity in the same way. This suggests auditing benchmarks for other cancelled factors, designing objectives and protocols around the comparisons a deployment actually makes, and learning such factors separately, as done here, rather than expecting the encoder to recover them.

\section*{AI use statement}

In this work, we used generative AI tools to assist in the writing of proofs, design and provide feedback on research methodology and experiments, implement methods, assist with translation, and support qualitative and thematic data analysis. We did not use generative AI tools for any other applicable tasks requiring disclosure. Generating synthetic datasets was not applicable to this work.

Additionally, we used generative AI tools to create or modify scientific figures and images and to edit the manuscript to improve readability.

We carefully reviewed all AI-assisted work. In particular, four human authors independently reviewed the LLM-generated code to rigorously verify that neither the main experiments nor the ablation studies involved data leakage or misleading implementation practices. We take full responsibility for the final content of this work, including all text, claims, and artifacts produced with the assistance of generative AI.

\section*{Ethics statement}

This work focuses on methodological advances in temporal
link prediction and evaluates the proposed approach on
existing benchmark datasets. We do not identify any
ethical concerns specific to the proposed methodology.
As with other methods applied to temporal interaction
data, downstream applications should respect applicable
privacy and data-use requirements.

\section*{Reproducibility statement}

Reproducibility was the first priority of this work, and a large part of its effort went into it. Every number in this paper, in every table and figure of the main text and the appendix, can be checked by the reader in two ways: the released result file of every run rebuilds all tables and figures byte for byte with one script, and the released training and evaluation scripts rerun every experiment from the public benchmark data, including our model, the retrained baselines, the transfer study, the ablations and the timing runs. Nothing reported here depends on a run that cannot be repeated. Every training configuration and seed is released (Appendix~\ref{app:our_training}, Table~\ref{tab:training_settings}), and a retraining from scratch should land within seed variance of the reported values, as it did in our own checks: a configuration retrained on another machine reproduced its six reported AP values within 0.35 points, a shipped checkpoint re-evaluated elsewhere reproduced them to five decimals, and the transfer pipeline reproduced the backbone-only results exactly (Appendix~\ref{app:reproducibility}). A script downloads the data from its Zenodo record; dataset statistics and evaluation protocols are documented in Appendix~\ref{app:datasets_protocols}, and the split, negative samplers and metric code are those of DyGLib, unmodified. Baseline numbers are either published values or produced by our own runs of the released code, and Table~\ref{tab:baseline_sources} states which. Where a released implementation let information from the query's own timestamp or label reach the score, Appendix~\ref{app:baseline_corrections} names the path, describes the correction, and explains why the model was retrained rather than re-evaluated; every change is marked in the released source code so that it can be audited against the original. The transfer procedure is specified in Appendix~\ref{app:transfer_details}. The theoretical claims are proved in Appendices~\ref{app:activity_motivation}, \ref{app:activity_likelihood} and~\ref{app:exponential_kernels}.



\bibliography{iclr2027_conference}

@inproceedings{yu2023dyglib,
  title     = {Towards Better Dynamic Graph Learning: New Architecture and Unified Library},
  author    = {Yu, Le and Sun, Leilei and Du, Bowen and Lv, Weifeng},
  booktitle = {Advances in Neural Information Processing Systems},
  volume    = {36},
  year      = {2023},
  url       = {https://proceedings.neurips.cc/paper_files/paper/2023/hash/d611019afba70d547bd595e8a4158f55-Abstract.html}
}

@inproceedings{poursafaei2022towards,
  title     = {Towards Better Evaluation for Dynamic Link Prediction},
  author    = {Poursafaei, Farimah and Huang, Shenyang and Pelrine, Kellin and Rabbany, Reihaneh},
  booktitle = {Advances in Neural Information Processing Systems},
  volume    = {35},
  year      = {2022},
  note      = {Datasets and Benchmarks Track},
  url       = {https://proceedings.neurips.cc/paper_files/paper/2022/hash/d49042a5d49818711c401d34172f9900-Abstract-Datasets_and_Benchmarks.html}
}

@inproceedings{lu2024tpnet,
  title     = {Improving Temporal Link Prediction via Temporal Walk Matrix Projection},
  author    = {Lu, Xiaodong and Sun, Leilei and Zhu, Tongyu and Lv, Weifeng},
  booktitle = {Advances in Neural Information Processing Systems},
  volume    = {37},
  year      = {2024},
  url       = {https://proceedings.neurips.cc/paper_files/paper/2024/hash/ff7bf6014f7826da531aa50f4538ee19-Abstract-Conference.html}
}

@inproceedings{chang2026grn,
  title     = {Graph Retention Networks for Dynamic Graphs},
  author    = {Chang, Qian and Li, Xia and Cheng, Xiufeng and Jia, Runsong and Yang, Jinqing and Hu, Guoping and Giurcaneanu, Ciprian Doru},
  booktitle = {Proceedings of the ACM Web Conference 2026},
  publisher = {Association for Computing Machinery},
  year      = {2026},
  doi       = {10.1145/3774904.3792107},
  url       = {https://doi.org/10.1145/3774904.3792107}
}

@article{chang2026dsrd,
  title     = {Forget Less, Generalize More: Unifying Temporal and Structural Adaptation for Dynamic Graphs},
  author    = {Chang, Qian and Giurcaneanu, Ciprian Doru and Jia, Runsong and Li, Xia and Hu, Guoping and Cheng, Xiufeng and Yang, Jinqing and Wu, Mengjia and Zhang, Yi},
  journal   = {arXiv preprint arXiv:2605.29453},
  year      = {2026},
  url       = {https://arxiv.org/abs/2605.29453}
}

@inproceedings{zaheer2017deepsets,
  title     = {Deep Sets},
  author    = {Zaheer, Manzil and Kottur, Satwik and Ravanbakhsh, Siamak and Poczos, Barnabas and Salakhutdinov, Ruslan and Smola, Alexander J.},
  booktitle = {Advances in Neural Information Processing Systems},
  volume    = {30},
  pages     = {3391--3401},
  year      = {2017},
  url       = {https://papers.nips.cc/paper/2017/hash/f22e4747da1aa27e363d86d40ff442fe-Abstract.html}
}

@inproceedings{qi2017pointnet,
  title     = {{PointNet}: Deep Learning on Point Sets for {3D} Classification and Segmentation},
  author    = {Qi, Charles R. and Su, Hao and Mo, Kaichun and Guibas, Leonidas J.},
  booktitle = {Proceedings of the IEEE Conference on Computer Vision and Pattern Recognition},
  year      = {2017},
  doi       = {10.1109/CVPR.2017.16},
  url       = {https://openaccess.thecvf.com/content_cvpr_2017/html/Qi_PointNet_Deep_Learning_CVPR_2017_paper.html}
}

@inproceedings{wagstaff2019limitations,
  title     = {On the Limitations of Representing Functions on Sets},
  author    = {Wagstaff, Edward and Fuchs, Fabian and Engelcke, Martin and Posner, Ingmar and Osborne, Michael A.},
  booktitle = {Proceedings of the 36th International Conference on Machine Learning},
  series    = {Proceedings of Machine Learning Research},
  volume    = {97},
  pages     = {6487--6494},
  publisher = {PMLR},
  year      = {2019},
  url       = {https://proceedings.mlr.press/v97/wagstaff19a.html}
}

@inproceedings{kumar2019predicting,
  title     = {Predicting Dynamic Embedding Trajectory in Temporal Interaction Networks},
  author    = {Kumar, Srijan and Zhang, Xikun and Leskovec, Jure},
  booktitle = {Proceedings of the 25th ACM SIGKDD International Conference on Knowledge Discovery and Data Mining},
  pages     = {1269--1278},
  publisher = {Association for Computing Machinery},
  year      = {2019},
  doi       = {10.1145/3292500.3330895},
  url       = {https://doi.org/10.1145/3292500.3330895}
}

@inproceedings{trivedi2019dyrep,
  title     = {{DyRep}: Learning Representations over Dynamic Graphs},
  author    = {Trivedi, Rakshit and Farajtabar, Mehrdad and Biswal, Prasenjeet and Zha, Hongyuan},
  booktitle = {International Conference on Learning Representations},
  year      = {2019},
  url       = {https://openreview.net/forum?id=HyePrhR5KX}
}

@inproceedings{rossi2020temporal,
  title     = {Temporal Graph Networks for Deep Learning on Dynamic Graphs},
  author    = {Rossi, Emanuele and Chamberlain, Ben and Frasca, Fabrizio and Eynard, Davide and Monti, Federico and Bronstein, Michael},
  booktitle = {ICML 2020 Workshop on Graph Representation Learning},
  year      = {2020},
  url       = {https://arxiv.org/abs/2006.10637}
}

@inproceedings{loshchilov2019decoupled,
  title     = {Decoupled Weight Decay Regularization},
  author    = {Loshchilov, Ilya and Hutter, Frank},
  booktitle = {International Conference on Learning Representations},
  year      = {2019},
  url       = {https://openreview.net/forum?id=Bkg6RiCqY7}
}

@inproceedings{xu2020inductive,
  title     = {Inductive Representation Learning on Temporal Graphs},
  author    = {Xu, Da and Ruan, Chuanwei and Korpeoglu, Evren and Kumar, Sushant and Achan, Kannan},
  booktitle = {International Conference on Learning Representations},
  year      = {2020},
  url       = {https://openreview.net/forum?id=rJeW1yHYwH}
}

@inproceedings{wang2021causal,
  title     = {Inductive Representation Learning in Temporal Networks via Causal Anonymous Walks},
  author    = {Wang, Yanbang and Chang, Yen-Yu and Liu, Yunyu and Leskovec, Jure and Li, Pan},
  booktitle = {International Conference on Learning Representations},
  year      = {2021},
  url       = {https://arxiv.org/abs/2101.05974}
}

@inproceedings{luo2022nat,
  title     = {Neighborhood-Aware Scalable Temporal Network Representation Learning},
  author    = {Luo, Yuhong and Li, Pan},
  booktitle = {Proceedings of the First Learning on Graphs Conference},
  series    = {Proceedings of Machine Learning Research},
  volume    = {198},
  pages     = {1:1--1:18},
  publisher = {PMLR},
  year      = {2022},
  url       = {https://proceedings.mlr.press/v198/luo22a.html}
}

@inproceedings{souza2022provably,
  title     = {Provably Expressive Temporal Graph Networks},
  author    = {Souza, Amauri H. and Mesquita, Diego and Kaski, Samuel and Garg, Vikas},
  booktitle = {Advances in Neural Information Processing Systems},
  volume    = {35},
  year      = {2022},
  url       = {https://proceedings.neurips.cc/paper_files/paper/2022/hash/d029c97ee0db162c60f2ebc9cb93387e-Abstract-Conference.html}
}

@inproceedings{cong2023graphmixer,
  title     = {Do We Really Need Complicated Model Architectures for Temporal Networks?},
  author    = {Cong, Weilin and Zhang, Si and Kang, Jian and Yuan, Baichuan and Wu, Hao and Zhou, Xin and Tong, Hanghang and Mahdavi, Mehrdad},
  booktitle = {International Conference on Learning Representations},
  year      = {2023},
  url       = {https://openreview.net/forum?id=ayPPc0SyLv1}
}

@inproceedings{li2025dygmamba,
  title     = {{DyG-Mamba}: Continuous State Space Modeling on Dynamic Graphs},
  author    = {Li, Dongyuan and Tan, Shiyin and Zhang, Ying and Jin, Ming and Pan, Shirui and Okumura, Manabu and Jiang, Renhe},
  booktitle = {Advances in Neural Information Processing Systems},
  volume    = {38},
  year      = {2025},
  doi       = {10.52202/085713-4299},
  url       = {https://arxiv.org/abs/2408.06966}
}

@inproceedings{yi2025craft,
  title     = {Future Link Prediction Without Memory or Aggregation},
  author    = {Yi, Lu and Lei, Runlin and Mo, Fengran and Zheng, Yanping and Wei, Zhewei and Ye, Yuhang},
  booktitle = {Advances in Neural Information Processing Systems},
  volume    = {38},
  year      = {2025},
  doi       = {10.52202/085713-0084},
  url       = {https://proceedings.neurips.cc/paper_files/paper/2025/hash/036912a83bdbb1fd792baf6532f102d8-Abstract-Conference.html}
}

@article{chung2025timeencoder,
  title     = {Between Linear and Sinusoidal: Rethinking the Time Encoder in Dynamic Graph Learning},
  author    = {Chung, Hsing-Huan and Chaudhari, Shravan S. and Han, Xing and Wald, Yoav and Saria, Suchi and Ghosh, Joydeep},
  journal   = {Transactions on Machine Learning Research},
  year      = {2025},
  url       = {https://openreview.net/forum?id=W6GQvdOGHg}
}

@inproceedings{tian2024freedyg,
  title     = {{FreeDyG}: Frequency Enhanced Continuous-Time Dynamic Graph Model for Link Prediction},
  author    = {Tian, Yuxing and Qi, Yiyan and Guo, Fan},
  booktitle = {International Conference on Learning Representations},
  year      = {2024},
  url       = {https://openreview.net/forum?id=82Mc5ilInM}
}

@article{hawkes1971spectra,
  title     = {Spectra of Some Self-Exciting and Mutually Exciting Point Processes},
  author    = {Hawkes, Alan G.},
  journal   = {Biometrika},
  volume    = {58},
  number    = {1},
  pages     = {83--90},
  year      = {1971},
  doi       = {10.1093/biomet/58.1.83},
  url       = {https://doi.org/10.1093/biomet/58.1.83}
}

@inproceedings{mei2017neural,
  title     = {The Neural {Hawkes} Process: A Neurally Self-Modulating Multivariate Point Process},
  author    = {Mei, Hongyuan and Eisner, Jason},
  booktitle = {Advances in Neural Information Processing Systems},
  volume    = {30},
  year      = {2017},
  url       = {https://papers.nips.cc/paper_files/paper/2017/hash/6463c88460bd63bbe256e495c63aa40b-Abstract.html}
}

@inproceedings{trivedi2017knowevolve,
  title     = {{Know-Evolve}: Deep Temporal Reasoning for Dynamic Knowledge Graphs},
  author    = {Trivedi, Rakshit and Dai, Hanjun and Wang, Yichen and Song, Le},
  booktitle = {Proceedings of the 34th International Conference on Machine Learning},
  series    = {Proceedings of Machine Learning Research},
  volume    = {70},
  pages     = {3462--3471},
  publisher = {PMLR},
  year      = {2017},
  url       = {https://proceedings.mlr.press/v70/trivedi17a.html}
}

@inproceedings{zuo2018embedding,
  title     = {Embedding Temporal Network via Neighborhood Formation},
  author    = {Zuo, Yuan and Liu, Guannan and Lin, Hao and Guo, Jia and Hu, Xiaoqian and Wu, Junjie},
  booktitle = {Proceedings of the 24th ACM SIGKDD International Conference on Knowledge Discovery and Data Mining},
  pages     = {2857--2866},
  publisher = {Association for Computing Machinery},
  year      = {2018},
  doi       = {10.1145/3219819.3220054},
  url       = {https://doi.org/10.1145/3219819.3220054}
}

@inproceedings{zuo2020transformer,
  title     = {Transformer {Hawkes} Process},
  author    = {Zuo, Simiao and Jiang, Haoming and Li, Zichong and Zhao, Tuo and Zha, Hongyuan},
  booktitle = {Proceedings of the 37th International Conference on Machine Learning},
  series    = {Proceedings of Machine Learning Research},
  volume    = {119},
  pages     = {11692--11702},
  publisher = {PMLR},
  year      = {2020},
  url       = {https://proceedings.mlr.press/v119/zuo20a.html}
}

@inproceedings{wen2022trend,
  title     = {{TREND}: {TempoRal} Event and Node Dynamics for Graph Representation Learning},
  author    = {Wen, Zhihao and Fang, Yuan},
  booktitle = {Proceedings of the ACM Web Conference 2022},
  pages     = {1159--1169},
  publisher = {Association for Computing Machinery},
  year      = {2022},
  doi       = {10.1145/3485447.3512164},
  url       = {https://doi.org/10.1145/3485447.3512164}
}

@inproceedings{huang2023tgb,
  title     = {Temporal Graph Benchmark for Machine Learning on Temporal Graphs},
  author    = {Huang, Shenyang and Poursafaei, Farimah and Danovitch, Jacob and Fey, Matthias and Hu, Weihua and Rossi, Emanuele and Leskovec, Jure and Bronstein, Michael and Rabusseau, Guillaume and Rabbany, Reihaneh},
  booktitle = {Advances in Neural Information Processing Systems},
  volume    = {36},
  year      = {2023},
  note      = {Datasets and Benchmarks Track},
  url       = {https://proceedings.neurips.cc/paper_files/paper/2023/hash/066b98e63313162f6562b35962671288-Abstract.html}
}

@inproceedings{cornell2025heuristics,
  title     = {On the Power of Heuristics in Temporal Graphs},
  author    = {Cornell, Filip and Smirnov, Oleg and Gandler, Gabriela Zarzar and Cao, Lele},
  booktitle = {Proceedings on "I Can't Believe It's Not Better: Challenges in Applied Deep Learning" at ICLR 2025 Workshops},
  series    = {Proceedings of Machine Learning Research},
  volume    = {296},
  pages     = {37--46},
  publisher = {PMLR},
  year      = {2025},
  url       = {https://proceedings.mlr.press/v296/cornell25a.html}
}

@inproceedings{daniluk2023global,
  title     = {Temporal Graph Models Fail to Capture Global Temporal Dynamics},
  author    = {Daniluk, Micha{\l} and D{\k{a}}browski, Jacek},
  booktitle = {Temporal Graph Learning Workshop at NeurIPS 2023},
  year      = {2023},
  url       = {https://openreview.net/forum?id=Ks94Yn5jqY}
}

@article{kazemi2019time2vec,
  title     = {{Time2Vec}: Learning a Vector Representation of Time},
  author    = {Kazemi, Seyed Mehran and Goel, Rishab and Eghbali, Sepehr and Ramanan, Janahan and Sahota, Jaspreet and Thakur, Sanjay and Wu, Stella and Smyth, Cathal and Poupart, Pascal and Brubaker, Marcus},
  journal   = {arXiv preprint arXiv:1907.05321},
  year      = {2019},
  url       = {https://arxiv.org/abs/1907.05321}
}

@inproceedings{xu2019functional,
  title     = {Self-attention with Functional Time Representation Learning},
  author    = {Xu, Da and Ruan, Chuanwei and Korpeoglu, Evren and Kumar, Sushant and Achan, Kannan},
  booktitle = {Advances in Neural Information Processing Systems},
  volume    = {32},
  year      = {2019},
  url       = {https://papers.nips.cc/paper_files/paper/2019/hash/cf34645d98a7630e2bcca98b3e29c8f2-Abstract.html}
}

@inproceedings{omi2019fully,
  title     = {Fully Neural Network based Model for General Temporal Point Processes},
  author    = {Omi, Takahiro and Ueda, Naonori and Aihara, Kazuyuki},
  booktitle = {Advances in Neural Information Processing Systems},
  volume    = {32},
  year      = {2019},
  url       = {https://proceedings.neurips.cc/paper/2019/hash/39e4973ba3321b80f37d9b55f63ed8b8-Abstract.html}
}

@inproceedings{shchur2020intensity,
  title     = {Intensity-Free Learning of Temporal Point Processes},
  author    = {Shchur, Oleksandr and Bilo{\v{s}}, Marin and G{\"u}nnemann, Stephan},
  booktitle = {International Conference on Learning Representations},
  year      = {2020},
  url       = {https://openreview.net/forum?id=HygOjhEYDH}
}

@inproceedings{gastinger2024tgb2,
  title     = {{TGB} 2.0: A Benchmark for Learning on Temporal Knowledge Graphs and Heterogeneous Graphs},
  author    = {Gastinger, Julia and Huang, Shenyang and Galkin, Mikhail and Loghmani, Erfan and Parviz, Ali and Poursafaei, Farimah and Danovitch, Jacob and Rossi, Emanuele and Koutis, Ioannis and Stuckenschmidt, Heiner and Rabbany, Reihaneh and Rabusseau, Guillaume},
  booktitle = {Advances in Neural Information Processing Systems},
  volume    = {37},
  year      = {2024},
  note      = {Datasets and Benchmarks Track},
  url       = {https://proceedings.neurips.cc/paper_files/paper/2024/hash/fda026cf2423a01fcbcf1e1e43ee9a50-Abstract-Datasets_and_Benchmarks_Track.html}
}

@inproceedings{xue2024easytpp,
  title     = {{EasyTPP}: Towards Open Benchmarking Temporal Point Processes},
  author    = {Xue, Siqiao and Shi, Xiaoming and Chu, Zhixuan and Wang, Yan and Hao, Hongyan and Zhou, Fan and Jiang, Caigao and Pan, Chen and Zhang, James Y. and Wen, Qingsong and Zhou, Jun and Mei, Hongyuan},
  booktitle = {International Conference on Learning Representations},
  year      = {2024},
  url       = {https://proceedings.iclr.cc/paper_files/paper/2024/hash/e66233a208ef32f56df6312263239fa0-Abstract-Conference.html}
}

@inproceedings{liu2025salom,
  title     = {{SALoM}: Structure Aware Temporal Graph Networks with Long-Short Memory Updater},
  author    = {Liu, Hanwen and Zhang, Longjiao and Wang, Rui and Zheng, Tongya and Wu, Sai and Yao, Chang and Song, Mingli},
  booktitle = {Advances in Neural Information Processing Systems},
  volume    = {38},
  year      = {2025},
  doi       = {10.52202/085713-0770},
  url       = {https://papers.nips.cc/paper_files/paper/2025/hash/20f94998511f25bb6378cae0e098bc46-Abstract-Conference.html}
}

@inproceedings{tieu2025lstep,
  title     = {Learnable Spatial-Temporal Positional Encoding for Link Prediction},
  author    = {Tieu, Katherine and Fu, Dongqi and Li, Zihao and Maciejewski, Ross and He, Jingrui},
  booktitle = {Proceedings of the 42nd International Conference on Machine Learning},
  series    = {Proceedings of Machine Learning Research},
  volume    = {267},
  pages     = {59570--59597},
  publisher = {PMLR},
  year      = {2025},
  url       = {https://proceedings.mlr.press/v267/tieu25a.html}
}

@inproceedings{xiong2025tetgn,
  title     = {Trajectory Encoding Temporal Graph Networks},
  author    = {Xiong, Jiafeng and Sakellariou, Rizos},
  booktitle = {Knowledge Graphs: 14th International Joint Conference, IJCKG 2025, Proceedings},
  series    = {Lecture Notes in Computer Science},
  volume    = {16297},
  pages     = {236--251},
  publisher = {Springer Nature},
  year      = {2026},
  doi       = {10.1007/978-981-95-5009-8_16},
  url       = {https://doi.org/10.1007/978-981-95-5009-8_16}
}

@article{zhou2025advances,
  title     = {Advances in Temporal Point Processes: {Bayesian}, Neural, and {LLM} Approaches},
  author    = {Zhou, Feng and Kong, Quyu and Qiao, Jie and Wan, Cheng and Zhang, Yixuan and Cai, Ruichu},
  journal   = {Transactions on Machine Learning Research},
  year      = {2026},
  url       = {https://openreview.net/forum?id=SXgGKkShhT}
}

@inproceedings{yi2025tgbseq,
  title     = {{TGB-Seq} Benchmark: Challenging Temporal {GNNs} with Complex Sequential Dynamics},
  author    = {Yi, Lu and Peng, Jie and Zheng, Yanping and Mo, Fengran and Wei, Zhewei and Ye, Yuhang and Yue, Zixuan and Huang, Zengfeng},
  booktitle = {International Conference on Learning Representations},
  year      = {2025},
  url       = {https://proceedings.iclr.cc/paper_files/paper/2025/hash/db5ca61dbc08cf5143c05ad2d1b0b2ca-Abstract-Conference.html}
}

@inproceedings{dizaji2026tgrab,
  title     = {{T-GRAB}: A Synthetic Diagnostic Benchmark for Learning on Temporal Graphs},
  author    = {Dizaji, Alireza and Tjandra, Benedict Aaron and Hamidi, Mehrab and Huang, Shenyang and Rabusseau, Guillaume},
  booktitle = {Proceedings of the 32nd ACM SIGKDD Conference on Knowledge Discovery and Data Mining V.1},
  pages     = {2628--2639},
  publisher = {Association for Computing Machinery},
  year      = {2026},
  doi       = {10.1145/3770854.3785699},
  url       = {https://doi.org/10.1145/3770854.3785699}
}
\bibliographystyle{iclr2027_conference}

\appendix

\section{Statistical Motivation and Mathematical Foundations of Source Activity Modeling}
\label{app:snam_foundations}

Our source activity model adopts a multi-timescale exponential
Hawkes-type intensity \citep{hawkes1971spectra}. Unlike neural
temporal-graph point-process models such as DyRep
\citep{trivedi2019dyrep}, however, it models source activity
independently of learned relation representations, using only
$K+1$ shared intensity parameters. For completeness, we derive
the relevant point-process formulation and likelihood from
first principles, without assuming prior familiarity with
Hawkes processes.

\subsection{From Negative Sampling to Source Activity Correction}
\label{app:activity_motivation}

Temporal link prediction is usually formulated as follows: given the interaction history and a candidate node pair, predict whether the pair will interact at the current time. However, a real interaction involves two different questions: \emph{Will the source node initiate an interaction at the current time? And, if it does, which destination node will it choose?} The first question concerns the rate at which events occur, while the second concerns the conditional relationship between nodes. Separating these two factors is the starting point of Source Node Activity Modeling (SNAM).

This distinction is closely related to how temporal link prediction models are trained and evaluated. In the DyGLib-style setting adopted in this work, a random destination negative sample $(u,v^-,t)$ is constructed from a real interaction $(u,v,t)$ by keeping the source node and timestamp unchanged and replacing only the destination node. In contrast, historical and inductive negative sampling draw complete negative edges from their respective candidate sets. As a result, a negative sample may have a different source node. Although both settings are formulated as binary classification, they require the model to use different information.

We first explain this distinction from a probabilistic perspective and then derive the motivation for adding an independent source-activity correction to the relation prediction score.

\begin{figure}[p]
\centering
\includegraphics[width=\linewidth]{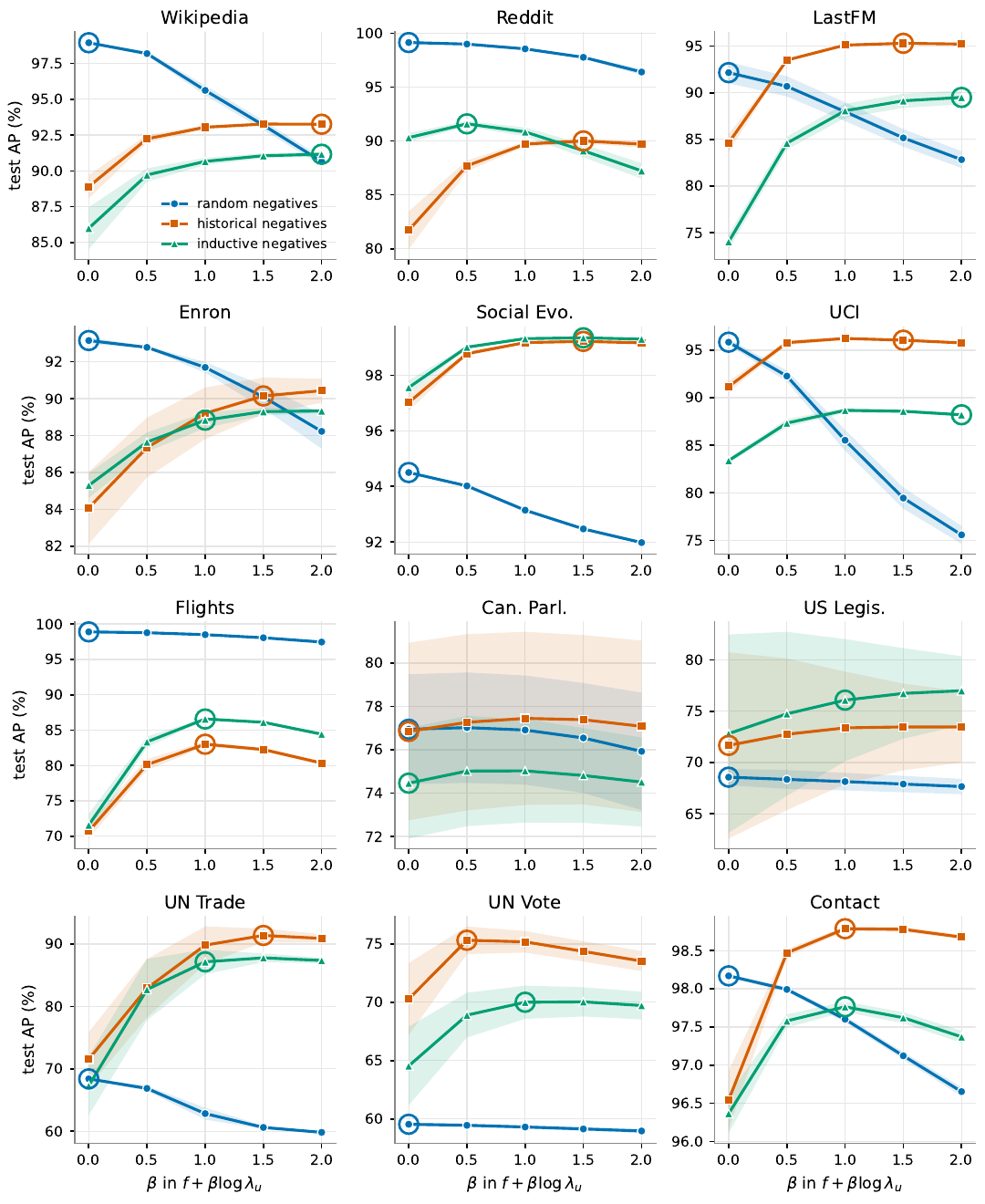}
\caption{Test AP versus $\beta$ for the remaining twelve datasets, same conventions as Figure~\ref{fig:beta_mooc}. The decline under random negatives has the same protocol origin as on MOOC and vanishes when candidates of one source are ranked. The wide bands on Can.\ Parl.\ and US Legis.\ are seed variance of the trunk itself.}
\label{fig:beta_appendix}
\end{figure}

\paragraph{A real interaction involves two different probabilistic factors.}

Consider a query time $t$ and suppose the model can observe only the interaction history strictly preceding $t$:
\[
\mathcal H_{t^-}=\{e_i:t_i<t\}.
\]

Let $\lambda_u(t)$ denote the \emph{conditional event intensity} of source node $u$ given this history. Intuitively, it measures how likely node $u$ is to initiate an interaction at the current time. For a sufficiently short interval of length $\Delta t>0$,
\begin{equation}
\begin{aligned}
&\Pr\bigl(
u\text{ initiates one interaction in }[t,t+\Delta t)
\mid\mathcal H_{t^-}
\bigr)\\
&\qquad =
\lambda_u(t)\Delta t+o(\Delta t).
\end{aligned}
\tag{A.1}
\label{eq:app-a1}
\end{equation}

Here, we assume that the probability of multiple events occurring within the same short interval is $o(\Delta t)$. Importantly, $\lambda_u(t)$ is an event rate per unit time, not a probability. It is $\lambda_u(t)\Delta t$ that approximates the probability of an event occurring within a short time interval.

If we further specify a destination node $v$, we obtain another statistical quantity: the \emph{pairwise conditional intensity} $\lambda_{uv}(t)$. It describes the instantaneous rate at which source node $u$ interacts with destination node $v$, given the current history:
\begin{equation}
\begin{aligned}
&\Pr\bigl(
u\to v\text{ occurs in }[t,t+\Delta t)
\mid\mathcal H_{t^-}
\bigr)\\
&\qquad =
\lambda_{uv}(t)\Delta t+o(\Delta t).
\end{aligned}
\tag{A.2}
\label{eq:app-a2}
\end{equation}

Since every interaction initiated by source node $u$ has a destination node, we have
\[
\lambda_u(t)=\sum_v\lambda_{uv}(t).
\]

When $\lambda_u(t)>0$, define
\[
\pi_{uv}(t)
=
\frac{\lambda_{uv}(t)}{\lambda_u(t)}.
\]

Then $\sum_v\pi_{uv}(t)=1$, and $\pi_{uv}(t)$ can be interpreted as the probability of choosing destination node $v$, given the history and the fact that source node $u$ initiates an interaction at the current time. We therefore obtain
\begin{equation}
\boxed{
\lambda_{uv}(t)=\lambda_u(t)\pi_{uv}(t).
}
\tag{A.3}
\label{eq:app-a3}
\end{equation}

\Eqref{eq:app-a3} separates the full pairwise interaction intensity into two factors with different meanings. The source intensity $\lambda_u(t)$ describes how active the source node is at the current time, while $\pi_{uv}(t)$ describes which destination the source is likely to choose once an interaction occurs.

This decomposition also reveals a potential difficulty in predicting current interactions from historical relationships alone. Two nodes may have interacted frequently in the past and therefore have a strong conditional relationship. However, if the source node has been inactive for a long time, the current event rate of that pair may still be low, even if the destination remains consistent with the source's historical preferences. Conversely, a highly active source node does not necessarily choose any particular destination.

Thus, if our goal is to compare how likely complete node pairs are to interact at the current time, we should, in principle, consider both factors. The question is whether commonly used negative-sampling objectives actually require a model to learn both of them.

\paragraph{What does random destination negative sampling actually supervise?}

To answer this question, we first consider a general binary classification problem.

Let $x$ denote a sample to be classified. Suppose real samples follow a distribution $p_+(x)$, while artificial negative samples follow a distribution $p_-(x)$. Assume that the two classes occur equally often during training and that a real-valued discriminator $g(x)$ is trained using binary cross-entropy. Its population risk is
\begin{equation}
\begin{aligned}
\mathcal R(g)
={}&-\mathbb E_{x\sim p_+}\log\sigma(g(x))\\
&-\mathbb E_{x\sim p_-}
\log\!\left(1-\sigma(g(x))\right),
\end{aligned}
\tag{A.4}
\label{eq:app-a4}
\end{equation}
where $\sigma(z)=1/(1+e^{-z})$.

An important point is that this binary classifier does not directly learn the probability that a real interaction will occur. Instead, it learns to distinguish real samples from samples generated by a particular negative sampler. Consequently, the optimal discriminator depends not only on the distribution of real interactions, but also on how negative samples are generated.

We can derive this result directly from~\eqref{eq:app-a4}. If the discriminator has sufficient expressive capacity, its risk can be minimized separately at each $x$ where both densities are positive. The corresponding pointwise objective is
\[
\ell_x(g)
=
-p_+(x)\log\sigma(g)
-p_-(x)\log(1-\sigma(g)).
\]

Differentiating with respect to $g$ gives
\[
\frac{\partial\ell_x}{\partial g}
=
-p_+(x)(1-\sigma(g))
+p_-(x)\sigma(g).
\]

Setting the derivative to zero and rearranging, we obtain
\[
\sigma(g^*(x))
=
\frac{p_+(x)}{p_+(x)+p_-(x)}.
\]

Applying the inverse sigmoid function yields
\begin{equation}
\boxed{
g^*(x)
=
\log\frac{p_+(x)}{p_-(x)}.
}
\tag{A.5}
\label{eq:app-a5}
\end{equation}

Therefore, under ideal population-risk minimization, binary cross-entropy learns the \emph{log-density ratio} between the real distribution and the negative-sampling distribution. This means that changing the negative-sampling protocol also changes the optimal discriminator, even when the real interaction data remain exactly the same.

We now apply this result to random destination negative sampling.

For each real interaction $(u,v,t)$, the negative sampler keeps $u$ and $t$ unchanged and independently draws a destination from a distribution $q_R(v)$. For simplicity, we omit explicit conditioning on the observed history in the distributions below.

The joint distribution of real interactions can be factorized as
\[
p_+(u,v,t)
=
p_+(u,t)\pi_{uv}(t).
\]

Since a negative sample inherits the source node and timestamp of a real interaction, its distribution satisfies
\[
p_-^R(u,v,t)
=
p_+(u,t)q_R(v).
\]

Substituting these distributions into~\eqref{eq:app-a5}, we obtain the ideal discriminator under random negative sampling:
\begin{equation}
\begin{aligned}
g_R^*(u,v,t)
&=
\log
\frac{p_+(u,t)\pi_{uv}(t)}
{p_+(u,t)q_R(v)}
\\[3pt]
&=
\boxed{
\log\pi_{uv}(t)-\log q_R(v).
}
\end{aligned}
\tag{A.6}
\label{eq:app-a6}
\end{equation}

The key observation in~\eqref{eq:app-a6} is that the joint source-time distribution $p_+(u,t)$ \emph{cancels exactly} in the density ratio.

This cancellation follows directly from the random negative-sampling mechanism. For a given real interaction, the positive sample and its randomly generated negative sample have exactly the same source node and timestamp. Whether the source is currently highly active or has been inactive for a long time, its activity level is shared by both samples. It therefore cannot help the discriminator distinguish this matched positive-negative pair.

When $q_R$ is uniform over a fixed set of destination nodes, $-\log q_R(v)$ becomes a constant that does not depend on the candidate destination. In this case, the ideal discriminator only needs to capture $\log\pi_{uv}(t)$: the relative preference for different destinations, given that the source node initiates an interaction at the current time.

This does not mean that a neural network trained with random negative samples cannot encode any activity-related information. Its node representations may still contain historical interaction frequencies, recent activity patterns, and other relevant statistics. The more precise conclusion from~\eqref{eq:app-a6} is that \emph{the random destination negative-sampling objective does not require the model to recover the absolute source event rate $\lambda_u(t)$.} Even a model that perfectly solves this training task cannot determine, from its discrimination score alone, how likely the source node is to initiate an interaction at the current time.

\paragraph{When the negative sample has a different source node.}

The cancellation above relies on a special property of random destination negative sampling: negative samples inherit both the source node and the query time of their corresponding positive samples.

Under historical and inductive negative sampling, however, negative pairs are drawn from candidate sets defined by the respective protocols. Their source nodes are no longer required to match those of the corresponding positive samples. The model may therefore need to compare source nodes with very different activity levels at the current time.

For example, a historical negative edge may connect two nodes that interacted frequently in the past. Because the edge genuinely appeared in the interaction history, a model based on historical neighborhoods and pairwise relationships may still assign it a high compatibility score. Yet, if its source node rarely initiates interactions at the current time, the actual event rate of this pair may be low. In this situation, source activity provides useful information that historical relation compatibility alone may not fully capture.

We can understand this difference more precisely through the optimal discriminator.

Fix a query time $t$ and the history preceding it. Define the total event intensity across all source nodes as
\[
\Lambda(t)=\sum_u\lambda_u(t).
\]

Conditioned on a real interaction occurring near the current time, the probability that its node pair is $(u,v)$ is
\begin{equation}
\begin{aligned}
p_+(u,v\mid t)
&=
\frac{\lambda_{uv}(t)}{\Lambda(t)}
\\
&=
\frac{\lambda_u(t)\pi_{uv}(t)}
{\Lambda(t)}.
\end{aligned}
\tag{A.7}
\label{eq:app-a7}
\end{equation}

Let $q_E(u,v\mid t)$ denote the negative-pair distribution induced by a particular hard negative-sampling protocol under the corresponding evaluation conditions. Applying~\eqref{eq:app-a5} on the common support of the positive and negative distributions gives
\begin{equation}
\begin{aligned}
g_E^*(u,v,t)
&=
\log\frac{p_+(u,v\mid t)}
{q_E(u,v\mid t)}
\\[3pt]
&=
\boxed{
\log\lambda_u(t)
+
\log\pi_{uv}(t)
-
\log\Lambda(t)
-
\log q_E(u,v\mid t).
}
\end{aligned}
\tag{A.8}
\label{eq:app-a8}
\end{equation}

Comparing~\eqref{eq:app-a6} and~\eqref{eq:app-a8} reveals the central difference between the two negative-sampling mechanisms.

Under random destination negative sampling, negative samples inherit the true source distribution, causing the source-activity factor to cancel automatically in the density ratio. When a hard negative-sampling protocol allows the source node to change, this cancellation is no longer guaranteed by the sampling mechanism. Source activity can therefore become an important factor in distinguishing real interactions from artificial negative edges.

In particular, consider an idealized setting where negative node pairs are sampled uniformly. If we compare candidate pairs at the same query time, both the sampling term and the total-intensity normalization term in~\eqref{eq:app-a8} are independent of the specific pair. The pair-dependent part of the ideal ranking score then becomes
\[
\log\lambda_u(t)+\log\pi_{uv}(t)
=
\log\lambda_{uv}(t),
\]
which is precisely the log-intensity of the complete node pair.

Actual historical and inductive negative sampling do not always satisfy these simplifying conditions. Their candidate sets may depend on the interaction history, evaluation batch, and exclusion rules, and their sampling procedures may include additional fallback mechanisms. As a result, $-\log q_E(u,v\mid t)$ is generally not constant. Comparisons across different query times may also be affected by the normalization term $-\log\Lambda(t)$.

\Eqref{eq:app-a8} therefore does not imply that adding source activity alone exactly recovers the optimal discriminator for every hard negative-sampling protocol. Rather, it identifies \emph{why source activity can become useful discriminative information and where this information appears mathematically in the optimal score}.

\paragraph{Recovering source-activity information from conditional destination preference.}

We can now return to the decomposition of the full interaction intensity in~\eqref{eq:app-a3}.

Random destination negative sampling primarily supervises conditional destination preference, whereas the event intensity of a complete node pair also depends on source activity. Combining ~\eqref{eq:app-a6} and~\eqref{eq:app-a3}, we obtain
\begin{equation}
\begin{aligned}
&g_R^*(u,v,t)+\log\lambda_u(t)
\\
&\quad=
\log\pi_{uv}(t)
-\log q_R(v)
+\log\lambda_u(t)
\\
&\quad=
\boxed{
\log\lambda_{uv}(t)-\log q_R(v).
}
\end{aligned}
\tag{A.9}
\label{eq:app-a9}
\end{equation}

When $q_R$ is uniform, the right-hand side differs from the log-intensity of the complete node pair only by a constant. This provides a direct statistical motivation for source-activity correction: independently estimate the current source event rate and use it to supplement the conditional destination preference learned by the relation model.

Following this idea, we combine the base score $f(u,v,t)$ produced by the relation branch with the source intensity estimated by SNAM:
\begin{equation}
\boxed{
s_\beta(u,v,t)
=
f(u,v,t)+\beta\log\lambda_u(t).
}
\tag{A.10}
\label{eq:app-a10}
\end{equation}

Here, $\beta\geq0$ controls the contribution of source-activity information. Our relation branch uses a smooth aggregation of two outputs during training and the base score $f$ during evaluation. Since a finite-capacity model need not produce the ideal density ratio in~\eqref{eq:app-a6}, and different evaluation protocols have different negative-sampling distributions, we select $\beta$ on the validation set rather than treating it as a theoretically optimal constant shared by all protocols.

It is also worth noting that, for a fixed source node $u$ and query time $t$, $\log\lambda_u(t)$ is the same additive term for every candidate destination. Thus, it does not change the ranking of destinations within that source-time query. Instead, it mainly contributes activity information when comparing samples associated with different source nodes or query times. In a global ranking that combines multiple queries, this information can change the relative ordering of samples.

SNAM is therefore not an arbitrary additional historical feature attached to a relation model. It provides an activity component with a clear statistical role, motivated by the difference between the information required by the random negative-sampling objective and the factors governing the occurrence of complete interactions.

Moreover, this component does not require a separate, complex temporal history encoder. The multi-timescale exponential historical states already maintained by Multi-Timescale Historical Representation (MTHR) yield exactly the decayed activity counts required by SNAM when the event content is set to a constant. This shared mathematical foundation allows relation representation and source-activity modeling to reuse historical computations while retaining independent training objectives.
\paragraph{Scope of the cancellation.}
The blindness to source activity is created during training, not by the evaluation protocol. The cancellation in~\eqref{eq:app-a6} uses one property only: a training negative keeps the source and timestamp of its positive. Every method compared here trains this way, as do other benchmarks such as TGB \citep{huang2023tgb}, so the factor is never rewarded regardless of architecture; the historical and inductive protocols do not create the gap but merely make it visible, because they are the first comparisons across sources, which is also what a deployment ranks. The signature appears on all 13 datasets: under random negatives the validation split selects $\beta=0$ in 123 of 130 cases (13 datasets, five seeds, two settings), whereas under historical and inductive negatives it selects $\beta>0$ in 111 and 116. A pattern that repeats across thirteen streams of different domains, sizes and time resolutions is a property of the objective, not a quirk of one protocol or one dataset. Because the gap belongs to the objective rather than to any model, the remedy transfers: the same activity term raises every backbone it is added to. Nor is source activity special: any factor a sampler holds fixed between a positive and its negatives drops out of the optimal discriminator, a constraint for the design of objectives and evaluation protocols.

\begin{table}[t]
\caption{Transfer of the source-activity term to TPNet on the eight streams outside the transfer study: test AP (\%) of the backbone alone and with $\beta\log\lambda_u$ added, in the transductive setting under historical (hist) and inductive (ind) negatives and in the inductive setting under historical negatives (its inductive protocol coincides), one seed per stream. $\lambda_u$ keeps the parameters fitted for our model on the same stream; $\beta$ is chosen on TPNet's validation split, and $\Delta$ is the change. Under random negatives $\beta=0$ is selected on every stream and setting except UN Vote in the inductive setting ($\beta=0.5$, 56.15 to 57.31); the random columns are therefore omitted. The streams on which the term does not help (Can.\ Parl., US Legis., UN Vote) are the ones on which it does not help our own model either (Tables~\ref{tab:trans_ap} and~\ref{tab:ind_ap}).}
\label{tab:transfer_tpnet_streams}
\centering
\scriptsize
\setlength{\tabcolsep}{4pt}
\renewcommand{\arraystretch}{0.95}
\begin{tabular}{lccccccccc}
\toprule
 & \multicolumn{3}{c}{Transductive, hist} & \multicolumn{3}{c}{Transductive, ind} & \multicolumn{3}{c}{Inductive, hist} \\
\cmidrule(lr){2-4} \cmidrule(lr){5-7} \cmidrule(lr){8-10}
Dataset & alone & $+\lambda$ & $\Delta$ & alone & $+\lambda$ & $\Delta$ & alone & $+\lambda$ & $\Delta$ \\
\midrule
MOOC & 93.30 & 99.42 & +6.12 & 88.81 & 97.78 & +8.97 & 83.49 & 97.02 & +13.52 \\
Social Evo. & 96.91 & 99.20 & +2.29 & 97.42 & 99.32 & +1.90 & 96.33 & 98.51 & +2.18 \\
Flights & 69.18 & 83.91 & +14.73 & 65.48 & 85.31 & +19.83 & 55.51 & 70.18 & +14.67 \\
Can. Parl. & 75.19 & 75.19 & +0.00 & 70.74 & 70.74 & +0.00 & 57.20 & 57.20 & +0.00 \\
US Legis. & 73.16 & 73.72 & +0.55 & 73.41 & 76.33 & +2.92 & 47.28 & 47.28 & +0.00 \\
UN Trade & 62.90 & 88.17 & +25.28 & 57.51 & 83.27 & +25.75 & 52.51 & 64.99 & +12.48 \\
UN Vote & 65.02 & 66.76 & +1.74 & 56.92 & 62.25 & +5.33 & 51.70 & 49.28 & -2.42 \\
Contact & 97.88 & 99.02 & +1.14 & 95.61 & 97.55 & +1.94 & 92.51 & 96.91 & +4.40 \\
\bottomrule
\end{tabular}
\end{table}

\subsection{The Shared Mathematical Foundation of Source Activity and Multi-Timescale Historical Representation}
\label{app:shared_historical_foundation}

The previous section showed that conditional destination preference and source activity are two distinct factors in interaction prediction. Random destination negative sampling primarily supervises the former, while the latter can provide important additional information when an evaluation protocol allows negative samples to have different source nodes. This raises a natural question: if we need to estimate the source intensity $\lambda_u(t)$ independently, must we design a separate temporal history encoder for it?

One possible approach would be to use a recurrent neural network, temporal attention, or another sequential model to encode the complete interaction history of each source node and predict its current activity rate from the resulting representation. However, this would introduce additional parameters and computational costs. It would also make the activity estimator dependent on a particular history encoder, limiting its ability to serve as an independent and reusable correction module.

Our design takes advantage of a more direct connection: \emph{the historical aggregation used by Multi-Timescale Historical Representation (MTHR) captures not only information about past interaction partners. When the content of each interaction is set to a constant, the same aggregation operator yields the multi-timescale activity counts needed by Source Node Activity Modeling (SNAM).} Thus, the historical states already maintained by MTHR contain the basic statistics required to construct SNAM. We only need to give these statistics an explicit interpretation in terms of event rates and learn how much each of them contributes to the current source intensity.

\paragraph{From historical interaction content to historical interaction counts.}

We first consider the general form of historical aggregation used by MTHR.

For a node $u$, let $e\ni u$ denote an interaction incident to $u$, let $t_e$ be its timestamp, and let $\psi_u(e)$ represent the content contributed by this interaction to node $u$. This content may be the embedding of the other endpoint, an interaction feature vector, or simply a scalar.

Given a fixed decay rate $\omega_k>0$ for the $k$-th timescale, define the historical aggregation operator as
\begin{equation}
\boxed{
H_{u,\psi}^{k}(t)
=
\sum_{\substack{e\ni u\\t_e<t}}
e^{-\omega_k(t-t_e)}\psi_u(e).
}
\tag{A.11}
\label{eq:app-a11}
\end{equation}

This operator has a simple interpretation. Each historical interaction contributes some content, but the contribution of an older interaction receives a smaller weight. Different timescales use different decay rates $\omega_k$, allowing the model to retain information from recent, intermediate, and more distant history. All interactions in the sum must satisfy $t_e<t$. Consequently, neither the interaction being predicted nor any other interaction with the same timestamp can enter the historical state.

\Eqref{eq:app-a11} separates two design choices that are easy to conflate. The exponential weight determines how historical information decays over time, while $\psi_u(e)$ determines what information we extract from each interaction. Changing the interaction content does not require changing the mathematical form of historical aggregation.

For example, let $p_u(e)$ denote the other endpoint of interaction $e$, and let $E_{p_u(e)}$ be its base node embedding. If we choose
\[
\psi_u(e)=E_{p_u(e)},
\]
then~\eqref{eq:app-a11} becomes a time-weighted sum of historical interaction partners' embeddings. This is the basic form of first-order neighborhood aggregation in MTHR: it captures which nodes have interacted with $u$ in the past and how long ago those interactions occurred.

Now consider a much simpler choice:
\[
\psi_u(e)=1.
\]

Each historical interaction now contributes a time-decayed unit count rather than a node embedding. Substituting this choice into the same aggregation operator gives
\begin{equation}
\boxed{
n_u^k(t)
=
H_{u,1}^{k}(t)
=
\sum_{\substack{e\ni u\\t_e<t}}
e^{-\omega_k(t-t_e)}.
}
\tag{A.12}
\label{eq:app-a12}
\end{equation}

These are exactly the multi-timescale activity counts already maintained by MTHR.

The relationship between first-order historical aggregation and activity counting is therefore deeper than two models happening to use the same exponential function. They are \emph{two instances of the same historical aggregation operator, applied to different interaction contents}. The former focuses on information carried by historical interaction partners, while the latter focuses on the number and temporal distribution of past interactions.

This distinction is particularly important for activity modeling. Two nodes may have very similar interaction partners but substantially different activity levels. Conversely, two nodes with similar numbers of interactions may have entirely different interaction partners. Retaining both historical content and historical counts allows relation modeling and activity modeling to use a common historical foundation while focusing on different predictive factors.

\paragraph{How the same historical states serve both MTHR and SNAM.}

The connection above is not limited to the mathematical definitions. It is also reflected directly in the computations performed by MTHR.

Recall the block-start historical aggregation introduced in the main text. Let $a_c$ denote the starting time of block $c$, and let $A_c^k$ be the time-decayed adjacency matrix constructed from interactions preceding $a_c$ at the $k$-th timescale. Its entries are
\[
A_c^k[u,w]
=
\sum_{\substack{
e\ni u,\ p_u(e)=w\\
t_e<a_c
}}
e^{-\omega_k(a_c-t_e)}.
\]

Given the base node embedding table $E$, MTHR computes its first-order historical aggregation at the beginning of the block:
\[
X_{1,c}^k=A_c^kE.
\]

At the same time, let $\mathbf 1$ denote the all-ones column vector. The block-start activity counts can be obtained using the same adjacency matrix:
\begin{equation}
\boxed{
X_{1,c}^k=A_c^kE,
\qquad
n_c^k=A_c^k\mathbf 1.
}
\tag{A.13}
\label{eq:app-a13}
\end{equation}

These two matrix operations use exactly the same historical weights. The only difference is the content being aggregated. The first operation aggregates node embeddings, while the second aggregates unit scalars. In other words, when MTHR constructs its historical weighted graph, it already has all the information needed to compute the activity counts.

Activity counts are not statistics introduced solely for SNAM. They are already used in several parts of MTHR, including the normalization of first- and second-order historical representations and the decay of the cold-start contribution. They also appear as endpoint activity features in Candidate Relation Modeling (CRM). Using these counts to estimate source intensity therefore does not require the backbone to perform another independent scan of the interaction history or maintain a separate set of learned temporal hidden states.

Moreover, the exponential kernel allows these counts to be updated exactly from the beginning of a block to any query time.

For a query time $t\geq a_c$ within block $c$, the interaction history can be divided into two parts: old interactions that occurred before the block started, and newly observed interactions that occurred within the block but strictly before the query time. Since
\[
e^{-\omega_k(t-t_e)}
=
e^{-\omega_k(t-a_c)}
e^{-\omega_k(a_c-t_e)}
\]
holds for every $t_e<a_c$,~\eqref{eq:app-a12} can be rewritten exactly as
\begin{equation}
\boxed{
n_u^k(t)
=
e^{-\omega_k(t-a_c)}n_c^k[u]
+
\sum_{\substack{
e\ni u\\a_c\leq t_e<t
}}
e^{-\omega_k(t-t_e)}.
}
\tag{A.14}
\label{eq:app-a14}
\end{equation}

The first term decays the previously aggregated block-start count to the current query time. The second term only needs to account for newly observed interactions within the block. Thus, even when a node has a long interaction history, we do not need to revisit its entire history for every query.

Importantly,~\eqref{eq:app-a14} is an exact identity for the exponential activity counts defined in this work, rather than an approximation. To share computations across queries, the default MTHR configuration may reuse its block-start first- and second-order node representations without immediately incorporating the embeddings of newly observed interaction partners. However, the activity counts still incorporate every interaction strictly preceding the query time according to~\eqref{eq:app-a14}. SNAM therefore reads the complete historical activity statistics available at the query time, rather than simply using the activity level recorded at the beginning of the block.
\begin{table}[p]
\caption{Number of training blocks and within-block updates, transductive and inductive link prediction: test AP (\%), mean{\scriptsize$\pm$std} over three seeds. $C$ = number of training blocks; $\star$ = the configuration of the main results. With within-block updates on, the events of the current block that precede the query time enter the counts, the feature states and the random projections; off, only the block-start state, decayed to the query time, is used. $f$ = the base score alone ($\beta=0$), $f{+}\lambda$ = with the source-activity term ($\beta$ chosen on validation). Bold: best of the six rows of a dataset. In the inductive setting both samplers draw from the same new-node subset, so only the historical columns are shown.}
\label{tab:a2_event_clock}
\centering
\footnotesize
\setlength{\tabcolsep}{2.5pt}
\renewcommand{\arraystretch}{0.92}
\begin{tabular}{ccccccccc}
\toprule
 & \multicolumn{5}{c}{Transductive} & \multicolumn{3}{c}{Inductive} \\
\cmidrule(lr){2-6} \cmidrule(lr){7-9}
$C$ & rnd & hist $f$ & hist $f{+}\lambda$ & ind $f$ & ind $f{+}\lambda$ & rnd & hist $f$ & hist $f{+}\lambda$ \\
\midrule
\multicolumn{9}{l}{\textit{Wikipedia, within-block updates on}} \\
48 & 98.92{\scriptsize$\pm$0.02} & 88.65{\scriptsize$\pm$0.46} & \textbf{93.29}{\scriptsize$\pm$0.05} & 85.20{\scriptsize$\pm$1.50} & 91.15{\scriptsize$\pm$0.12} & 98.35{\scriptsize$\pm$0.01} & 78.11{\scriptsize$\pm$1.50} & \textbf{88.35}{\scriptsize$\pm$0.13} \\
96$^\star$ & \textbf{98.95}{\scriptsize$\pm$0.02} & 88.71{\scriptsize$\pm$0.80} & 93.25{\scriptsize$\pm$0.05} & \textbf{86.19}{\scriptsize$\pm$1.21} & \textbf{91.23}{\scriptsize$\pm$0.09} & 98.41{\scriptsize$\pm$0.03} & 77.78{\scriptsize$\pm$1.86} & 88.29{\scriptsize$\pm$0.13} \\
192 & \textbf{98.95}{\scriptsize$\pm$0.04} & 86.98{\scriptsize$\pm$0.90} & 93.18{\scriptsize$\pm$0.18} & 83.11{\scriptsize$\pm$2.41} & 91.03{\scriptsize$\pm$0.26} & \textbf{98.42}{\scriptsize$\pm$0.03} & 75.75{\scriptsize$\pm$2.93} & 88.34{\scriptsize$\pm$0.01} \\
\cmidrule(lr){1-9}
\multicolumn{9}{l}{\textit{Wikipedia, within-block updates off}} \\
48 & 98.72{\scriptsize$\pm$0.03} & 88.28{\scriptsize$\pm$0.98} & 89.26{\scriptsize$\pm$0.48} & 85.21{\scriptsize$\pm$0.71} & 87.48{\scriptsize$\pm$0.23} & 98.09{\scriptsize$\pm$0.02} & 78.44{\scriptsize$\pm$0.80} & 79.14{\scriptsize$\pm$0.52} \\
96 & 98.77{\scriptsize$\pm$0.00} & \textbf{89.69}{\scriptsize$\pm$0.63} & 90.23{\scriptsize$\pm$0.45} & 85.48{\scriptsize$\pm$1.55} & 87.10{\scriptsize$\pm$0.79} & 98.19{\scriptsize$\pm$0.01} & \textbf{80.03}{\scriptsize$\pm$2.29} & 80.47{\scriptsize$\pm$1.67} \\
192 & 98.83{\scriptsize$\pm$0.05} & 87.08{\scriptsize$\pm$1.23} & 89.72{\scriptsize$\pm$0.38} & 82.89{\scriptsize$\pm$1.91} & 86.04{\scriptsize$\pm$0.51} & 98.20{\scriptsize$\pm$0.06} & 77.60{\scriptsize$\pm$1.68} & 79.63{\scriptsize$\pm$1.04} \\
\midrule
\multicolumn{9}{l}{\textit{Reddit, within-block updates on}} \\
48$^\star$ & 99.15{\scriptsize$\pm$0.00} & 82.74{\scriptsize$\pm$1.44} & \textbf{90.16}{\scriptsize$\pm$0.17} & \textbf{90.62}{\scriptsize$\pm$0.27} & \textbf{91.79}{\scriptsize$\pm$0.13} & 98.22{\scriptsize$\pm$0.02} & 67.54{\scriptsize$\pm$2.74} & 79.20{\scriptsize$\pm$0.34} \\
96 & 99.16{\scriptsize$\pm$0.01} & 82.23{\scriptsize$\pm$1.35} & 90.07{\scriptsize$\pm$0.02} & 90.23{\scriptsize$\pm$0.36} & 91.57{\scriptsize$\pm$0.05} & 98.20{\scriptsize$\pm$0.04} & 67.09{\scriptsize$\pm$2.05} & 79.21{\scriptsize$\pm$0.13} \\
192 & \textbf{99.18}{\scriptsize$\pm$0.01} & 82.96{\scriptsize$\pm$1.14} & 90.08{\scriptsize$\pm$0.13} & 90.52{\scriptsize$\pm$0.31} & 91.71{\scriptsize$\pm$0.19} & \textbf{98.24}{\scriptsize$\pm$0.03} & \textbf{69.01}{\scriptsize$\pm$1.35} & \textbf{79.36}{\scriptsize$\pm$0.14} \\
\cmidrule(lr){1-9}
\multicolumn{9}{l}{\textit{Reddit, within-block updates off}} \\
48 & 99.11{\scriptsize$\pm$0.03} & \textbf{83.23}{\scriptsize$\pm$0.29} & 86.35{\scriptsize$\pm$0.15} & 90.19{\scriptsize$\pm$0.26} & 91.17{\scriptsize$\pm$0.02} & 98.03{\scriptsize$\pm$0.09} & 67.85{\scriptsize$\pm$0.35} & 71.14{\scriptsize$\pm$0.21} \\
96 & 99.13{\scriptsize$\pm$0.02} & 81.27{\scriptsize$\pm$0.31} & 85.89{\scriptsize$\pm$0.04} & 89.88{\scriptsize$\pm$0.39} & 90.77{\scriptsize$\pm$0.33} & 98.08{\scriptsize$\pm$0.03} & 66.16{\scriptsize$\pm$1.47} & 70.63{\scriptsize$\pm$0.48} \\
192 & 99.15{\scriptsize$\pm$0.01} & 82.45{\scriptsize$\pm$1.25} & 86.92{\scriptsize$\pm$0.21} & 90.28{\scriptsize$\pm$0.16} & 90.87{\scriptsize$\pm$0.17} & 98.09{\scriptsize$\pm$0.03} & 68.68{\scriptsize$\pm$1.46} & 72.08{\scriptsize$\pm$0.61} \\
\midrule
\multicolumn{9}{l}{\textit{UCI, within-block updates on}} \\
48 & 95.79{\scriptsize$\pm$0.32} & 90.77{\scriptsize$\pm$0.34} & 95.61{\scriptsize$\pm$0.19} & 82.87{\scriptsize$\pm$0.19} & 87.90{\scriptsize$\pm$0.11} & 94.67{\scriptsize$\pm$0.09} & 83.61{\scriptsize$\pm$0.18} & 88.76{\scriptsize$\pm$0.10} \\
96 & 95.59{\scriptsize$\pm$0.02} & 91.11{\scriptsize$\pm$0.54} & 95.96{\scriptsize$\pm$0.09} & 83.30{\scriptsize$\pm$0.23} & 87.90{\scriptsize$\pm$0.23} & 94.70{\scriptsize$\pm$0.09} & \textbf{84.13}{\scriptsize$\pm$0.22} & 88.85{\scriptsize$\pm$0.18} \\
192$^\star$ & \textbf{95.81}{\scriptsize$\pm$0.41} & \textbf{91.47}{\scriptsize$\pm$0.52} & \textbf{96.04}{\scriptsize$\pm$0.17} & \textbf{83.35}{\scriptsize$\pm$0.29} & \textbf{88.18}{\scriptsize$\pm$0.04} & \textbf{94.74}{\scriptsize$\pm$0.12} & 84.08{\scriptsize$\pm$0.35} & \textbf{89.03}{\scriptsize$\pm$0.05} \\
\cmidrule(lr){1-9}
\multicolumn{9}{l}{\textit{UCI, within-block updates off}} \\
48 & 94.81{\scriptsize$\pm$0.32} & 89.01{\scriptsize$\pm$0.22} & 89.01{\scriptsize$\pm$0.22} & 82.52{\scriptsize$\pm$0.19} & 82.52{\scriptsize$\pm$0.19} & 93.84{\scriptsize$\pm$0.05} & 82.72{\scriptsize$\pm$0.24} & 82.72{\scriptsize$\pm$0.24} \\
96 & 94.77{\scriptsize$\pm$0.48} & 90.37{\scriptsize$\pm$0.18} & 91.56{\scriptsize$\pm$0.16} & 83.08{\scriptsize$\pm$0.40} & 83.08{\scriptsize$\pm$0.40} & 93.99{\scriptsize$\pm$0.20} & 83.41{\scriptsize$\pm$0.49} & 83.41{\scriptsize$\pm$0.49} \\
192 & 95.26{\scriptsize$\pm$0.31} & 90.62{\scriptsize$\pm$0.71} & 92.35{\scriptsize$\pm$0.26} & 82.87{\scriptsize$\pm$0.04} & 82.87{\scriptsize$\pm$0.04} & 94.02{\scriptsize$\pm$0.20} & 83.32{\scriptsize$\pm$0.17} & 83.32{\scriptsize$\pm$0.17} \\
\midrule
\multicolumn{9}{l}{\textit{Enron, within-block updates on}} \\
48 & 92.86{\scriptsize$\pm$0.17} & 73.85{\scriptsize$\pm$3.02} & 89.13{\scriptsize$\pm$0.50} & 79.09{\scriptsize$\pm$2.17} & 86.85{\scriptsize$\pm$1.12} & \textbf{89.74}{\scriptsize$\pm$0.28} & 69.11{\scriptsize$\pm$0.97} & 84.29{\scriptsize$\pm$0.39} \\
96 & 92.86{\scriptsize$\pm$0.21} & 79.75{\scriptsize$\pm$1.44} & 89.82{\scriptsize$\pm$0.11} & 83.39{\scriptsize$\pm$0.78} & \textbf{89.37}{\scriptsize$\pm$0.16} & 89.70{\scriptsize$\pm$0.03} & 73.45{\scriptsize$\pm$1.81} & 84.55{\scriptsize$\pm$0.63} \\
192$^\star$ & \textbf{93.09}{\scriptsize$\pm$0.21} & \textbf{84.04}{\scriptsize$\pm$2.43} & \textbf{90.26}{\scriptsize$\pm$0.77} & \textbf{85.32}{\scriptsize$\pm$0.85} & 89.04{\scriptsize$\pm$0.18} & 89.51{\scriptsize$\pm$0.10} & \textbf{78.56}{\scriptsize$\pm$2.85} & \textbf{85.57}{\scriptsize$\pm$0.91} \\
\cmidrule(lr){1-9}
\multicolumn{9}{l}{\textit{Enron, within-block updates off}} \\
48 & 92.77{\scriptsize$\pm$0.27} & 76.04{\scriptsize$\pm$3.64} & 82.49{\scriptsize$\pm$1.11} & 80.08{\scriptsize$\pm$2.45} & 84.32{\scriptsize$\pm$1.42} & 89.38{\scriptsize$\pm$0.38} & 70.57{\scriptsize$\pm$2.89} & 74.42{\scriptsize$\pm$1.91} \\
96 & 92.40{\scriptsize$\pm$0.29} & 76.90{\scriptsize$\pm$3.60} & 84.13{\scriptsize$\pm$1.21} & 81.36{\scriptsize$\pm$2.46} & 84.61{\scriptsize$\pm$1.73} & 89.20{\scriptsize$\pm$0.23} & 70.45{\scriptsize$\pm$3.53} & 75.47{\scriptsize$\pm$1.90} \\
192 & \textbf{93.09}{\scriptsize$\pm$0.08} & 82.39{\scriptsize$\pm$5.58} & 88.13{\scriptsize$\pm$1.23} & 84.32{\scriptsize$\pm$3.19} & 87.55{\scriptsize$\pm$0.83} & 89.49{\scriptsize$\pm$0.26} & 76.47{\scriptsize$\pm$5.29} & 81.26{\scriptsize$\pm$1.61} \\
\midrule
\multicolumn{9}{l}{\textit{LastFM, within-block updates on}} \\
48 & 90.30{\scriptsize$\pm$3.57} & 81.19{\scriptsize$\pm$2.41} & 94.86{\scriptsize$\pm$0.50} & 70.43{\scriptsize$\pm$3.01} & 89.48{\scriptsize$\pm$0.07} & 88.87{\scriptsize$\pm$6.30} & 73.36{\scriptsize$\pm$5.85} & 89.27{\scriptsize$\pm$0.65} \\
96 & 91.17{\scriptsize$\pm$2.96} & 83.32{\scriptsize$\pm$0.86} & 95.32{\scriptsize$\pm$0.05} & 72.01{\scriptsize$\pm$1.12} & \textbf{89.83}{\scriptsize$\pm$0.57} & 90.12{\scriptsize$\pm$5.24} & 76.09{\scriptsize$\pm$2.46} & \textbf{90.13}{\scriptsize$\pm$0.57} \\
192$^\star$ & 91.87{\scriptsize$\pm$1.35} & \textbf{84.60}{\scriptsize$\pm$1.20} & \textbf{95.36}{\scriptsize$\pm$0.03} & \textbf{73.96}{\scriptsize$\pm$0.95} & 89.82{\scriptsize$\pm$0.71} & 91.60{\scriptsize$\pm$2.72} & \textbf{78.38}{\scriptsize$\pm$0.70} & 90.12{\scriptsize$\pm$0.61} \\
\cmidrule(lr){1-9}
\multicolumn{9}{l}{\textit{LastFM, within-block updates off}} \\
48 & \textbf{92.58}{\scriptsize$\pm$0.38} & 82.17{\scriptsize$\pm$1.68} & 87.80{\scriptsize$\pm$0.39} & 71.12{\scriptsize$\pm$0.75} & 78.48{\scriptsize$\pm$0.29} & \textbf{93.24}{\scriptsize$\pm$0.33} & 76.53{\scriptsize$\pm$0.39} & 77.96{\scriptsize$\pm$0.38} \\
96 & 89.42{\scriptsize$\pm$2.52} & 81.00{\scriptsize$\pm$1.67} & 87.21{\scriptsize$\pm$1.27} & 69.08{\scriptsize$\pm$2.08} & 77.07{\scriptsize$\pm$0.82} & 87.47{\scriptsize$\pm$4.55} & 71.82{\scriptsize$\pm$4.01} & 74.68{\scriptsize$\pm$2.72} \\
192 & 92.41{\scriptsize$\pm$0.40} & 84.30{\scriptsize$\pm$0.33} & 90.06{\scriptsize$\pm$0.07} & 73.40{\scriptsize$\pm$0.20} & 78.91{\scriptsize$\pm$0.14} & 92.99{\scriptsize$\pm$0.34} & 78.28{\scriptsize$\pm$0.02} & 79.77{\scriptsize$\pm$0.24} \\
\bottomrule
\end{tabular}
\end{table}

This shared exponential historical state provides two closely related benefits. MTHR can reuse block-start graph aggregation to reduce repeated computation, while SNAM can directly access exact activity counts from the same historical computation framework. The two modules share historical statistics and computational structure without requiring their trainable hidden representations to be shared.

\paragraph{From multi-timescale activity counts to conditional event intensity.}

The next question is why these activity counts can be used to estimate the conditional event intensity $\lambda_u(t)$ introduced in the previous section.

The count $n_u^k(t)$ is not itself an event rate. It records the level of historical activity at a particular timescale, but it does not directly specify how many interactions are expected to occur per unit time in the future. To turn these historical statistics into a conditional intensity, we must specify how past activity influences the rate of future events.

We start from a simple self-excitation assumption: an interaction that has already occurred may temporarily increase the rate at which the same source node initiates another interaction, and this influence gradually weakens over time. Different activity patterns may persist for different lengths of time, so we do not want to restrict the influence of historical interactions to a single timescale.

MTHR already provides $K$ fixed exponential timescales, with the corresponding activity counts
\[
n_u^1(t),\ldots,n_u^K(t).
\]

We therefore model the source conditional intensity as
\begin{equation}
\boxed{
\lambda_u(t)
=
\mu+\sum_{k=1}^{K}w_kn_u^k(t),
\qquad
\mu>0,\quad w_k>0.
}
\tag{A.15}
\label{eq:app-a15}
\end{equation}

Here, $\mu$ is the baseline event rate. It ensures that the model retains a positive intensity even when the node has little recent activity. The coefficient $w_k$ determines how strongly historical activity at the $k$-th timescale contributes to the current event rate.

\Eqref{eq:app-a15} has a clear temporal interpretation. Suppose source node $u$ initiates an interaction at time $t_e$. At a later query time $t>t_e$, this event contributes
\[
e^{-\omega_k(t-t_e)}
\]
to its activity count at the $k$-th timescale, and therefore contributes
\[
w_ke^{-\omega_k(t-t_e)}
\]
to its current conditional intensity.

Summing over all timescales, the total influence of this historical event on the subsequent activity rate is
\[
\varphi(t-t_e)
=
\sum_{k=1}^{K}
w_ke^{-\omega_k(t-t_e)}.
\]

Consequently,~\eqref{eq:app-a15} can also be written as
\begin{equation}
\boxed{
\lambda_u(t)
=
\mu+
\sum_{\substack{e\ni u\\t_e<t}}
\varphi(t-t_e),
\qquad
\varphi(s)=\sum_{k=1}^{K}w_ke^{-\omega_ks}.
}
\tag{A.16}
\label{eq:app-a16}
\end{equation}

This expression shows that SNAM can be viewed as a linear self-exciting intensity model with a multi-exponential excitation kernel. Each historical interaction makes a positive contribution to the subsequent event rate, and that contribution decays over time. The different timescales jointly determine the short- and long-term structure of this influence.

It also shows that the multi-timescale structure of SNAM is not an additional, unrelated design choice. MTHR already requires multiple timescales to represent historical relationships over different temporal ranges. SNAM uses the same timescales to assign an event-rate interpretation to historical activity. Short-timescale counts can capture recent bursts of interactions, while longer-timescale counts retain more persistent activity patterns. By learning the contribution of each timescale, the model determines how strongly these different patterns affect the future event rate.

It is important to distinguish the roles of the decay rates and the activity coefficients. The decay rates $\omega_k$ are determined from the temporal range of the training data and remain fixed, whereas the coefficients $w_k$ are learned from data. The former determine how long the influence of a historical interaction persists; the latter determine how strong that influence is. Fixing the timescales therefore does not mean that the model prespecifies the relative importance of different activity patterns.

\paragraph{A small set of shared parameters and their mathematical interpretation.}

To ensure that the conditional intensity in~\eqref{eq:app-a15} remains positive, while keeping the parameters of different timescales dimensionally consistent, we use the following parameterization:
\begin{equation}
\boxed{
w_k=\omega_ke^{\theta_k},
\qquad
\mu=\omega_Ke^{\theta_\mu}.
}
\tag{A.17}
\label{eq:app-a17}
\end{equation}

Here, $\theta_\mu,\theta_1,\ldots,\theta_K$ are trainable real-valued parameters, and $\omega_K$ is the decay rate of the slowest timescale. The exponential parameterization guarantees $w_k>0$ and $\mu>0$. It also ensures that both quantities have units of inverse time, consistent with the decay rates $\omega_k$.

This parameterization has an additional useful interpretation. Consider the contribution of a single historical interaction to the future intensity at the $k$-th timescale. Its contribution as a function of the elapsed time $s\geq0$ is $w_ke^{-\omega_ks}$. Integrating this curve over the entire future time axis gives
\begin{equation}
\boxed{
\int_0^\infty
w_ke^{-\omega_ks}\,\mathrm ds
=
\frac{w_k}{\omega_k}
=
e^{\theta_k}.
}
\tag{A.18}
\label{eq:app-a18}
\end{equation}

Thus, $\theta_k$ is not merely an uninterpretable mixture weight. The quantity $e^{\theta_k}$ represents the total direct excitation contributed by one event at the $k$-th timescale, and is dimensionless. Under the usual interpretation of a linear self-exciting process, it can also correspond to the expected number of direct offspring events at that timescale, rather than the total number of events resulting from all subsequent generations of excitation.

This interpretation further separates the roles of the two parameters. The decay rate $\omega_k$ determines how quickly the excitation disappears, while $e^{\theta_k}$ determines its total direct contribution at that timescale. The model can therefore learn how strongly historical activity with different durations affects future event rates, while keeping the multi-timescale temporal grid fixed.

More importantly, all source nodes share the same parameter vector
\[
\vartheta=(\theta_\mu,\theta_1,\ldots,\theta_K).
\]

The model does not need to learn a separate set of intensity parameters for each source node. Different source nodes can have different values of $\lambda_u(t)$ at the same time because they have experienced different interaction histories and therefore have different activity counts $n_u^k(t)$, not because the model memorizes a node-specific activity rate.

As a result, SNAM contains only $K+1$ trainable parameters, independent of the number of nodes. Its modeling capacity comes from the multi-timescale activity information preserved in the historical states, rather than from a large number of node-specific parameters.

Of course,~\eqref{eq:app-a15} remains an explicit modeling assumption, rather than the unique conditional intensity implied by the historical aggregation operator. Its positive linear combination describes the self-exciting effect through which past interactions may increase subsequent event rates. It cannot directly represent an interaction that suppresses future activity, nor does it automatically guarantee long-term stationarity. We adopt this form because it provides an interpretable historical excitation structure and is fully compatible with the exponential states already used by MTHR.

\paragraph{A shared historical foundation with independent model parameters.}

We can now describe the relationship between MTHR and SNAM more precisely.

Both modules use the same timescales $\omega_1,\ldots,\omega_K$ and extract information from the same strictly preceding interaction history. However, they use this information differently. MTHR aggregates interaction partners and their neighborhood information to learn node representations for relation prediction. SNAM turns the activity counts produced by the same historical aggregation operator into a conditional event intensity.

The sharing takes place at the level of historical statistics, rather than by building the activity estimator on top of MTHR's trainable node representations. Given the interaction history and fixed decay rates, the counts $n_u^k(t)$ are already determined. They do not depend on the base node embedding table, the learned projections used in graph aggregation, or the parameters of the CRM relation scoring network. Therefore, the conditional intensity in~\eqref{eq:app-a15} depends only on the activity counts and its own $K+1$ parameters.

This property naturally connects to the statistical motivation established in the previous section. Since conditional destination preference and source activity are two different factors in the ideal probabilistic model, there is no need to force them to be produced by the same trainable representation. Instead, sharing historical statistics avoids redundant computation, while separate parameter sets allow the two branches to receive supervision that matches their respective predictive objectives.

For SNAM, this supervision does not require constructing any negative node pairs. The times at which real interactions occur provide evidence that the activity rate should be high, while intervals with no interactions also provide information that the predicted rate should not remain excessively high. In the next section, we derive the point-process likelihood of SNAM from these two sources of information. We further show that, because its intensity shares the same exponential kernels as the historical states, the time integral in its training objective can also be computed exactly.

This allows us to learn source activity independently of the internal parameters of the relation branch, and provides the foundation for reusing the same activity model with other temporal link prediction backbones.

\subsection{Independent Learning and Closed-Form Likelihood of Source Activity}
\label{app:activity_likelihood}

\paragraph{Deriving the activity likelihood from observed event times.}

The previous section established a conditional event intensity $\lambda_u(t)$ for each source node. We now address how to learn its parameters from observed interactions. Unlike the relation branch, which constructs a classification task using artificial negative samples, SNAM directly uses the times at which real interactions occur. The model must explain both why interactions occur at their observed times and why no interactions are observed during the remaining periods.

These two sources of information can be captured by a single point-process likelihood. We derive its form directly from the definition of conditional intensity.

Consider an observation window $[a,b)$, and let $\mathcal U_{\mathrm{train}}$ denote the set of training source nodes. Given the history preceding the window, suppose we observe $m$ interactions, with source nodes and event times
\[
(u_1,t_1),\ldots,(u_m,t_m),
\qquad
a<t_1<\cdots<t_m<b.
\]

We first consider the ideal continuous-time setting in which event times are distinct. The treatment of identical timestamps in real datasets is discussed later when describing the implementation boundaries.

Since SNAM models only the source nodes and occurrence times of interactions, we do not need to specify their destination nodes here. Define the total event intensity of all training source nodes as
\[
\Lambda_{\mathrm{train}}(t)
=
\sum_{u\in\mathcal U_{\mathrm{train}}}
\lambda_u(t).
\]

By the definition of conditional intensity, for a sufficiently short interval of length $\delta>0$, the conditional probability that no training source node initiates an interaction is
\[
1-\Lambda_{\mathrm{train}}(t)\delta+o(\delta).
\]

If exactly one interaction occurs in this interval and its source node is $u$, the corresponding conditional probability is
\[
\lambda_u(t)\delta+o(\delta).
\]

Our goal is to use these two local probabilities to obtain the likelihood of the entire observed event sequence.

First, divide the observation window into short intervals of length $\delta$. When $\delta$ is sufficiently small, we can distinguish intervals containing observed events from those containing no events. Let $\mathcal Z_\delta$ denote the set of intervals without events, and let $r_j$ be the starting time of the $j$-th interval.

By considering the intervals in chronological order and applying the chain rule of conditional probability, we can write the probability of observing this event pattern as
\[
\begin{aligned}
P_\delta
={}&
\underbrace{
\prod_{i=1}^{m}
\left[
\lambda_{u_i}(t_i^-)\delta+o(\delta)
\right]
}_{\text{Intervals containing observed events}}
\\[6pt]
&\times
\underbrace{
\prod_{j\in\mathcal Z_\delta}
\left[
1-\Lambda_{\mathrm{train}}(r_j)\delta
+o(\delta)
\right]
}_{\text{Intervals without events}}.
\end{aligned}
\]

Here, $\lambda_{u_i}(t_i^-)$ is computed using only the history strictly preceding the current event. We do not assume that different time intervals are independent. Instead, the conditional probability of each interval depends on the history observed up to that point, so earlier events may change the intensity in subsequent intervals.

We now consider the two products separately.

For intervals without events, taking the logarithm turns their product into a sum:
\[
\log P_{\mathrm{no\ event},\delta}
=
\sum_{j\in\mathcal Z_\delta}
\log\left(
1-\Lambda_{\mathrm{train}}(r_j)\delta
+o(\delta)
\right).
\]

Using $\log(1-x)=-x+o(x)$, as $\delta\to0$, we obtain
\[
\begin{aligned}
\log P_{\mathrm{no\ event},\delta}
&=
-\sum_{j\in\mathcal Z_\delta}
\Lambda_{\mathrm{train}}(r_j)\delta
+
\sum_{j\in\mathcal Z_\delta}o(\delta)
\\
&\longrightarrow
-\int_a^b
\Lambda_{\mathrm{train}}(t)\,\mathrm dt.
\end{aligned}
\]

This limit holds under the usual regularity conditions, including local boundedness and sufficiently uniform small-interval approximations. Since the observation window contains only finitely many events, excluding the intervals containing those events does not change the limit of the Riemann sum. Therefore, the product of the no-event probabilities converges to
\[
P_{\mathrm{no\ event},\delta}
\longrightarrow
\exp\left(
-\int_a^b
\Lambda_{\mathrm{train}}(t)\,\mathrm dt
\right).
\]

Next, consider the intervals containing observed events. The interval containing the $i$-th event contributes a probability of approximately $\lambda_{u_i}(t_i^-)\delta$. The product over all $m$ event intervals therefore satisfies
\[
\begin{aligned}
P_{\mathrm{event},\delta}
&=
\prod_{i=1}^{m}
\left[
\lambda_{u_i}(t_i^-)\delta+o(\delta)
\right]
\\
&=
\delta^m
\left[
\prod_{i=1}^{m}
\lambda_{u_i}(t_i^-)+o(1)
\right].
\end{aligned}
\]

Combining the contributions from event intervals and no-event intervals gives
\[
\begin{aligned}
P_\delta
={}&
\delta^m
\left[
\prod_{i=1}^{m}\lambda_{u_i}(t_i^-)+o(1)
\right]
\\
&\times
\exp\left(
-\int_a^b\Lambda_{\mathrm{train}}(t)\,\mathrm dt
+o(1)
\right).
\end{aligned}
\]

Importantly, $P_\delta$ is still the probability that the events fall into their specified short intervals. What we need instead is the joint density of the continuous event times $(t_1,\ldots,t_m)$. Each event time belongs to an interval of width $\delta$, so these intervals form a region of volume $\delta^m$ in the $m$-dimensional event-time space. Dividing the probability above by $\delta^m$ and taking $\delta\to0$ therefore gives the joint density of the event times.

Thus, conditioned on the history preceding the observation window, the likelihood of the observed source nodes and event times is
\begin{equation}
\boxed{
\begin{aligned}
\mathcal P_{\mathrm{act}}
&=
\lim_{\delta\to0}\frac{P_\delta}{\delta^m}
\\
&=
\left(
\prod_{i=1}^{m}
\lambda_{u_i}(t_i^-)
\right)
\exp\left(
-\int_a^b\Lambda_{\mathrm{train}}(t)\,\mathrm dt
\right)
\\
&=
\left(
\prod_{i=1}^{m}
\lambda_{u_i}(t_i^-)
\right)
\exp\left(
-\sum_{u\in\mathcal U_{\mathrm{train}}}
\int_a^b\lambda_u(t)\,\mathrm dt
\right).
\end{aligned}
}
\tag{A.19}
\label{eq:app-a19}
\end{equation}

Finally, taking the negative logarithm of~\eqref{eq:app-a19} gives the negative log-likelihood of SNAM over the observation window:
\begin{equation}
\boxed{
\mathcal L_{\mathrm{act}}^{\mathrm{raw}}
=
\sum_{u\in\mathcal U_{\mathrm{train}}}
\int_a^b\lambda_u(t)\,\mathrm dt
-
\sum_{i=1}^{m}
\log\lambda_{u_i}(t_i^-).
}
\tag{A.20}
\label{eq:app-a20}
\end{equation}

This objective separates the supervision provided by the entire observation window into two terms. The observed-event term encourages the model to assign high source intensities at the times when real interactions occur. The cumulative-intensity term prevents the model from maintaining excessively high predicted event rates throughout the observation window. In particular, even if a training source node does not initiate any interaction during the current window, it is still constrained by the first term.

\Eqref{eq:app-a20} therefore allows us to learn source activity directly from real events and their occurrence times, without constructing additional negative node pairs.

\paragraph{Decomposing the full likelihood into source activity and conditional destination preference.}

\Eqref{eq:app-a20} also reveals a deeper connection between SNAM and relation modeling.

In Appendix~\ref{app:activity_motivation}, we established that the interaction intensity of a complete node pair can be decomposed as
\[
\lambda_{uv}(t)
=
\lambda_u(t)\pi_{uv}(t),
\]
where $\lambda_u(t)$ describes the source event rate and $\pi_{uv}(t)$ describes the conditional destination choice given that an interaction occurs.

Now consider an ideal marked point process in which every event has not only a source node and occurrence time, but also a destination node. If we directly construct the likelihood of the complete interaction sequence using the pairwise intensity $\lambda_{uv}(t)$, the log-intensity of each observed event can be decomposed as
\[
\log\lambda_{u_iv_i}(t_i^-)
=
\log\lambda_{u_i}(t_i^-)
+
\log\pi_{u_iv_i}(t_i).
\]

At the same time, since every source event has a destination, the total intensity over all possible node pairs satisfies
\[
\sum_{u,v}\lambda_{uv}(t)
=
\sum_u\lambda_u(t).
\]

Consequently, the log-likelihood of the complete event sequence can also be decomposed as
\begin{equation}
\boxed{
\begin{aligned}
\ell_{\mathrm{full}}
={}&
\underbrace{
\sum_{i=1}^{m}\log\lambda_{u_i}(t_i^-)
-
\sum_u\int_a^b\lambda_u(t)\,\mathrm dt
}_{\text{Source-activity log-likelihood}}
\\[4pt]
&+
\underbrace{
\sum_{i=1}^{m}\log\pi_{u_iv_i}(t_i)
}_{\text{Conditional destination-choice log-likelihood}}.
\end{aligned}
}
\tag{A.21}
\label{eq:app-a21}
\end{equation}

This decomposition shows that source activity and conditional destination preference have distinct roles not only in determining the intensity of a complete interaction, but also in the probabilistic modeling of an entire event sequence.

Our method does not directly maximize the full joint likelihood in~\eqref{eq:app-a21}. The relation branch uses binary cross-entropy with random destination negative sampling to learn a score that distinguishes real destinations from artificially sampled ones. SNAM, on the other hand, uses the point-process negative log-likelihood in~\eqref{eq:app-a20} to estimate the absolute event rate of each source node.

We therefore use two complementary but different training objectives. The former learns relational discrimination by contrasting real and sampled node pairs, while the latter estimates source activity from observed event times and cumulative intensity over the observation window. \Eqref{eq:app-a21} explains the probabilistic structure behind this division of labor, but does not imply that random-negative binary cross-entropy is exactly equivalent to the conditional destination-choice likelihood.
\begin{table}[t]
\caption{Form of the source-activity term, remaining protocols: test AP (\%), mean{\scriptsize$\pm$std} over three seeds, for the score $f+\beta\,b$ with the term $b$ replaced on the same checkpoints ($\beta$ chosen on validation; under random negatives every variant selects $\beta=0$). $n_u^k$ is the source's decayed event count at scale $k$ ($K$ the slowest); $\Delta t_u$ is the time elapsed since the source's previous event, so $-\log(1+\Delta t_u)$ is a recency score; the MLP is a two-layer network on $\log(1+n_u)$ fitted on the training period. In the inductive setting the historical and inductive samplers draw from the same new-node subset, so only the historical protocol is shown. $\star$: the main results. Bold: best within a block.}
\label{tab:a3_appendix}
\centering
\footnotesize
\setlength{\tabcolsep}{3pt}
\renewcommand{\arraystretch}{0.92}
\begin{tabular}{lccccc}
\toprule
term $b$ & Wikipedia & Reddit & UCI & Enron & LastFM \\
\midrule
\multicolumn{6}{l}{\textit{Transductive setting, historical negatives}} \\
\midrule
none ($f$ alone, $\beta=0$) & 88.71{\scriptsize$\pm$0.80} & 82.74{\scriptsize$\pm$1.44} & 91.47{\scriptsize$\pm$0.52} & 84.04{\scriptsize$\pm$2.43} & 84.60{\scriptsize$\pm$1.20} \\
$\log\lambda_u$ (SNAM)$^\star$ & 93.25{\scriptsize$\pm$0.05} & 90.16{\scriptsize$\pm$0.17} & 96.04{\scriptsize$\pm$0.17} & 90.26{\scriptsize$\pm$0.77} & 95.36{\scriptsize$\pm$0.03} \\
\cmidrule(lr){1-6}
$\log(1+n_u^{K})$ & 89.03{\scriptsize$\pm$0.54} & 85.25{\scriptsize$\pm$0.61} & 91.47{\scriptsize$\pm$0.52} & 84.03{\scriptsize$\pm$2.43} & 87.15{\scriptsize$\pm$0.54} \\
$\log(1+n_u^{K/2})$ & 91.67{\scriptsize$\pm$0.19} & 83.92{\scriptsize$\pm$1.23} & 95.03{\scriptsize$\pm$0.34} & 88.98{\scriptsize$\pm$1.56} & 92.71{\scriptsize$\pm$0.32} \\
$-\log(1+\Delta t_u)$ & 93.33{\scriptsize$\pm$0.04} & 90.29{\scriptsize$\pm$0.21} & 96.02{\scriptsize$\pm$0.08} & 88.96{\scriptsize$\pm$0.89} & 95.30{\scriptsize$\pm$0.03} \\
MLP on $\log(1+n_u)$ & \textbf{93.42}{\scriptsize$\pm$0.08} & \textbf{90.51}{\scriptsize$\pm$0.15} & \textbf{96.22}{\scriptsize$\pm$0.08} & \textbf{90.55}{\scriptsize$\pm$0.88} & \textbf{95.61}{\scriptsize$\pm$0.02} \\
\midrule
\multicolumn{6}{l}{\textit{Inductive setting, historical negatives}} \\
\midrule
none ($f$ alone, $\beta=0$) & 77.78{\scriptsize$\pm$1.86} & 67.54{\scriptsize$\pm$2.74} & 84.08{\scriptsize$\pm$0.35} & 78.56{\scriptsize$\pm$2.85} & 78.38{\scriptsize$\pm$0.70} \\
$\log\lambda_u$ (SNAM)$^\star$ & \textbf{88.29}{\scriptsize$\pm$0.13} & 79.20{\scriptsize$\pm$0.34} & 89.03{\scriptsize$\pm$0.05} & \textbf{85.57}{\scriptsize$\pm$0.91} & 90.12{\scriptsize$\pm$0.61} \\
\cmidrule(lr){1-6}
$\log(1+n_u^{K})$ & 78.47{\scriptsize$\pm$1.67} & 71.19{\scriptsize$\pm$1.46} & 84.08{\scriptsize$\pm$0.35} & 78.96{\scriptsize$\pm$2.76} & 79.72{\scriptsize$\pm$0.23} \\
$\log(1+n_u^{K/2})$ & 84.85{\scriptsize$\pm$0.76} & 69.50{\scriptsize$\pm$2.27} & 87.20{\scriptsize$\pm$0.52} & 83.88{\scriptsize$\pm$1.54} & 86.70{\scriptsize$\pm$0.67} \\
$-\log(1+\Delta t_u)$ & 88.15{\scriptsize$\pm$0.08} & 79.37{\scriptsize$\pm$0.36} & \textbf{89.51}{\scriptsize$\pm$0.03} & 83.83{\scriptsize$\pm$0.57} & \textbf{90.50}{\scriptsize$\pm$0.59} \\
MLP on $\log(1+n_u)$ & 88.10{\scriptsize$\pm$0.15} & \textbf{79.65}{\scriptsize$\pm$0.31} & 89.26{\scriptsize$\pm$0.08} & 85.32{\scriptsize$\pm$1.05} & 90.41{\scriptsize$\pm$0.62} \\
\bottomrule
\end{tabular}
\end{table}

\paragraph{Closed-form temporal integration from shared exponential states.}

Although the point-process likelihood does not require artificial negative samples, it contains a temporal integral:
\[
\int_a^b\lambda_u(t)\,\mathrm dt.
\]

For a general neural intensity model, this integral may not have an analytic form and may require temporal discretization, numerical quadrature, or Monte Carlo approximation. If the intensity must be evaluated at many additional time points during training, the activity branch itself can introduce substantial computational overhead.

The exponential structure of SNAM makes this integral directly tractable. Recall from~\eqref{eq:app-a15} that
\[
\lambda_u(t)
=
\mu+\sum_{k=1}^{K}w_kn_u^k(t).
\]

Because each activity count is a sum of exponential terms, it is sufficient to first integrate one exponential kernel and then reuse the resulting expression for the complete history.

Let
\[
T=b-a
\]
denote the length of the observation window, and define
\begin{equation}
\boxed{
G_k(T)
=
\int_0^T e^{-\omega_ks}\,\mathrm ds
=
\frac{1-e^{-\omega_kT}}{\omega_k}.
}
\tag{A.22}
\label{eq:app-a22}
\end{equation}

The function $G_k(T)$ is the total contribution accumulated by a unit exponential decay over a time interval of length $T$.

To compute the integral of $n_u^k(t)$ over $[a,b)$, we divide the historical interactions into two groups.

First consider interactions that occurred before the observation window began. At time $a$, their total contribution at the $k$-th timescale has already been summarized by $n_u^k(a)$. As time advances from $a$ to $t$, this accumulated state decays as
\[
e^{-\omega_k(t-a)}n_u^k(a).
\]

Its contribution over the full observation window is therefore
\[
\begin{aligned}
\int_a^b
e^{-\omega_k(t-a)}n_u^k(a)\,\mathrm dt
&=
n_u^k(a)
\int_a^b
e^{-\omega_k(t-a)}\,\mathrm dt
\\
&=
n_u^k(a)G_k(T).
\end{aligned}
\]

Next consider an interaction occurring inside the observation window at time $t_e\in[a,b)$. It contributes nothing before $t_e$. From $t_e$ onward, its contribution to the activity count is
\[
e^{-\omega_k(t-t_e)}.
\]

Its accumulated contribution over the remainder of the observation window is therefore
\[
\begin{aligned}
\int_{t_e}^{b}
e^{-\omega_k(t-t_e)}\,\mathrm dt
=
G_k(b-t_e).
\end{aligned}
\]

Adding the contribution from the pre-window history and all interactions observed inside the window gives
\begin{equation}
\boxed{
\begin{aligned}
\int_a^b n_u^k(t)\,\mathrm dt
={}&
n_u^k(a)G_k(T)
\\
&+
\sum_{\substack{
e\ni u\\
a\leq t_e<b
}}
G_k(b-t_e).
\end{aligned}
}
\tag{A.23}
\label{eq:app-a23}
\end{equation}

Substituting~\eqref{eq:app-a23} into the SNAM intensity and summing over all timescales yields
\begin{equation}
\boxed{
\begin{aligned}
\int_a^b\lambda_u(t)\,\mathrm dt
={}&
\mu T
\\
&+
\sum_{k=1}^{K}w_k
\left[
n_u^k(a)G_k(T)
+
\sum_{\substack{
e\ni u\\
a\leq t_e<b
}}
G_k(b-t_e)
\right].
\end{aligned}
}
\tag{A.24}
\label{eq:app-a24}
\end{equation}

\Eqref{eq:app-a24} gives a closed-form expression for the cumulative source intensity over the observation window. It requires only the multi-timescale activity counts at the beginning of the window, the timestamps of observed interactions within the window, and the learned SNAM parameters. No additional temporal sampling or numerical integration is required.

This result also connects naturally to the interpretation of the SNAM parameters in~\eqref{eq:app-a18}. A single event contributes a total direct excitation of $e^{\theta_k}$ over the entire future time axis at the $k$-th timescale. If we observe only a finite interval of length $T$ after that event, its accumulated contribution is
\[
w_kG_k(T)
=
e^{\theta_k}
\left(
1-e^{-\omega_kT}
\right).
\]

Thus, the closed-form integral can be viewed as accumulating the contribution of every historical interaction over the portion of the observation window in which it remains active. Events that occurred before the window are already compressed into the entry state $n_u^k(a)$, while events occurring inside the window contribute only from their own occurrence times onward.

This is another consequence of using the same exponential states for MTHR and SNAM. The multiplicative structure of the exponential kernel allows historical states to be propagated exactly from a block boundary to a query time, as shown in~\eqref{eq:app-a14}. The same kernel also admits the exact temporal integral in~\eqref{eq:app-a24}. The two properties arise from the same mathematical structure: one supports efficient state reuse, while the other makes point-process likelihood training inexpensive.

\paragraph{Block-wise optimization and independent parameter learning.}

In practice, we optimize SNAM using the same temporal block structure as the relation branch.

Let $c$ denote a training block, let $a_c$ and $b_c$ denote the boundaries of its associated observation window, and let $B_c$ be the number of positive interactions supervised in the block. For non-final training blocks, $b_c$ is typically the start time of the next block.

Using the point-process negative log-likelihood derived above, we define the activity loss for block $c$ as
\begin{equation}
\boxed{
\mathcal L_{\mathrm{act},c}
=
\frac{1}{B_c}
\left[
\sum_{u\in\mathcal U_{\mathrm{train}}}
\int_{a_c}^{b_c}
\lambda_u(t)\,\mathrm dt
-
\sum_{i\in c}
\log\lambda_{u_i}(t_i^-)
\right].
}
\tag{A.25}
\label{eq:app-a25}
\end{equation}

The integral term is evaluated using the closed-form expression in~\eqref{eq:app-a24}, while the event term uses the activity counts computed from the history strictly preceding each observed interaction.

The integral in~\eqref{eq:app-a25} covers all training source nodes, not only those that appear in the current block. This is important because a source node that remains inactive throughout the observation window still provides information: assigning a persistently high intensity to such a node should be penalized even though it contributes no event term.

The total training objective for block $c$ is
\begin{equation}
\boxed{
\mathcal L_c(\Theta,\vartheta)
=
\mathcal L_{\mathrm{link},c}(\Theta)
+
\mathcal L_{\mathrm{act},c}(\vartheta),
\qquad
\vartheta=
(\theta_\mu,\theta_1,\ldots,\theta_K).
}
\tag{A.26}
\label{eq:app-a26}
\end{equation}

Here, $\Theta$ denotes the trainable parameters of MTHR, CRM, and the relation scoring network, while $\vartheta$ contains the $K+1$ trainable parameters of SNAM.

Although both losses are evaluated within the same training block and optimized in the same training loop, they act on disjoint parameter sets. Once the interaction history and fixed decay rates are given, the activity counts $n_u^k(t)$ contain no trainable parameters. Consequently,
\[
\mathcal L_{\mathrm{act},c}
\]
updates only $\vartheta$ and does not require backpropagation through MTHR's graph aggregation or CRM's relation scoring network. Conversely, the random-negative relation loss does not train the SNAM intensity parameters.

This separation preserves the statistical meaning of the two objectives. The relation branch learns to distinguish observed node pairs from randomly sampled alternatives, while SNAM learns source event rates from observed event times and cumulative intensity over the entire observation window.

Our implementation performs this optimization block by block. The first training block is used only to establish the initial historical state, and supervised optimization begins from the following block. Each block loss is normalized by its own number of supervised interactions $B_c$, followed by one optimizer update. The resulting training procedure is therefore a block-wise likelihood optimization scheme and should not be interpreted as exactly the same objective as maximizing one unnormalized point-process likelihood over the entire training horizon in a single step.

Real datasets may also contain discretized timestamps, and multiple records may share the same timestamp. Equal-event-count partitioning can place such records near or across block boundaries. In all event terms, we consistently use only history with timestamps strictly smaller than the current query time. The exactness claimed in~\eqref{eq:app-a24} refers to the temporal integral of the exponential intensity defined by our model over a specified observation window. It does not imply that a simple continuous-time point-process observation model exactly describes every artifact introduced by discretized timestamps or tied events.

\paragraph{Why independent learning enables cross-model reuse.}

The separation above has an additional consequence: SNAM learns source activity from the event stream itself rather than from the hidden representation of a particular relation encoder.

Given the same training interaction history, time units, decay-rate grid, and strict-history convention, the activity counts
\[
n_u^1(t),\ldots,n_u^K(t)
\]
are fully determined independently of MTHR and CRM. The SNAM objective in~\eqref{eq:app-a25} depends only on these counts, the observed event times, and its own parameter vector $\vartheta$.

Thus, after learning
\[
\hat{\vartheta}
=
(\hat{\theta}_\mu,\hat{\theta}_1,\ldots,\hat{\theta}_K)
\]
on a given dataset, the resulting source-intensity model can be evaluated without using the internal representations of our relation branch.

Suppose an independently trained temporal link prediction model produces a base score
\[
f_{\mathrm{base}}(u,v,t).
\]
As long as the same observable event history is available, we can independently compute the source activity counts and obtain
\[
\lambda_u(t;\hat{\vartheta}).
\]

The final score can then be formed as
\[
s_\beta(u,v,t)
=
f_{\mathrm{base}}(u,v,t)
+
\beta\log\lambda_u(t;\hat{\vartheta}).
\]

The external model does not need to adopt MTHR's historical encoder or explicitly maintain our multi-timescale counts internally. These statistics can be computed by a separate SNAM-side history index. The parameters of the external model remain unchanged, and the already learned SNAM parameters $\hat{\vartheta}$ can be reused on the same dataset without model-specific retraining. Only the score-combination coefficient $\beta$ is selected using the validation split for the corresponding model and evaluation protocol.

This reuse requires consistency in the quantities that define the activity process: the visible interaction history, time units, decay-rate configuration, and strict-history rule must agree with those used when SNAM was trained. The claim is therefore that one dataset-level activity model can be reused across different relation encoders under consistent historical statistics, rather than that a single set of intensity parameters can be transferred unchanged across arbitrary datasets.

\begin{table}[t]
\caption{Transfer of the source-activity term, remaining protocols: what each backbone gains over itself. Test AP (\%) of the backbone alone and with $\beta\log\lambda_u$ added to its logits, and the difference $\Delta$, in the transductive setting under inductive (ind) negatives and in the inductive setting under historical (hist) negatives; the transductive setting under historical negatives is in Table~\ref{tab:transfer}. Mean{\scriptsize$\pm$std} over three seeds (DyGFormer$^\dagger$ on LastFM: one), the std of $\Delta$ over the per-seed differences; the mean row averages the five streams. $\lambda_u$ keeps the parameters fitted for our model on the same stream and $\beta$ is chosen on the backbone's validation split. Under random negatives the validation split selects $\beta=0$ for every backbone and seed, so the backbone's score is unchanged there and those columns are omitted. In the inductive setting the two samplers draw from the same new-node subset, so only the historical protocol is shown. $\dagger$: retrained by us with the corrected released code.}
\label{tab:transfer_appendix}
\centering
\scriptsize
\setlength{\tabcolsep}{4pt}
\renewcommand{\arraystretch}{0.95}
\begin{tabular}{lcccccc}
\toprule
 & \multicolumn{3}{c}{Transductive, ind} & \multicolumn{3}{c}{Inductive, hist} \\
\cmidrule(lr){2-4} \cmidrule(lr){5-7}
backbone / stream & alone & $+\lambda$ & $\Delta$ & alone & $+\lambda$ & $\Delta$ \\
\midrule
\multicolumn{7}{l}{\textit{TPNet}} \\
Wikipedia & 82.13{\tiny$\pm$3.35} & 91.18{\tiny$\pm$0.26} & +9.05{\tiny$\pm$3.10} & 74.81{\tiny$\pm$1.88} & 88.83{\tiny$\pm$0.06} & +14.02{\tiny$\pm$1.83} \\
Reddit & 87.92{\tiny$\pm$0.54} & 90.10{\tiny$\pm$0.36} & +2.19{\tiny$\pm$0.18} & 61.70{\tiny$\pm$0.32} & 80.27{\tiny$\pm$0.25} & +18.57{\tiny$\pm$0.08} \\
UCI & 77.42{\tiny$\pm$2.63} & 87.40{\tiny$\pm$0.58} & +9.98{\tiny$\pm$2.54} & 78.65{\tiny$\pm$1.83} & 88.61{\tiny$\pm$0.30} & +9.96{\tiny$\pm$1.75} \\
Enron & 76.75{\tiny$\pm$0.44} & 87.28{\tiny$\pm$0.19} & +10.53{\tiny$\pm$0.25} & 74.96{\tiny$\pm$0.40} & 86.38{\tiny$\pm$0.21} & +11.42{\tiny$\pm$0.61} \\
LastFM & 76.63{\tiny$\pm$1.29} & 91.71{\tiny$\pm$0.13} & +15.08{\tiny$\pm$1.20} & 80.76{\tiny$\pm$1.22} & 92.21{\tiny$\pm$0.05} & +11.46{\tiny$\pm$1.22} \\
\textit{mean} & 80.17 & 89.53 & +9.37 & 74.18 & 87.26 & +13.08 \\
\midrule
\multicolumn{7}{l}{\textit{TGN$^\dagger$}} \\
Wikipedia & 85.17{\tiny$\pm$0.98} & 91.33{\tiny$\pm$0.11} & +6.16{\tiny$\pm$0.88} & 81.86{\tiny$\pm$1.01} & 89.03{\tiny$\pm$0.02} & +7.17{\tiny$\pm$1.00} \\
Reddit & 87.66{\tiny$\pm$0.40} & 89.39{\tiny$\pm$0.28} & +1.72{\tiny$\pm$0.12} & 63.95{\tiny$\pm$0.22} & 79.37{\tiny$\pm$0.17} & +15.42{\tiny$\pm$0.15} \\
UCI & 70.75{\tiny$\pm$0.29} & 84.65{\tiny$\pm$0.10} & +13.91{\tiny$\pm$0.29} & 70.67{\tiny$\pm$0.36} & 85.78{\tiny$\pm$0.13} & +15.12{\tiny$\pm$0.25} \\
Enron & 72.80{\tiny$\pm$0.33} & 85.73{\tiny$\pm$0.73} & +12.93{\tiny$\pm$0.57} & 62.50{\tiny$\pm$0.56} & 82.93{\tiny$\pm$0.18} & +20.43{\tiny$\pm$0.68} \\
LastFM & 69.84{\tiny$\pm$4.59} & 90.21{\tiny$\pm$0.25} & +20.37{\tiny$\pm$4.35} & 75.05{\tiny$\pm$2.11} & 90.50{\tiny$\pm$0.10} & +15.46{\tiny$\pm$2.03} \\
\textit{mean} & 77.25 & 88.26 & +11.02 & 70.80 & 85.52 & +14.72 \\
\midrule
\multicolumn{7}{l}{\textit{DyGFormer$^\dagger$}} \\
Wikipedia & 69.34{\tiny$\pm$0.32} & 90.48{\tiny$\pm$0.06} & +21.14{\tiny$\pm$0.35} & 63.40{\tiny$\pm$1.23} & 88.51{\tiny$\pm$0.04} & +25.11{\tiny$\pm$1.25} \\
Reddit & 91.51{\tiny$\pm$0.51} & 92.63{\tiny$\pm$0.40} & +1.11{\tiny$\pm$0.14} & 64.83{\tiny$\pm$2.14} & 80.25{\tiny$\pm$0.24} & +15.42{\tiny$\pm$2.38} \\
UCI & 72.05{\tiny$\pm$0.62} & 85.73{\tiny$\pm$0.11} & +13.68{\tiny$\pm$0.52} & 71.74{\tiny$\pm$0.77} & 86.89{\tiny$\pm$0.09} & +15.16{\tiny$\pm$0.68} \\
Enron & 77.46{\tiny$\pm$0.21} & 85.59{\tiny$\pm$0.40} & +8.12{\tiny$\pm$0.19} & 66.13{\tiny$\pm$0.49} & 82.43{\tiny$\pm$1.13} & +16.30{\tiny$\pm$0.66} \\
LastFM & 73.56 & 90.88 & +17.32 & 76.57 & 90.96 & +14.39 \\
\textit{mean} & 76.78 & 89.06 & +12.28 & 68.53 & 85.81 & +17.27 \\
\midrule
\multicolumn{7}{l}{\textit{DSRD$^\dagger$}} \\
Wikipedia & 87.88{\tiny$\pm$0.14} & 91.34{\tiny$\pm$0.06} & +3.47{\tiny$\pm$0.12} & 86.41{\tiny$\pm$0.44} & 89.72{\tiny$\pm$0.02} & +3.31{\tiny$\pm$0.45} \\
Reddit & 93.46{\tiny$\pm$0.04} & 94.09{\tiny$\pm$0.04} & +0.63{\tiny$\pm$0.01} & 80.20{\tiny$\pm$1.04} & 86.87{\tiny$\pm$0.25} & +6.67{\tiny$\pm$0.79} \\
UCI & 80.64{\tiny$\pm$0.19} & 88.86{\tiny$\pm$0.18} & +8.21{\tiny$\pm$0.26} & 82.51{\tiny$\pm$0.21} & 90.13{\tiny$\pm$0.34} & +7.61{\tiny$\pm$0.47} \\
Enron & 85.32{\tiny$\pm$0.05} & 91.22{\tiny$\pm$0.20} & +5.91{\tiny$\pm$0.19} & 83.60{\tiny$\pm$0.18} & 89.21{\tiny$\pm$0.42} & +5.61{\tiny$\pm$0.26} \\
LastFM & 83.87{\tiny$\pm$1.43} & 94.95{\tiny$\pm$0.02} & +11.08{\tiny$\pm$1.42} & 86.50{\tiny$\pm$2.55} & 95.40{\tiny$\pm$0.03} & +8.90{\tiny$\pm$2.55} \\
\textit{mean} & 86.23 & 92.09 & +5.86 & 83.84 & 90.27 & +6.42 \\
\bottomrule
\end{tabular}
\end{table}

\section{Mathematical Properties of Multi-Timescale Exponential Kernels}
\label{app:exponential_kernels}

MTHR uses multiple exponential kernels with different decay rates to aggregate historical interactions. This section establishes the mathematical properties underlying this design. We first show that exponential kernels support exact state propagation under a scalar-update requirement. We then motivate the use of multiple timescales and logarithmically spaced decay rates, and show how exponential mixtures can represent slowly decaying temporal patterns. Finally, we describe the timescale configuration used in our implementation.

\subsection{Exponential Kernels and Exact State Propagation}
\label{app:exact_state_propagation}

Consider a general linear historical representation:
\[
H_{u,\psi}(t)
=
\sum_{\substack{e\ni u\\t_e<t}}
\kappa(t-t_e)\psi_u(e),
\]
where $\kappa(s)$ denotes the weight retained by an interaction after an elapsed time $s\geq0$, and $\psi_u(e)$ denotes the content contributed by the interaction.

We require that, if no new interaction occurs between $t$ and $t+\Delta$, the historical state can be updated by multiplying it by a scalar $a(\Delta)$ that is independent of the interaction history:
\[
H_{u,\psi}(t+\Delta)
=
a(\Delta)H_{u,\psi}(t).
\]

If this requirement holds for arbitrary interaction histories and contents, it must also hold for a history containing only one interaction. Therefore,
\begin{equation}
\kappa(s+\Delta)
=
a(\Delta)\kappa(s),
\qquad s,\Delta\geq0.
\tag{B.1}
\label{eq:app-b1}
\end{equation}

Assume that $\kappa$ is continuous, strictly positive, and normalized such that $\kappa(0)=1$. Setting $s=0$ in \eqref{eq:app-b1} gives $a(\Delta)=\kappa(\Delta)$. Hence,
\[
\kappa(s+\Delta)
=
\kappa(s)\kappa(\Delta).
\]

Let $F(s)=\log\kappa(s)$. We then have
\[
F(s+\Delta)=F(s)+F(\Delta).
\]

By the continuity of $F$ and the additive functional equation, $F(s)=cs$ for some constant $c$. Requiring the historical weight to decrease strictly with elapsed time gives $c<0$. Writing $c=-\omega$, we obtain
\begin{equation}
\boxed{
\kappa(s)=e^{-\omega s},
\qquad \omega>0.
}
\tag{B.2}
\label{eq:app-b2}
\end{equation}

Thus, under the stated assumptions, the exponential kernel is the unique decaying kernel that permits exact scalar propagation of an arbitrary linear historical state.

The exponential kernel also provides an exact update when new interactions occur. For any $a\leq t$, partition the historical interactions into those occurring before $a$ and those occurring in $[a,t)$. Using the multiplicative property of the exponential function, we obtain
\begin{equation}
\boxed{
\begin{aligned}
H_{u,\psi}^{k}(t)
={}&
e^{-\omega_k(t-a)}
H_{u,\psi}^{k}(a)
\\
&+
\sum_{\substack{e\ni u\\a\leq t_e<t}}
e^{-\omega_k(t-t_e)}\psi_u(e).
\end{aligned}
}
\tag{B.3}
\label{eq:app-b3}
\end{equation}

This identity provides the mathematical basis for block-start caching and query-time historical updates in MTHR. Once earlier interactions have been summarized, their contributions can be propagated to a later query time without revisiting them. Only newly observed interactions need to be incorporated.

The same identity applies to the activity counts introduced in Appendix~\ref{app:shared_historical_foundation}, since the update rule is independent of whether the interaction content is a node embedding or a unit scalar.

\subsection{Multiple Timescales and Logarithmically Spaced Decay Rates}
\label{app:timescale_grid}

Although a single exponential kernel supports exact state propagation, a fixed decay rate provides temporal resolution over a limited range.

Define the characteristic time constant of the $k$-th kernel as
\[
\tau_k=\frac{1}{\omega_k}.
\]

After an interaction has aged by $\tau_k$, its weight decreases to $e^{-1}$ of its initial value. For elapsed times much smaller than $\tau_k$, the exponential weights remain close to one, making nearby interactions difficult to distinguish. For elapsed times much larger than $\tau_k$, the weights become very small, and the corresponding interactions contribute little to the current state.

A single timescale is therefore insufficient to retain fine-grained information across widely different temporal ranges. MTHR instead uses $K$ decay rates:
\[
\omega_1>\omega_2>\cdots>\omega_K>0.
\]

Larger decay rates emphasize recent interactions, while smaller decay rates retain information from more distant history.

To cover multiple orders of magnitude, we space the decay rates uniformly on a logarithmic scale. Equivalently, consecutive characteristic time constants have a constant ratio:
\begin{equation}
\boxed{
\frac{\tau_{k+1}}{\tau_k}=r>1.
}
\tag{B.4}
\label{eq:app-b4}
\end{equation}

Unlike uniform spacing on the linear time axis, logarithmic spacing provides approximately uniform relative resolution across different orders of magnitude. For example, the interval from one second to ten seconds occupies the same logarithmic range as the interval from one day to ten days.

This construction is also consistent with changes in time units. If all timestamps are scaled by a common factor and the decay rates are scaled inversely, the exponential weights and the ratios between consecutive timescales remain unchanged.

Logarithmic spacing therefore offers a practical way to allocate a finite number of temporal scales across a wide range of interaction histories, without assuming that all interactions follow one particular decay rate.

\subsection{Representing Slowly Decaying Patterns with Exponential Mixtures}
\label{app:slow_decay}

The role of multiple exponential kernels extends beyond retaining information at several characteristic timescales. An appropriate mixture of exponential kernels can also produce an overall temporal response that decays more slowly than any individual exponential over a range of interest.

To establish this property, consider the power-law function $s^{-\alpha}$, where $s>0$ and $\alpha>0$. Recall the Gamma function:
\[
\Gamma(\alpha)
=
\int_0^\infty
z^{\alpha-1}e^{-z}\,\mathrm dz.
\]

Applying the substitution $z=\omega s$, we obtain
\[
\begin{aligned}
\int_0^\infty
\omega^{\alpha-1}e^{-\omega s}
\,\mathrm d\omega
&=
s^{-\alpha}
\int_0^\infty
z^{\alpha-1}e^{-z}
\,\mathrm dz
\\
&=
s^{-\alpha}\Gamma(\alpha).
\end{aligned}
\]

Consequently,
\begin{equation}
\boxed{
s^{-\alpha}
=
\frac{1}{\Gamma(\alpha)}
\int_0^\infty
\omega^{\alpha-1}
e^{-\omega s}
\,\mathrm d\omega.
}
\tag{B.5}
\label{eq:app-b5}
\end{equation}

\Eqref{eq:app-b5} expresses power-law decay exactly as a continuous positive mixture of exponential kernels. By combining kernels with different decay rates, the overall temporal response is no longer restricted to the decay profile of a single exponential.

To connect this representation to the logarithmically spaced grid used in MTHR, let
\[
\ell=\log\omega,
\qquad
\omega=e^\ell,
\qquad
\mathrm d\omega=e^\ell\,\mathrm d\ell.
\]

Substituting these expressions into \eqref{eq:app-b5} gives
\begin{equation}
\boxed{
s^{-\alpha}
=
\frac{1}{\Gamma(\alpha)}
\int_{-\infty}^{\infty}
e^{\alpha\ell}
e^{-e^\ell s}
\,\mathrm d\ell.
}
\tag{B.6}
\label{eq:app-b6}
\end{equation}

Approximating this integral using finitely many uniformly spaced points on the logarithmic axis yields a finite exponential mixture with logarithmically spaced decay rates:
\[
s^{-\alpha}
\approx
\sum_{k=1}^{K}
c_ke^{-\omega_ks},
\qquad c_k>0,
\]
where the coefficients $c_k$ incorporate the corresponding integration weights.

This representation provides a mathematical motivation for using logarithmically spaced exponential basis functions. Their combinations can capture slowly decaying temporal influences while retaining the exact state-propagation property of each individual kernel.

A finite mixture introduces approximation error through the truncation of the integration range and the discretization of the logarithmic axis. In our model, the exponential kernels serve as temporal basis functions whose historical states are subsequently combined through learnable transformations and nonlinear networks. The resulting representations are therefore more general than a single positive exponential mixture.

\subsection{Timescale Configuration}
\label{app:timescale_configuration}

We determine the timescale grid from the temporal range of the training data rather than manually specifying a fixed set of durations for each dataset.

Using only the timestamps of the filtered training interactions, let $\Delta t_{\min}$ denote the smallest positive difference between consecutive distinct timestamps, and let $L$ denote the time span between the first and last training timestamps.

For training data containing at least two distinct timestamps, we define the range of decay rates as
\begin{equation}
\boxed{
\omega_{\max}
=
\frac{1}{\Delta t_{\min}},
\qquad
\omega_{\min}
=
\frac{1}{L}.
}
\tag{B.7}
\label{eq:app-b7}
\end{equation}

Here, $\omega_{\max}$ corresponds to the shortest characteristic time constant, while $\omega_{\min}$ corresponds to the longest.

We allocate approximately two intervals per decade on the logarithmic time axis, subject to a minimum of two and a maximum of sixteen timescales:
\begin{equation}
\boxed{
K=
\min\left(
16,\,
\max\left(
2,\,
\left\lceil
2\log_{10}
\frac{L}{\Delta t_{\min}}
\right\rceil+1
\right)
\right).
}
\tag{B.8}
\label{eq:app-b8}
\end{equation}

Given $K$, the decay rates are uniformly spaced on the logarithmic axis:
\begin{equation}
\boxed{
\omega_k
=
\omega_{\max}
\left(
\frac{\omega_{\min}}{\omega_{\max}}
\right)^{\frac{k-1}{K-1}},
\qquad
k=1,\ldots,K.
}
\tag{B.9}
\label{eq:app-b9}
\end{equation}

Thus, $\omega_1=\omega_{\max}$ and $\omega_K=\omega_{\min}$, with a constant ratio between consecutive characteristic time constants.

For MOOC, the saved timescale configuration records $\Delta t_{\min}=1$ second and $L=1{,}909{,}964$ seconds, resulting in $K=14$. The characteristic time constants therefore range from approximately one second to 22.1 days, with consecutive time constants differing by a factor of approximately $3.04$.

All decay rates are determined before training and remain fixed. The model learns how to combine the historical representations at different timescales, while SNAM separately learns their contributions to source event intensity. The timescale grid determines the available temporal resolutions, and the relative importance of those resolutions is learned from the training data.
\begin{table}[p]
\caption{Number of timescales $K$, transductive and inductive link prediction: test AP (\%), mean{\scriptsize$\pm$std} over three seeds, under random (rnd), historical (hist) and inductive (ind) negative sampling; $f$ = the base score alone ($\beta=0$), $f{+}\lambda$ = with the source-activity term ($\beta$ chosen on validation). $\star$: the configuration of the main results; $K=1$ keeps a single timescale at the centre of the covered range, the other rows use one, two and four intervals per decade of it. Bold: best of the four rows of a dataset. In the inductive setting both samplers draw from the same new-node subset and agree to within 0.06 at a fixed $\beta$, so only the historical columns are shown.}
\label{tab:a1_scales_full}
\centering
\footnotesize
\setlength{\tabcolsep}{2.5pt}
\renewcommand{\arraystretch}{0.92}
\begin{tabular}{ccccccccc}
\toprule
 & \multicolumn{5}{c}{Transductive} & \multicolumn{3}{c}{Inductive} \\
\cmidrule(lr){2-6} \cmidrule(lr){7-9}
$K$ & rnd & hist $f$ & hist $f{+}\lambda$ & ind $f$ & ind $f{+}\lambda$ & rnd & hist $f$ & hist $f{+}\lambda$ \\
\midrule
\multicolumn{9}{l}{\textit{Wikipedia}} \\
1 & 95.19{\scriptsize$\pm$0.02} & \textbf{89.51}{\scriptsize$\pm$0.03} & 91.91{\scriptsize$\pm$0.03} & \textbf{87.25}{\scriptsize$\pm$0.03} & 90.05{\scriptsize$\pm$0.07} & 95.61{\scriptsize$\pm$0.02} & \textbf{87.67}{\scriptsize$\pm$0.04} & \textbf{90.18}{\scriptsize$\pm$0.05} \\
8 & \textbf{98.96}{\scriptsize$\pm$0.02} & 87.87{\scriptsize$\pm$1.75} & 93.22{\scriptsize$\pm$0.01} & 85.68{\scriptsize$\pm$1.47} & \textbf{91.43}{\scriptsize$\pm$0.03} & 98.40{\scriptsize$\pm$0.02} & 77.58{\scriptsize$\pm$2.64} & 88.49{\scriptsize$\pm$0.07} \\
14$^\star$ & 98.95{\scriptsize$\pm$0.02} & 88.71{\scriptsize$\pm$0.80} & 93.25{\scriptsize$\pm$0.05} & 86.19{\scriptsize$\pm$1.21} & 91.23{\scriptsize$\pm$0.09} & \textbf{98.41}{\scriptsize$\pm$0.03} & 77.78{\scriptsize$\pm$1.86} & 88.29{\scriptsize$\pm$0.13} \\
27 & 98.94{\scriptsize$\pm$0.02} & 88.39{\scriptsize$\pm$0.94} & \textbf{93.26}{\scriptsize$\pm$0.04} & 84.16{\scriptsize$\pm$2.83} & 90.85{\scriptsize$\pm$0.21} & 98.40{\scriptsize$\pm$0.06} & 77.66{\scriptsize$\pm$2.40} & 88.23{\scriptsize$\pm$0.11} \\
\midrule
\multicolumn{9}{l}{\textit{Reddit}} \\
1 & 89.32{\scriptsize$\pm$0.05} & 73.97{\scriptsize$\pm$0.12} & 77.50{\scriptsize$\pm$0.30} & 80.17{\scriptsize$\pm$0.11} & 82.47{\scriptsize$\pm$0.11} & 85.42{\scriptsize$\pm$0.14} & 63.10{\scriptsize$\pm$0.25} & 67.71{\scriptsize$\pm$0.52} \\
11 & \textbf{99.15}{\scriptsize$\pm$0.01} & 82.24{\scriptsize$\pm$0.22} & 90.14{\scriptsize$\pm$0.13} & 90.54{\scriptsize$\pm$0.10} & 91.78{\scriptsize$\pm$0.05} & 98.20{\scriptsize$\pm$0.03} & \textbf{67.66}{\scriptsize$\pm$0.67} & \textbf{79.24}{\scriptsize$\pm$0.06} \\
16$^\star$ & \textbf{99.15}{\scriptsize$\pm$0.00} & \textbf{82.74}{\scriptsize$\pm$1.44} & \textbf{90.16}{\scriptsize$\pm$0.17} & \textbf{90.62}{\scriptsize$\pm$0.27} & \textbf{91.79}{\scriptsize$\pm$0.13} & 98.22{\scriptsize$\pm$0.02} & 67.54{\scriptsize$\pm$2.74} & 79.20{\scriptsize$\pm$0.34} \\
32 & \textbf{99.15}{\scriptsize$\pm$0.01} & 81.73{\scriptsize$\pm$0.75} & 90.01{\scriptsize$\pm$0.07} & 90.15{\scriptsize$\pm$0.21} & 91.45{\scriptsize$\pm$0.17} & \textbf{98.23}{\scriptsize$\pm$0.02} & 65.83{\scriptsize$\pm$0.53} & 78.99{\scriptsize$\pm$0.15} \\
\midrule
\multicolumn{9}{l}{\textit{UCI}} \\
1 & 83.62{\scriptsize$\pm$0.65} & 63.43{\scriptsize$\pm$0.85} & 79.48{\scriptsize$\pm$0.41} & 75.49{\scriptsize$\pm$0.96} & 84.60{\scriptsize$\pm$0.72} & 82.47{\scriptsize$\pm$1.11} & 74.77{\scriptsize$\pm$0.61} & 84.73{\scriptsize$\pm$0.52} \\
8 & \textbf{96.18}{\scriptsize$\pm$0.14} & 90.52{\scriptsize$\pm$1.20} & 95.43{\scriptsize$\pm$0.32} & 83.26{\scriptsize$\pm$0.52} & 87.93{\scriptsize$\pm$0.41} & \textbf{94.77}{\scriptsize$\pm$0.16} & 83.71{\scriptsize$\pm$0.45} & 88.55{\scriptsize$\pm$0.20} \\
15$^\star$ & 95.81{\scriptsize$\pm$0.41} & \textbf{91.47}{\scriptsize$\pm$0.52} & 96.04{\scriptsize$\pm$0.17} & \textbf{83.35}{\scriptsize$\pm$0.29} & 88.18{\scriptsize$\pm$0.04} & 94.74{\scriptsize$\pm$0.12} & \textbf{84.08}{\scriptsize$\pm$0.35} & 89.03{\scriptsize$\pm$0.05} \\
28 & 95.86{\scriptsize$\pm$0.57} & 91.42{\scriptsize$\pm$0.21} & \textbf{96.12}{\scriptsize$\pm$0.10} & 83.09{\scriptsize$\pm$0.05} & \textbf{88.25}{\scriptsize$\pm$0.03} & 94.75{\scriptsize$\pm$0.06} & 84.01{\scriptsize$\pm$0.18} & \textbf{89.16}{\scriptsize$\pm$0.09} \\
\midrule
\multicolumn{9}{l}{\textit{Enron}} \\
1 & 82.67{\scriptsize$\pm$0.23} & 74.24{\scriptsize$\pm$0.63} & 83.47{\scriptsize$\pm$0.26} & 72.51{\scriptsize$\pm$0.39} & 81.08{\scriptsize$\pm$0.40} & 79.19{\scriptsize$\pm$0.28} & 72.86{\scriptsize$\pm$0.88} & 81.09{\scriptsize$\pm$0.51} \\
9 & \textbf{93.16}{\scriptsize$\pm$0.09} & 83.45{\scriptsize$\pm$1.18} & 90.46{\scriptsize$\pm$0.12} & 84.98{\scriptsize$\pm$0.48} & 88.92{\scriptsize$\pm$0.21} & \textbf{89.74}{\scriptsize$\pm$0.13} & 77.63{\scriptsize$\pm$1.03} & 85.44{\scriptsize$\pm$0.50} \\
16$^\star$ & 93.09{\scriptsize$\pm$0.21} & 84.04{\scriptsize$\pm$2.43} & 90.26{\scriptsize$\pm$0.77} & 85.32{\scriptsize$\pm$0.85} & 89.04{\scriptsize$\pm$0.18} & 89.51{\scriptsize$\pm$0.10} & 78.56{\scriptsize$\pm$2.85} & 85.57{\scriptsize$\pm$0.91} \\
32 & 93.07{\scriptsize$\pm$0.06} & \textbf{85.74}{\scriptsize$\pm$0.71} & \textbf{90.62}{\scriptsize$\pm$0.13} & \textbf{86.01}{\scriptsize$\pm$0.43} & \textbf{89.06}{\scriptsize$\pm$0.18} & 89.51{\scriptsize$\pm$0.11} & \textbf{80.45}{\scriptsize$\pm$0.56} & \textbf{85.82}{\scriptsize$\pm$0.19} \\
\midrule
\multicolumn{9}{l}{\textit{LastFM}} \\
1 & 79.04{\scriptsize$\pm$0.12} & 73.39{\scriptsize$\pm$0.17} & 92.67{\scriptsize$\pm$0.01} & 68.91{\scriptsize$\pm$0.22} & 88.25{\scriptsize$\pm$0.25} & 84.32{\scriptsize$\pm$0.20} & 76.88{\scriptsize$\pm$0.14} & 89.87{\scriptsize$\pm$0.19} \\
10 & 92.16{\scriptsize$\pm$0.66} & 84.53{\scriptsize$\pm$0.57} & 95.22{\scriptsize$\pm$0.08} & 73.16{\scriptsize$\pm$0.64} & 88.77{\scriptsize$\pm$0.32} & 92.80{\scriptsize$\pm$0.62} & 78.21{\scriptsize$\pm$0.59} & 89.26{\scriptsize$\pm$0.26} \\
16$^\star$ & 91.87{\scriptsize$\pm$1.35} & \textbf{84.60}{\scriptsize$\pm$1.20} & \textbf{95.36}{\scriptsize$\pm$0.03} & \textbf{73.96}{\scriptsize$\pm$0.95} & \textbf{89.82}{\scriptsize$\pm$0.71} & 91.60{\scriptsize$\pm$2.72} & 78.38{\scriptsize$\pm$0.70} & \textbf{90.12}{\scriptsize$\pm$0.61} \\
32 & \textbf{92.43}{\scriptsize$\pm$0.27} & 84.44{\scriptsize$\pm$0.09} & 95.29{\scriptsize$\pm$0.04} & 73.60{\scriptsize$\pm$0.54} & 89.43{\scriptsize$\pm$0.09} & \textbf{93.12}{\scriptsize$\pm$0.14} & \textbf{78.54}{\scriptsize$\pm$0.47} & 89.84{\scriptsize$\pm$0.19} \\
\bottomrule
\end{tabular}
\end{table}

\section{Detailed Construction of Candidate Relation Modeling}
\label{app:crm}

MTHR aggregates historical interactions into multi-timescale node representations. However, concatenating the representations of two endpoints still leaves the scoring network to recover their specific historical relationships. To provide this information explicitly, Candidate Relation Modeling (CRM) introduces two complementary types of candidate-specific features. The first extracts interaction and structural information from the complete observable history. The second encodes individual recent interactions of the source node to measure their relevance to the current candidate destination.

We denote these features by $\eta_{uv}(t)$ and $c_{uv}(t)$, respectively. Together with the MTHR node representations $h_u(t)$ and $h_v(t)$, they form the inputs to the final relation scoring network. This section specifies their construction, computation, and network architecture. All historical queries use only interactions satisfying $t_e<t$.

\subsection{Node-Pair Features from the Complete History}
\label{app:pair_features}

For a candidate interaction $(u,v,t)$, CRM first constructs the node-pair historical feature vector $\eta_{uv}(t)$. It combines direct interaction counts, endpoint activity levels, learnable historical representation matching, and historical edge-feature matching. Structural features obtained from fixed random projections are also included in $\eta_{uv}(t)$ and are described in the next subsection.

\paragraph{Direct interactions and endpoint activity counts.}

For the $k$-th timescale, define the time-decayed interaction count of an unordered node pair as
\[
n_{uv}^{k}(t)
=
\sum_{\substack{
e:\{u_e,v_e\}=\{u,v\}\\
t_e<t
}}
e^{-\omega_k(t-t_e)}.
\]

This quantity records repeated interactions between the candidate endpoints, weighting each interaction according to its age. In bipartite interaction datasets such as MOOC, node types distinguish students from content nodes, so using unordered pairs does not confuse the roles of the two endpoints.

Together with the endpoint activity counts defined in Appendix~\ref{app:shared_historical_foundation}, CRM retains three scalar features at each timescale:
\begin{equation}
\boxed{
\begin{gathered}
\log(1+n_{uv}^{k}(t)),\\
\log(1+n_u^{k}(t)),\qquad \log(1+n_v^{k}(t)).
\end{gathered}
}
\tag{C.1}
\label{eq:app-c1}
\end{equation}

These features contribute $3K$ dimensions in total. The pairwise count captures repeated interactions, while the two endpoint counts provide the activity context needed to assess the pairwise history. The logarithmic transformation compresses the dynamic range of large counts while preserving their monotonicity.

\paragraph{Learnable low-rank historical representation matching.}

Direct interaction counts alone cannot fully capture the neighborhood information of the two endpoints. We therefore introduce learnable matching between the first- and second-order historical representations produced by MTHR.

For two $d$-dimensional vectors $x$ and $y$, a general bilinear matching function takes the form $x^\top Cy$, where $C\in\mathbb R^{d\times d}$. We factorize $C=L^\top R$, with $L,R\in\mathbb R^{r\times d}$, obtaining
\begin{equation}
\boxed{
x^\top Cy
=
\langle Lx,Ry\rangle,
\qquad \operatorname{rank}(C)\leq r.
}
\tag{C.2}
\label{eq:app-c2}
\end{equation}

This formulation performs matching in a lower-dimensional space. We use $r=16$ and construct three matching scores at each timescale:
\begin{equation}
\boxed{
\begin{aligned}
q_{11}^{k}
&=\left\langle
L_{11}^{k}\widetilde x_{1,u}^{k},
R_{11}^{k}\widetilde x_{1,v}^{k}
\right\rangle,\\
q_{12}^{k}
&=\left\langle
L_{12}^{k}\widetilde x_{1,u}^{k},
R_{12}^{k}\widetilde x_{2,v}^{k}
\right\rangle,\\
q_{21}^{k}
&=\left\langle
L_{21}^{k}\widetilde x_{2,u}^{k},
R_{21}^{k}\widetilde x_{1,v}^{k}
\right\rangle.
\end{aligned}
}
\tag{C.3}
\label{eq:app-c3}
\end{equation}

Here, $\widetilde x_{1,u}^{k}$ and $\widetilde x_{2,u}^{k}$ denote the normalized first- and second-order MTHR representations of node $u$ for the current query, with the query time and block index suppressed for readability. The three scores capture first-order matching and cross-order matching in both directions.

In bipartite graphs such as MOOC, the first-order representation of one endpoint and the second-order representation of the other can aggregate information from the same node type. Cross-order matching is therefore particularly suitable for capturing indirect associations between a student and a candidate content node.

Since these matching scores can be positive or negative, we apply a signed logarithmic transformation to reduce the influence of large magnitudes while preserving their signs:
\begin{equation}
\boxed{
\operatorname{slog}(q)
=\operatorname{sign}(q)\log(1+|q|).
}
\tag{C.4}
\label{eq:app-c4}
\end{equation}

The three transformed low-rank matching scores contribute $3K$ feature dimensions.

We additionally retain a static matching score between the base node embeddings:
\[
q_{\mathrm{static}}
=\frac{\langle E_u,E_v\rangle}{\sqrt d}.
\]

This term does not vary with query time and can capture stable node-specific relationships learned during training.

When valid historical edge features are available, let $\widetilde x_{f,u}^{k}(t)\in\mathbb R^F$ denote the normalized edge-feature aggregate at the corresponding timescale. We further compute
\[
q_{ff}^{k}
=\left\langle
C_f^{k}\widetilde x_{f,u}^{k},
C_f^{k}\widetilde x_{f,v}^{k}
\right\rangle,
\qquad C_f^{k}\in\mathbb R^{r\times F}.
\]

This score also undergoes the transformation in \eqref{eq:app-c4} and contributes one scalar per timescale.

Edge features are standardized using the means and standard deviations computed from training interactions, with zero-variance columns removed. The same statistics are reused during validation and testing. Historical edge-feature aggregation uses only previously observed interactions and never accesses the features of the interaction currently being predicted.

\subsection{Structural Features from Fixed Random Projections}
\label{app:random_projection_features}

The low-rank matching scores above are based on trainable node representations, which reflect both historical structure and the supervised prediction objective. CRM introduces an additional structural channel based on fixed random projections. This channel compresses weighted neighborhood overlap into low-dimensional inner products without relying on the trainable node embedding table.

Let $N$ denote the total number of nodes. We assign each node $w$ a fixed random vector $R_w\in\mathbb R^D$ and collect these vectors into a matrix $R\in\mathbb R^{N\times D}$. Its entries are sampled independently according to
\begin{equation}
\boxed{
R_{w,j}
=\begin{cases}
+1/\sqrt D,&\text{with probability }1/2,\\
-1/\sqrt D,&\text{with probability }1/2.
\end{cases}
}
\tag{C.5}
\label{eq:app-c5}
\end{equation}

We use $D=256$ and fix the random initialization seed to $7$. The projection matrix remains unchanged throughout training.

For the same node, $\langle R_w,R_w\rangle=1$. For two distinct nodes $w\neq z$,
\[
\mathbb E_R\langle R_w,R_z\rangle=0,
\qquad
\operatorname{Var}_R
\bigl(\langle R_w,R_z\rangle\bigr)
=\frac1D.
\]

Both the expectation and variance are taken over the random initialization of the projection matrix.

Let $a,b\in\mathbb R^N$ be two weighted neighborhood vectors. Compressing them into $a^\top R$ and $b^\top R$ gives
\begin{equation}
\begin{aligned}
&\mathbb E_R
\bigl[\langle a^\top R,b^\top R\rangle\bigr]\\
&\quad=\sum_{w,z}a_wb_z\,
\mathbb E_R\langle R_w,R_z\rangle\\
&\quad=\sum_w a_wb_w=a^\top b.
\end{aligned}
\tag{C.6}
\label{eq:app-c6}
\end{equation}

Thus, the inner product of the compressed vectors is an unbiased estimator of the original weighted neighborhood inner product. The representation dimension is fixed at $D$ and does not increase with the total number of nodes.

\paragraph{First- and second-order structural projections.}

For a fixed timescale $k$, let $A_c^k$ denote the time-decayed adjacency matrix at the beginning of block $c$. We construct
\begin{equation}
\boxed{
\begin{aligned}
S_{1,c}^{k}&=A_c^kR,\\
S_{2,c}^{k}&=A_c^kS_{1,c}^{k}
=(A_c^k)^2R.
\end{aligned}
}
\tag{C.7}
\label{eq:app-c7}
\end{equation}

Since the historical adjacency matrix is symmetric, \eqref{eq:app-c6} provides the following interpretations of inner products between projections of different orders:
\begin{equation}
\boxed{
\begin{aligned}
\mathbb E_R
\langle S_{1,c,u}^{k},S_{1,c,v}^{k}\rangle
&=\bigl[(A_c^k)^2\bigr]_{uv},\\
\mathbb E_R
\langle S_{1,c,u}^{k},S_{2,c,v}^{k}\rangle
&=\bigl[(A_c^k)^3\bigr]_{uv},\\
\mathbb E_R
\langle S_{2,c,u}^{k},S_{1,c,v}^{k}\rangle
&=\bigl[(A_c^k)^3\bigr]_{uv},\\
\mathbb E_R
\langle S_{2,c,u}^{k},S_{2,c,v}^{k}\rangle
&=\bigl[(A_c^k)^4\bigr]_{uv}.
\end{aligned}
}
\tag{C.8}
\label{eq:app-c8}
\end{equation}

These matrix powers correspond to weighted walks of different lengths. For example, $\bigl[(A_c^k)^3\bigr]_{uv}$ sums the products of edge weights over all three-step walks from $u$ to $v$.

For the MOOC bipartite graph, placing student nodes before content nodes gives the block adjacency matrix
\[
A=
\begin{bmatrix}
0 & B\\
B^\top & 0
\end{bmatrix},
\qquad
A^2=
\begin{bmatrix}
BB^\top & 0\\
0 & B^\top B
\end{bmatrix}.
\]

Consequently, even-length walk weights between a student and a content node are zero, while three-step walks can capture indirect associations of the form student--content--student--candidate content. This motivates retaining both same-order and cross-order structural projections in CRM.

\paragraph{Query-time structural states.}

For a query time $t\geq a_c$ within block $c$, the first-order random projection incorporates all newly observed interactions strictly preceding $t$:
\[
\begin{aligned}
s_{1,u}^{k}(t)
={}&e^{-\omega_k(t-a_c)}S_{1,c,u}^{k}\\
&+\sum_{\substack{e\ni u\\a_c\leq t_e<t}}
e^{-\omega_k(t-t_e)}R_{p_u(e)}.
\end{aligned}
\]

The second-order projection uses the cached block-start first-order projections as neighbor content:
\begin{equation}
\begin{aligned}
s_{2,u}^{k}(t;c)
={}&e^{-\omega_k(t-a_c)}S_{2,c,u}^{k}\\
&+\sum_{\substack{e\ni u\\a_c\leq t_e<t}}
e^{-\omega_k(t-t_e)}
S_{1,c,p_u(e)}^{k}.
\end{aligned}
\tag{C.9}
\label{eq:app-c9}
\end{equation}

This construction reuses the block-start representations and avoids recomputing complete two-layer random projections for every query.

The structural interpretation differs slightly between block-start and within-block states. If $B_t^k$ denotes the time-decayed adjacency matrix constructed from the complete history preceding the query time, then
\[
s_1^k(t)=B_t^kR,
\qquad
s_2^k(t;c)=B_t^kA_c^kR.
\]

Thus, the powers of a single adjacency matrix in \eqref{eq:app-c8} describe the block-start states exactly. Within a block, the second-order structural channel combines historical graphs from the query time and the block-start time.

Finally, CRM applies the following transformation to the four query-time projection inner products:
\begin{equation}
\boxed{
\chi(q)=0.1\log(1+\max(q,0)).
}
\tag{C.10}
\label{eq:app-c10}
\end{equation}

The resulting structural features are
\[
\begin{aligned}
\kappa_{ij}^k(u,v,t)
&=\chi\left(\left\langle
s_{i,u}^k(t;c),s_{j,v}^k(t;c)
\right\rangle\right),\\
&(i,j)\in\{1,2\}^2,
\end{aligned}
\]
where $s_{1,u}^k(t;c)$ is shorthand for $s_{1,u}^k(t)$.

This transformation clips negative values and compresses the dynamic range of large inner products. The unbiasedness in \eqref{eq:app-c6} applies to the original inner products before nonlinear processing. After applying $\chi$, the resulting features are no longer unbiased estimators of the original weighted walk sums.

\paragraph{Final composition of node-pair features.}

We concatenate the features from Appendix~\ref{app:pair_features} and the structural features defined above across all timescales to obtain $\eta_{uv}(t)$. Its components and dimensions are summarized in Table~\ref{tab:crm_pair_dimensions}.

\begin{table}[t]
\centering
\caption{Composition of the node-pair historical feature vector $\eta_{uv}(t)$. The edge-feature matching term is included when valid historical edge features are available.}
\label{tab:crm_pair_dimensions}
\small
\setlength{\tabcolsep}{4pt}
\begin{tabular}{lc}
\hline
Feature group & Dimension\\
\hline
Low-rank representation matches (3 types) & $3K$\\
Pair and endpoint activity counts & $3K$\\
Static node matching & $1$\\
Random-projection matches (4 types) & $4K$\\
Historical edge-feature matching & $K$\\
\hline
Total & $11K+1$\\
\hline
\end{tabular}
\end{table}

For MOOC, $K=14$, giving $\eta_{uv}(t)\in\mathbb R^{155}$. If the dataset contains no valid historical edge features, the corresponding $K$ matching terms are omitted, reducing the dimension to $10K+1$.

\subsection{Candidate-Conditioned Recent Interaction Encoding}
\label{app:recent_interaction_encoding}

Aggregated historical states efficiently preserve information across multiple timescales, but their linear aggregation takes place before nonlinear processing. Different sets of historical interactions may therefore produce identical or similar aggregated representations, making their individual event attributes difficult to recover.

To complement these aggregated representations, CRM introduces a recent-interaction encoding branch. For a candidate interaction $(u,v,t)$, we select up to the $M$ most recent interactions of source node $u$ with timestamps strictly smaller than $t$. Denote this set by $\mathcal R_u(t)$. We use $M=20$ by default and include all available interactions if the source has fewer than $M$ historical records. Selected interactions may come from earlier blocks or from the current block, provided that they strictly precede the query time.

For each $e\in\mathcal R_u(t)$, let $p=p_u(e)$ denote its historical interaction partner. We construct four groups of features relative to the current candidate destination $v$:
\begin{equation}
\boxed{
\begin{aligned}
a_e^k(t)
&=e^{-\omega_k(t-t_e)},\\
b_{e,v}^{k}
&=\chi\left(
\langle S_{1,c,p}^{k},S_{1,c,v}^{k}\rangle
\right),\\
j_{e,v}
&=\frac{\langle W_{\mathrm{sim}}m_{c,p},m_{c,v}\rangle}{\sqrt d},\\
\zeta_{e,v}(t)
&=\left[
(a_e^k(t))_{k=1}^{K};
(b_{e,v}^{k})_{k=1}^{K};
 j_{e,v};\bar\phi_e
\right].
\end{aligned}
}
\tag{C.11}
\label{eq:app-c11}
\end{equation}

Here, $m_{c,p}$ and $m_{c,v}$ are the intermediate node representations computed by MTHR at the block start, $W_{\mathrm{sim}}\in\mathbb R^{d\times d}$ is a learnable matching matrix, and $\bar\phi_e\in\mathbb R^F$ denotes the standardized features of the historical interaction.

The four groups capture complementary information. The temporal weights $a_e^k(t)$ describe the age of the historical interaction. The structural scores $b_{e,v}^k$ measure the historical neighborhood overlap between the previous interaction partner and the candidate destination. The learnable score $j_{e,v}$ measures their compatibility in the representation space. Finally, $\bar\phi_e$ retains the observed features of the historical interaction itself.

Interaction selection and temporal weights are computed at the exact query time, whereas the structural projections and intermediate node representations are reused from the block-start cache. Each event therefore produces a feature vector of dimension $2K+1+F$, which equals $33$ for the common MOOC configuration with $K=14$ and $F=4$.

We then apply a shared two-layer network to each selected event:
\[
\begin{aligned}
\rho(z)
&=\operatorname{GELU}\!\left(
W_b\operatorname{GELU}(W_az+b_a)+b_b
\right)\\
&\in\mathbb R^{64},
\end{aligned}
\]
where $W_a$ maps the event input to $64$ dimensions and $W_b\in\mathbb R^{64\times64}$.

The event embeddings are aggregated by summation:
\begin{equation}
\boxed{
c_{uv}(t)
=\sum_{e\in\mathcal R_u(t)}
\rho\bigl(\zeta_{e,v}(t)\bigr).
}
\tag{C.12}
\label{eq:app-c12}
\end{equation}

This architecture follows the shared element-wise encoding and permutation-invariant aggregation pattern of Deep Sets \citep{zaheer2017deepsets}. Related set-based representations have been studied in PointNet \citep{qi2017pointnet}, while the expressive properties and limitations of fixed-dimensional sum aggregation have been further examined by \cite{wagstaff2019limitations}. Here, we use a candidate-conditioned variant: each historical interaction is encoded relative to the current candidate destination before aggregation.

If the source node has no available historical interactions, we set $c_{uv}(t)=\mathbf 0$.

Unlike representations that aggregate the complete history before nonlinear processing, this branch applies a candidate-conditioned nonlinear transformation to each selected event before pooling. Summation makes the resulting representation invariant to the ordering of the selected event features, while temporal information remains explicitly encoded through $a_e^k(t)$. The selection of the most recent $M$ events still depends on their temporal order.

Since each event feature also depends on the candidate destination $v$, $c_{uv}(t)$ is a node-pair representation rather than a general source-node embedding that can be computed once and reused for all candidate destinations.

\subsection{Relation Scoring Architecture}
\label{app:relation_scoring_architecture}

The model now has three types of inputs for candidate relation prediction:
\[
x_{uv}(t)=[h_u(t);h_v(t)]\in\mathbb R^{2d},
\]
\[
\eta_{uv}(t)\in\mathbb R^{11K+1},
\qquad c_{uv}(t)\in\mathbb R^{64},
\]
where the stated dimension of $\eta_{uv}(t)$ assumes that valid historical edge features are available.

We combine these inputs using a scoring network with a nonlinear path and a direct linear path. First, we compute a $64$-dimensional hidden preactivation:
\begin{equation}
a=W_1x_{uv}+b_1+V_1\eta_{uv}+Q_1c_{uv}
\in\mathbb R^{64}.
\tag{C.13}
\label{eq:app-c13}
\end{equation}

We then retain the positive and negative parts of this preactivation separately:
\[
\varphi(a)=[\operatorname{ReLU}(a);\operatorname{ReLU}(-a)]
\in\mathbb R^{128}.
\]

Finally, the nonlinear and direct linear paths jointly produce two real-valued outputs:
\begin{equation}
\boxed{
\begin{aligned}
z={}&W_2\varphi(a)+Ux_{uv}+b_z\\
&+V_2\eta_{uv}+Q_2c_{uv}
\in\mathbb R^2.
\end{aligned}
}
\tag{C.14}
\label{eq:app-c14}
\end{equation}

The direct linear path provides an explicit mapping from the concatenated node representations to the outputs. The positive-negative activation split allows the nonlinear path to process positive and negative preactivations separately.

The parameters $V_1,V_2,Q_1,Q_2$ are initialized to zero, so the additional pairwise and recent-interaction channels initially leave the output of the base scoring head unchanged and gradually contribute during training. We apply dropout with probability $0.2$ to the concatenated node representations $x_{uv}$ and the complete-history pairwise features $\eta_{uv}$. No additional dropout is applied to the recent-interaction encoding $c_{uv}$ at this stage.

The two outputs $z_0,z_1$ do not represent separate ``no-edge'' and ``edge'' classes. Instead, they jointly define a scalar scoring function. During training, we use the smooth aggregation
\[
g(u,v,t)=\log\left(e^{z_0}+e^{z_1}\right),
\]
followed by binary cross-entropy with random destination negative sampling.

During evaluation, the base score is
\begin{equation}
\boxed{f(u,v,t)=\max(z_0,z_1).}
\tag{C.15}
\label{eq:app-c15}
\end{equation}

The two aggregation functions satisfy
\[
0\leq g(u,v,t)-f(u,v,t)\leq\log2.
\]

Thus, the training score is a smooth approximation to the evaluation score, although their difference varies across samples.

Finally, $f(u,v,t)$ serves as the base relation score produced by CRM. It is combined with the independently learned source intensity from Appendix~\ref{app:snam_foundations} to obtain the complete evaluation score $s_\beta(u,v,t)$. SNAM does not participate in the internal feature fusion of the relation scoring network; instead, it provides an independent source-activity correction at the final scoring stage.
\begin{table}[p]
\caption{Which inputs of the relation branch carry the pair information: test AP (\%), mean{\scriptsize$\pm$std} over three seeds, transductive and inductive link prediction under random (rnd), historical (hist) and inductive (ind) negative sampling; $f$ = the base score alone ($\beta=0$), $f{+}\lambda$ = with the source-activity term ($\beta$ chosen on validation). $\star$: the configuration of the main results. The other rows retrain the model without the recent interaction encoding $c_{uv}(t)$ of the source's last 20 events, without the structural features from random projections (which removes the recent interaction encoding as well, since it reads them), without the pair and endpoint counts $n_{uv}^k$, $n_u^k$, $n_v^k$, and without all three (the node embeddings, the bilinear matches and the static term only). Bold: best of the rows of a stream. In the inductive setting both samplers draw from the same new-node subset and agree to within 0.06 at a fixed $\beta$, so only the historical columns are shown.}
\label{tab:a4_trunk_channels}
\centering
\scriptsize
\setlength{\tabcolsep}{2pt}
\renewcommand{\arraystretch}{0.92}
\begin{tabular}{lcccccccc}
\toprule
 & \multicolumn{5}{c}{Transductive} & \multicolumn{3}{c}{Inductive} \\
\cmidrule(lr){2-6} \cmidrule(lr){7-9}
variant & rnd & hist $f$ & hist $f{+}\lambda$ & ind $f$ & ind $f{+}\lambda$ & rnd & hist $f$ & hist $f{+}\lambda$ \\
\midrule
\multicolumn{9}{l}{\textit{Wikipedia}} \\
all inputs$^\star$ & \textbf{98.95}{\tiny$\pm$0.02} & 88.71{\tiny$\pm$0.80} & 93.25{\tiny$\pm$0.05} & 86.19{\tiny$\pm$1.21} & \textbf{91.23}{\tiny$\pm$0.09} & \textbf{98.41}{\tiny$\pm$0.03} & 77.78{\tiny$\pm$1.86} & 88.29{\tiny$\pm$0.13} \\
no $c_{uv}$ & \textbf{98.95}{\tiny$\pm$0.02} & \textbf{90.37}{\tiny$\pm$0.69} & \textbf{93.36}{\tiny$\pm$0.08} & \textbf{87.80}{\tiny$\pm$0.37} & 91.16{\tiny$\pm$0.05} & 98.38{\tiny$\pm$0.01} & \textbf{81.79}{\tiny$\pm$1.32} & \textbf{88.41}{\tiny$\pm$0.05} \\
no projections & 98.91{\tiny$\pm$0.02} & 89.54{\tiny$\pm$0.36} & 93.35{\tiny$\pm$0.07} & 84.95{\tiny$\pm$0.89} & 90.93{\tiny$\pm$0.13} & 98.30{\tiny$\pm$0.02} & 79.72{\tiny$\pm$1.04} & 88.39{\tiny$\pm$0.03} \\
no counts & 98.25{\tiny$\pm$0.04} & 86.49{\tiny$\pm$1.09} & 92.94{\tiny$\pm$0.06} & 78.37{\tiny$\pm$1.97} & 90.54{\tiny$\pm$0.24} & 97.56{\tiny$\pm$0.09} & 74.19{\tiny$\pm$1.58} & 88.12{\tiny$\pm$0.11} \\
none of the three & 97.83{\tiny$\pm$0.11} & 88.86{\tiny$\pm$0.40} & 93.11{\tiny$\pm$0.11} & 83.68{\tiny$\pm$0.56} & 91.11{\tiny$\pm$0.04} & 96.90{\tiny$\pm$0.15} & 78.41{\tiny$\pm$0.89} & 88.33{\tiny$\pm$0.14} \\
\midrule
\multicolumn{9}{l}{\textit{Reddit}} \\
all inputs$^\star$ & \textbf{99.15}{\tiny$\pm$0.00} & 82.74{\tiny$\pm$1.44} & \textbf{90.16}{\tiny$\pm$0.17} & \textbf{90.62}{\tiny$\pm$0.27} & \textbf{91.79}{\tiny$\pm$0.13} & \textbf{98.22}{\tiny$\pm$0.02} & 67.54{\tiny$\pm$2.74} & 79.20{\tiny$\pm$0.34} \\
no $c_{uv}$ & 99.13{\tiny$\pm$0.01} & 82.36{\tiny$\pm$0.33} & 90.01{\tiny$\pm$0.03} & 90.55{\tiny$\pm$0.12} & 91.69{\tiny$\pm$0.04} & 98.19{\tiny$\pm$0.02} & 68.83{\tiny$\pm$0.05} & 79.04{\tiny$\pm$0.05} \\
no projections & 99.11{\tiny$\pm$0.01} & \textbf{83.12}{\tiny$\pm$1.18} & 89.96{\tiny$\pm$0.15} & 90.37{\tiny$\pm$0.06} & 91.55{\tiny$\pm$0.06} & 98.13{\tiny$\pm$0.02} & \textbf{69.08}{\tiny$\pm$1.26} & 79.04{\tiny$\pm$0.13} \\
no counts & 98.72{\tiny$\pm$0.03} & 82.77{\tiny$\pm$0.51} & 90.00{\tiny$\pm$0.08} & 89.63{\tiny$\pm$0.18} & 91.12{\tiny$\pm$0.04} & 97.33{\tiny$\pm$0.07} & 65.53{\tiny$\pm$0.89} & \textbf{79.30}{\tiny$\pm$0.13} \\
none of the three & 98.52{\tiny$\pm$0.01} & 82.46{\tiny$\pm$0.16} & 89.82{\tiny$\pm$0.03} & 89.87{\tiny$\pm$0.13} & 91.42{\tiny$\pm$0.06} & 96.57{\tiny$\pm$0.07} & 65.45{\tiny$\pm$0.36} & 78.38{\tiny$\pm$0.38} \\
\midrule
\multicolumn{9}{l}{\textit{UCI}} \\
all inputs$^\star$ & 95.81{\tiny$\pm$0.41} & \textbf{91.47}{\tiny$\pm$0.52} & \textbf{96.04}{\tiny$\pm$0.17} & \textbf{83.35}{\tiny$\pm$0.29} & \textbf{88.18}{\tiny$\pm$0.04} & 94.74{\tiny$\pm$0.12} & 84.08{\tiny$\pm$0.35} & \textbf{89.03}{\tiny$\pm$0.05} \\
no $c_{uv}$ & 95.91{\tiny$\pm$0.25} & 90.63{\tiny$\pm$0.15} & 95.83{\tiny$\pm$0.04} & 83.29{\tiny$\pm$0.16} & 87.80{\tiny$\pm$0.13} & 94.72{\tiny$\pm$0.13} & \textbf{84.23}{\tiny$\pm$0.22} & 88.78{\tiny$\pm$0.06} \\
no projections & \textbf{95.92}{\tiny$\pm$0.23} & 90.26{\tiny$\pm$0.04} & 95.82{\tiny$\pm$0.01} & 82.92{\tiny$\pm$0.48} & 87.78{\tiny$\pm$0.03} & \textbf{94.80}{\tiny$\pm$0.11} & 84.02{\tiny$\pm$0.28} & 88.84{\tiny$\pm$0.13} \\
no counts & 95.24{\tiny$\pm$0.07} & 89.68{\tiny$\pm$0.18} & 95.13{\tiny$\pm$0.17} & 76.19{\tiny$\pm$0.93} & 86.23{\tiny$\pm$0.28} & 92.65{\tiny$\pm$0.01} & 78.04{\tiny$\pm$1.04} & 87.55{\tiny$\pm$0.27} \\
none of the three & 83.09{\tiny$\pm$1.05} & 62.12{\tiny$\pm$0.68} & 91.69{\tiny$\pm$0.29} & 59.73{\tiny$\pm$0.53} & 84.15{\tiny$\pm$0.08} & 74.90{\tiny$\pm$1.27} & 59.66{\tiny$\pm$1.66} & 85.46{\tiny$\pm$0.06} \\
\midrule
\multicolumn{9}{l}{\textit{Enron}} \\
all inputs$^\star$ & 93.09{\tiny$\pm$0.21} & 84.04{\tiny$\pm$2.43} & 90.26{\tiny$\pm$0.77} & 85.32{\tiny$\pm$0.85} & 89.04{\tiny$\pm$0.18} & 89.51{\tiny$\pm$0.10} & 78.56{\tiny$\pm$2.85} & 85.57{\tiny$\pm$0.91} \\
no $c_{uv}$ & \textbf{93.12}{\tiny$\pm$0.05} & \textbf{85.54}{\tiny$\pm$0.28} & \textbf{90.62}{\tiny$\pm$0.09} & \textbf{86.07}{\tiny$\pm$0.23} & \textbf{89.16}{\tiny$\pm$0.23} & \textbf{89.62}{\tiny$\pm$0.21} & \textbf{80.32}{\tiny$\pm$0.25} & \textbf{85.79}{\tiny$\pm$0.47} \\
no projections & 92.67{\tiny$\pm$0.11} & 81.73{\tiny$\pm$1.39} & 89.88{\tiny$\pm$0.11} & 84.35{\tiny$\pm$0.87} & 89.02{\tiny$\pm$0.21} & 89.25{\tiny$\pm$0.21} & 76.04{\tiny$\pm$2.46} & 84.59{\tiny$\pm$0.75} \\
no counts & 91.16{\tiny$\pm$0.19} & 78.87{\tiny$\pm$1.41} & 89.33{\tiny$\pm$0.22} & 78.27{\tiny$\pm$2.16} & 87.00{\tiny$\pm$0.29} & 84.45{\tiny$\pm$0.43} & 70.73{\tiny$\pm$1.82} & 84.21{\tiny$\pm$0.27} \\
none of the three & 87.59{\tiny$\pm$0.28} & 73.89{\tiny$\pm$0.27} & 87.24{\tiny$\pm$0.47} & 77.60{\tiny$\pm$0.74} & 86.50{\tiny$\pm$0.90} & 77.80{\tiny$\pm$0.38} & 64.10{\tiny$\pm$0.62} & 81.23{\tiny$\pm$0.29} \\
\midrule
\multicolumn{9}{l}{\textit{LastFM}} \\
all inputs$^\star$ & 91.87{\tiny$\pm$1.35} & \textbf{84.60}{\tiny$\pm$1.20} & \textbf{95.36}{\tiny$\pm$0.03} & \textbf{73.96}{\tiny$\pm$0.95} & 89.82{\tiny$\pm$0.71} & 91.60{\tiny$\pm$2.72} & \textbf{78.38}{\tiny$\pm$0.70} & 90.12{\tiny$\pm$0.61} \\
no $c_{uv}$ & \textbf{92.86}{\tiny$\pm$0.15} & 82.57{\tiny$\pm$0.64} & 95.26{\tiny$\pm$0.03} & 70.33{\tiny$\pm$0.18} & 90.28{\tiny$\pm$0.09} & \textbf{93.69}{\tiny$\pm$0.11} & 76.59{\tiny$\pm$0.40} & 90.78{\tiny$\pm$0.02} \\
no projections & 92.77{\tiny$\pm$0.05} & 82.34{\tiny$\pm$0.45} & 95.32{\tiny$\pm$0.03} & 71.63{\tiny$\pm$0.47} & \textbf{90.49}{\tiny$\pm$0.11} & 93.47{\tiny$\pm$0.01} & 77.08{\tiny$\pm$0.46} & \textbf{90.86}{\tiny$\pm$0.05} \\
no counts & 87.81{\tiny$\pm$0.77} & 83.87{\tiny$\pm$0.44} & 94.56{\tiny$\pm$0.07} & 73.10{\tiny$\pm$0.99} & 88.35{\tiny$\pm$0.29} & 89.36{\tiny$\pm$0.89} & 78.08{\tiny$\pm$1.11} & 88.75{\tiny$\pm$0.27} \\
none of the three & 81.53{\tiny$\pm$0.28} & 75.33{\tiny$\pm$0.55} & 93.95{\tiny$\pm$0.04} & 67.95{\tiny$\pm$0.43} & 89.38{\tiny$\pm$0.56} & 84.06{\tiny$\pm$0.42} & 70.27{\tiny$\pm$0.96} & 89.18{\tiny$\pm$0.49} \\
\bottomrule
\end{tabular}
\end{table}

\section{Computational Complexity and Efficient Training}
\label{app:computational_efficiency}

Our computational efficiency comes from three properties of the proposed architecture: shared block-start aggregation, closed-form historical states that eliminate temporal backpropagation through time (BPTT), and the reuse of activity statistics for SNAM. This section analyzes the resulting time and memory complexity, including the trade-off between block-start computation and within-block historical updates.

\subsection{Shared Block-Start Computation}
\label{app:block_computation}

Consider a temporal graph with $N$ nodes and $m$ interaction records, partitioned chronologically into $C$ blocks of approximately $B=m/C$ interactions each. Let $a_c$ be the starting time of block $c$, and let $H_c$ denote the number of interactions preceding $a_c$.

For the $k$-th timescale, MTHR constructs the time-decayed adjacency matrix
\[
A_c^k[u,w]
=
\sum_{\substack{
e:\{u_e,v_e\}=\{u,w\}\\
t_e<a_c
}}
e^{-\omega_k(a_c-t_e)}.
\]

All queries within the block share the historical states computed from these matrices:
\begin{equation}
\boxed{
\begin{aligned}
X_{1,c}^k &= A_c^kE,\\
M_c &= \mathcal F_\Theta
\left(
X_{1,c}^1,\ldots,X_{1,c}^K,n_c
\right),\\
X_{2,c}^k &= A_c^kM_c,\\
S_{1,c}^k &= A_c^kR,\\
S_{2,c}^k &= A_c^kS_{1,c}^k.
\end{aligned}
}
\tag{D.1}
\label{eq:app-d1}
\end{equation}

Here, $E\in\mathbb R^{N\times d}$ is the trainable node embedding table, $\mathcal F_\Theta$ denotes first-order normalization and cross-timescale fusion, and $R\in\mathbb R^{N\times D}$ is the fixed random projection matrix. The resulting states provide first- and second-order node representations and structural projections for MTHR and CRM.

Let $P_c$ denote the number of directed node pairs with nonzero historical weights at the beginning of block $c$. A sparse-dense matrix multiplication with $d$-dimensional node representations requires $O(P_cd)$ operations. Across $K$ timescales, the first- and second-order aggregation costs $O(KP_cd)$, while structural random projections require $O(KP_cD)$.

Including the learnable transformations of the node representations, the main block-start computation has complexity
\begin{equation}
\boxed{
T_{\mathrm{state},c}
=
O\left(
KP_c(d+D)+NKd^2
\right).
}
\tag{D.2}
\label{eq:app-d2}
\end{equation}

For $B_c$ positive interactions and an equal number of randomly sampled negatives, approximately $2B_c$ candidate pairs reuse these states. Complete graph aggregation is therefore performed once per block rather than separately for each candidate.

\begin{figure}[p]
\centering
\includegraphics[width=0.94\linewidth]{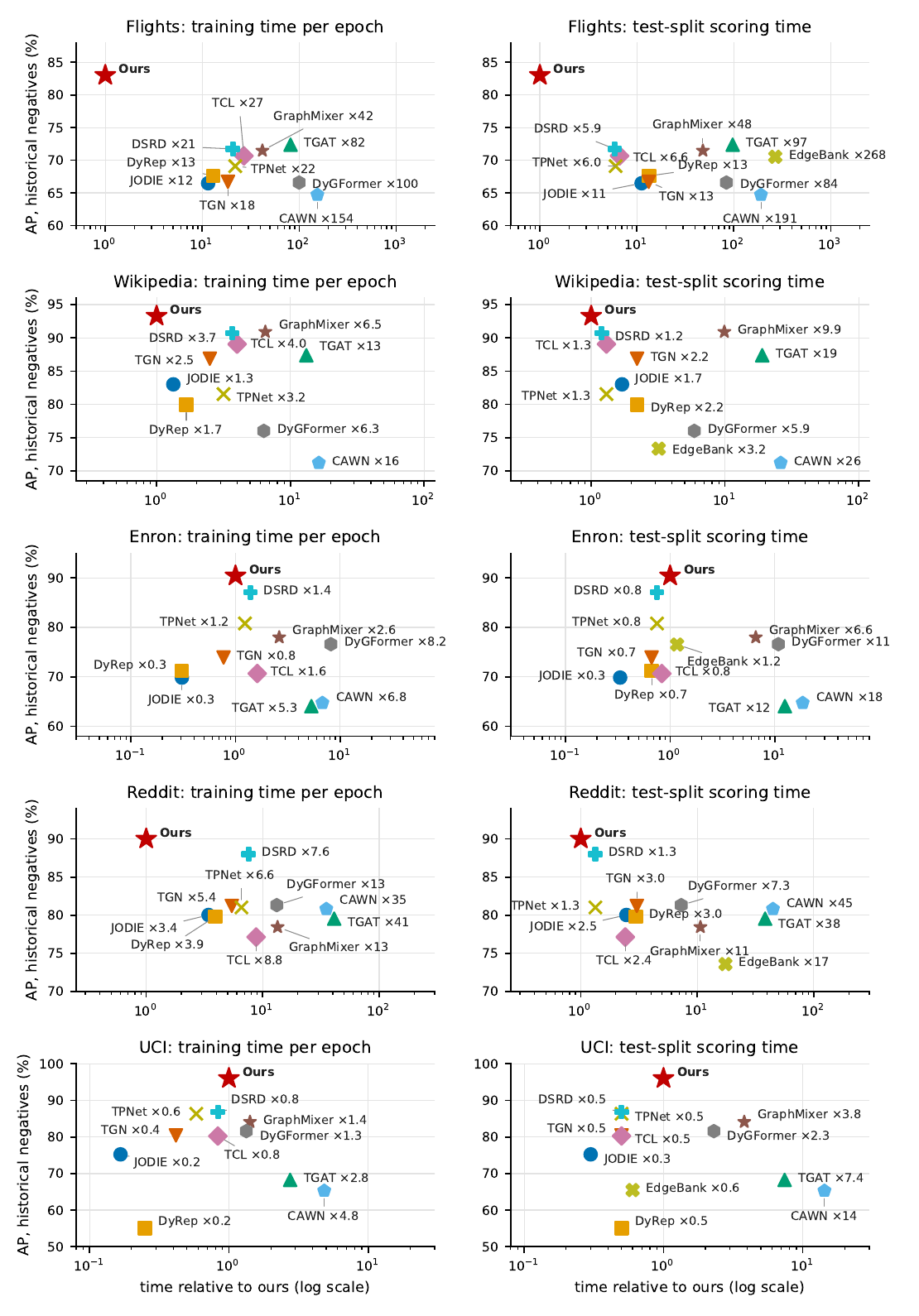}
\caption{Speed on the remaining datasets (Flights, Wikipedia, Enron, Reddit, UCI), one dataset per row. Left: wall-clock time of one training epoch; right: wall-clock time to score the whole test split (random negatives, batches of 200). Both are relative to our model on a log scale; the vertical axis is transductive test AP under historical negative sampling. Every model was run from its released code on the same idle RTX~5090, one process at a time; each number is the median of three runs. For the baselines we timed 300 training batches and 500 scoring batches after a warm-up and scaled to the full epoch or split; our model was timed over the full epoch and the full split. Same conventions as Figure~\ref{fig:speed_lastfm}. Our scoring pass is slower than TPNet's on UCI, the smallest stream, where a fixed per-chunk overhead dominates.}
\label{fig:speed_appendix}
\end{figure}

After each optimizer update, the next block reconstructs its representations using the updated parameters. State sharing is confined to the current block.

\subsection{Query-Time Computation and Within-Block Updates}
\label{app:query_complexity}

For a query $(u,v,t)$ in block $c$, cached historical states are propagated to the query time using the exponential kernel. Under the default \texttt{fresh\_state=0} configuration, the first- and second-order node aggregates are
\[
\widehat X_{1,u}^k(t;c)
=
e^{-\omega_k(t-a_c)}X_{1,c,u}^k,
\]
\[
\widehat X_{2,u}^k(t;c)
=
e^{-\omega_k(t-a_c)}X_{2,c,u}^k.
\]

Within-block interactions are incorporated into the activity counts, edge-feature aggregates, and structural projections. In particular, activity counts are updated exactly:
\begin{equation}
\boxed{
n_u^k(t)
=
e^{-\omega_k(t-a_c)}n_c^k[u]
+
\sum_{\substack{
e\ni u\\
a_c\leq t_e<t
}}
e^{-\omega_k(t-t_e)}.
}
\tag{D.3}
\label{eq:app-d3}
\end{equation}

The second-order random projection uses cached block-start first-order projections as the content of newly observed interactions, avoiding query-wise recomputation of complete two-layer graph propagation.

Query-time computation consists of state retrieval and fusion, processing of within-block records, and candidate-specific feature construction.

Retrieving and decaying the cached node representations requires $O(Kd)$ basic vector operations per node, followed by approximately $O(Kd^2)$ operations for learnable projections and fusion.

Let $r_u(t;c)$ denote the number of within-block historical records accessed for a query. Computing their weights across all timescales costs
\[
O\bigl(Kr_u(t;c)\bigr).
\]
Aggregation of structural projections and edge features introduces additional factors depending on $D$ and $F$, respectively. Repeated interaction partners can be grouped by their accumulated temporal weights before high-dimensional aggregation.

CRM uses low-rank bilinear matching with rank $r$, requiring $O(Krd)$ operations across all timescales. Its four structural matching scores require $O(KD)$ inner-product operations.

The candidate-conditioned recent-interaction encoder processes at most $M$ historical records per query. Each record contributes $K$ temporal weights, $K$ structural matching scores, a learnable representation matching score, and an event-wise MLP output. Its event-processing cost is therefore bounded by the fixed history length $M$.

Exact pairwise interaction counts additionally access the historical records of the candidate pair. Their cost depends on the number of repeated interactions associated with that pair.

Under random destination negative sampling, a positive sample $(u,v,t)$ and its corresponding negative sample $(u,v^-,t)$ share the same source node and query time. Their source-side historical states can therefore be reused, while destination representations and candidate-specific features are computed separately.

\subsection{Gradient Propagation without Temporal BPTT}
\label{app:gradient_propagation}

A central advantage of the proposed historical representation is that gradients propagate through sparse matrix operations rather than an event-wise recurrent computation graph.

Consider a single aggregation layer
\[
X=AE,
\]
where $A$ is the historical weight matrix constructed from observed interactions and $E$ is the trainable node embedding table. Let
\[
G=\frac{\partial\mathcal L}{\partial X}.
\]

Since $\mathrm dX=A\,\mathrm dE$, matrix differentiation gives
\[
\begin{aligned}
\mathrm d\mathcal L
&=
\operatorname{tr}(G^\top\mathrm dX)\\
&=
\operatorname{tr}(G^\top A\,\mathrm dE)\\
&=
\operatorname{tr}
\left(
(A^\top G)^\top\mathrm dE
\right).
\end{aligned}
\]

Therefore,
\begin{equation}
\boxed{
\frac{\partial\mathcal L}{\partial E}
=
A^\top G.
}
\tag{D.4}
\label{eq:app-d4}
\end{equation}

Historical interactions determine the weights in $A$, while gradients reach the embeddings of historical interaction partners through a transposed sparse matrix multiplication.

This property extends to two-layer aggregation. For a single timescale, write
\[
X_1=AE,
\qquad
M=\mathcal F_\Theta(X_1),
\qquad
X_2=AM.
\]

Let $G_1$ and $G_2$ denote the gradients contributed directly to $X_1$ and $X_2$, and let $G_M$ denote the gradient contributed directly to $M$. Backpropagation through the two aggregation layers gives
\begin{equation}
\boxed{
\frac{\partial\mathcal L}{\partial E}
=
A^\top
\left[
G_1+
J_{\mathcal F_\Theta}(X_1)^\top
\left(
G_M+A^\top G_2
\right)
\right]
+
G_E^{\mathrm{direct}}.
}
\tag{D.5}
\label{eq:app-d5}
\end{equation}

Here, $J_{\mathcal F_\Theta}(X_1)$ is the Jacobian of the first-order fusion mapping, and $G_E^{\mathrm{direct}}$ collects other gradient paths that depend directly on the base node embeddings.

For the complete multi-timescale model, define the aggregation operator
\[
\mathcal A_c(E)
=
(A_c^1E,\ldots,A_c^KE)
\]
and its adjoint
\[
\mathcal A_c^\ast(G)
=
\sum_{k=1}^{K}
(A_c^k)^\top G_k.
\]

The corresponding gradient relation becomes
\begin{equation}
\boxed{
\nabla_E\mathcal L
=
\mathcal A_c^\ast
\left[
G_1+
J_{\mathcal F_\Theta}^{\top}
\left(
G_M+\mathcal A_c^\ast(G_2)
\right)
\right]
+
G_E^{\mathrm{direct}}.
}
\tag{D.6}
\label{eq:app-d6}
\end{equation}

The Jacobian-transpose operation accounts for the joint dependence of the intermediate representation on all timescales.

\Eqref{eq:app-d6} shows that the depth of the historical gradient path is determined by the aggregation architecture, rather than the length of the event history. Historical interactions contribute through the fixed adjacency weights, and the corresponding trainable node embeddings receive gradients directly through sparse matrix operations.

By contrast, an event-wise recurrent representation
\[
h_i=F_\Theta(h_{i-1},e_i)
\]
requires the temporal Jacobian product
\[
\frac{\partial h_i}{\partial h_j}
=
\prod_{\ell=j+1}^{i}
\frac{\partial h_\ell}{\partial h_{\ell-1}},
\qquad j<i.
\]

Full BPTT through this recurrence requires an unfolded computation graph whose depth grows with the number of historical updates. MTHR replaces this temporal dependency with a fixed-depth graph aggregation computation.

Each training block builds its computation graph using the current parameters. After its backward pass and optimizer update, the graph can be released before processing the next block. Thus, historical interactions remain available to the current loss without retaining computation graphs from previous blocks.

\subsection{Overall Complexity, Memory Usage, and the Block-Size Trade-off}
\label{app:overall_complexity}

We now combine block-start aggregation and query-time processing to analyze the overall cost.

Let $F$ denote the valid edge-feature dimension and $P$ the number of directed node-pair positions stored in the historical index. For block $c$, let $Q_c$ be the number of scored positive and negative candidates, $R_c$ the total number of within-block records accessed by these queries, and $J_c$ the total number of historical records traversed for exact pairwise counts.

\paragraph{Block-start computation.}

In the current implementation, block-start historical weights are reconstructed from interactions preceding each block. Constructing the adjacency weights and activity counts requires $O(KH_c+KP)$ operations, while historical edge-feature aggregation adds $O(KH_cF)$.

Combining these operations with the graph aggregation cost in \eqref{eq:app-d2} yields
\begin{equation}
\boxed{
T_{\mathrm{entry},c}
=
O\left(
KH_c(1+F)
+
KP(d+D)
+
NKd^2
\right).
}
\tag{D.7}
\label{eq:app-d7}
\end{equation}

Here, $P$ accounts for the indexed sparse structure, including stored node-pair positions whose historical weights may currently be zero.

Under approximately equal-sized blocks,
\[
H_c\approx\frac{cm}{C}.
\]

The number of historical interaction records revisited across one training epoch is therefore approximately
\[
\sum_{c=0}^{C-1}H_c
\approx
\frac{m(C-1)}2.
\]

Consequently, the total cost of reconstructing block-start historical weights scales as
\begin{equation}
\boxed{
T_{\mathrm{history}}
=
O(KmC).
}
\tag{D.8}
\label{eq:app-d8}
\end{equation}

The graph aggregation itself remains shared across all queries within each block.

\paragraph{Query-time computation.}

The fixed-dimensional query operations include node representation fusion, low-rank matching, structural projection matching, and recent-interaction encoding. With fixed $K,d,D,r,M$, their cost per candidate is bounded independently of the total historical sequence length.

The history-dependent operations are within-block record processing and exact pairwise counting. Together, their temporal-weight computations require
\[
O\bigl(K(R_c+J_c)\bigr)
\]
operations per block, with additional vector aggregation costs determined by $D$ and $F$.

The recent-interaction branch processes at most $M$ events per candidate, so its total cost scales with $Q_cM$.

\paragraph{Block-size trade-off.}

Increasing the number of blocks reduces the amount of within-block history available to each query but increases the frequency of block-start reconstruction.

To quantify the first effect, suppose each block contains approximately $m/C$ interactions and the number of scored candidates per block is also of order $m/C$. Since each query can access at most the interactions preceding it within its current block, an upper bound is
\[
R_c
=
O\left(\frac{m^2}{C^2}\right).
\]

Summing over all blocks gives
\begin{equation}
\boxed{
\sum_c R_c
=
O\left(\frac{m^2}{C}\right).
}
\tag{D.9}
\label{eq:app-d9}
\end{equation}

The actual number of accessed records also depends on node activity and the distribution of repeated interactions.

Combining \eqref{eq:app-d8} and \eqref{eq:app-d9}, the two principal history-processing costs have opposing dependence on the block count:
\[
T_{\mathrm{history}}=O(KmC),
\qquad
T_{\mathrm{records}}
=O\left(\frac{Km^2}{C}\right)
\]
for the temporal-weight component of within-block record processing under the stated upper-bound setting.

Finer partitioning reduces within-block record processing but increases block-start reconstruction, graph aggregation, and optimizer-update frequency. The block count therefore controls the balance between shared computation and query-time historical updates.

\begin{figure}[t]
\vspace{-\intextsep}
\centering
\includegraphics[width=\linewidth]{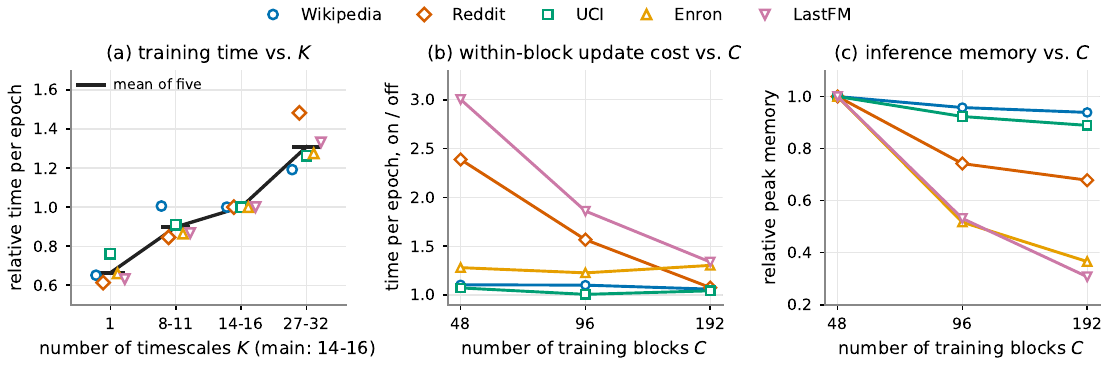}
\caption{What the two structural parameters of the state cost (one RTX~5090, median of three runs, five datasets; the matching accuracies are in Tables~\ref{tab:a1_scales_full} and~\ref{tab:a2_event_clock}). (a) Training time per epoch as the number of time scales $K$ changes, relative to the main configuration ($K$ = 14--16); each point is a dataset, the line is their mean. Doubling $K$ adds about 30\% and a single scale saves about 35\%, so the multi-scale state is nearly free; since accuracy saturates around $K=8$, nothing has to be traded away. (b) The price of the event clock: epoch time with within-chunk records on, divided by the time with them off, as a function of the number of training chunks $C$. Shorter chunks have fewer records to read, so the overhead falls from up to $3\times$ at $C=48$ to 1.1--1.3$\times$ at $C=192$. Rebuilding the chunk-start state costs a fixed 65\,ms per chunk, so the absolute epoch time does not keep falling with $C$: the small streams train fastest at $C=48$, Reddit and LastFM at $C=96$. (c) Peak GPU memory when scoring the test split, relative to $C=48$. Finer chunks lower memory by up to $3\times$ on the large streams without changing accuracy, which makes $T$ a free memory knob: on Reddit, peak memory drops from 3.8\,GB at $C=48$ to 2.6\,GB at $C=192$.}
\label{fig:cost_s3}
\end{figure}

\paragraph{Additional cost of SNAM.}

SNAM reuses the multi-timescale activity counts already maintained by MTHR. Its cumulative intensity is computed using the closed-form integral derived in Appendix~\ref{app:activity_likelihood}.

Let $N_s$ be the number of training source nodes and $I_c$ the number of relevant event-endpoint records within the current observation window. Given the block-start activity counts, computing the cumulative intensity and event log-intensities requires
\begin{equation}
\boxed{
T_{\mathrm{SNAM},c}
=
O\left(
K(N_s+I_c+B_c)
\right).
}
\tag{D.10}
\label{eq:app-d10}
\end{equation}

The three terms correspond to integrating the entry states of all training source nodes, accumulating the contributions of newly observed events, and evaluating the observed-event log-intensities.

SNAM adds only $K+1$ trainable parameters. Its likelihood requires neither additional graph aggregation nor numerical temporal integration, and its gradients are independent of the relation branch.

\paragraph{Memory complexity.}

The model stores the global event index, the node-pair index, trainable node embeddings, and the fixed random projection matrix. Within a block, the main cached states comprise the first- and second-order node aggregates, historical edge-feature aggregates, and first- and second-order random projections.

Their principal storage cost is
\begin{equation}
\boxed{
M_{\mathrm{state}}
=
O\left(
NK(d+D+F)+KP
\right).
}
\tag{D.11}
\label{eq:app-d11}
\end{equation}

Additional memory is required for the current block's candidate representations, historical records, recent-interaction features, and temporary computation tensors.

During training, automatic differentiation retains only the computation graph of the current block. Unlike event-wise recurrent models with full BPTT, it does not need to retain an unfolded computation graph spanning the historical event sequence.

During inference, disabling gradient computation leaves the shared historical states and current query tensors as the main GPU memory components. Finer temporal partitioning reduces the number of queries and within-block records processed at once, while the global indices and cached block-start states remain.

In summary, shared block-start aggregation reduces repeated graph computation, fixed-depth matrix operations eliminate temporal BPTT, and closed-form intensity integration keeps SNAM inexpensive. The remaining costs are governed by block-start reconstruction, within-block historical record processing, and CRM's candidate-conditioned features.
\begin{table}[t]
\caption{Cost as the stream gets longer. We cut LastFM and Flights to their first 10\%, 30\% and 100\% of events, split each prefix 70/15/15 by time on its own, trained the main configuration on it and timed one training epoch and the scoring of its test split (one RTX~5090, median of three runs). The last two columns divide by the amount of work: both times grow at most linearly with the length of the stream, and on the short prefixes the fixed cost per block dominates (LastFM keeps $C=192$ training blocks at every length, so its 10\% prefix pays about 12\,s of block overhead per epoch; Flights uses one block per time step: 5 / 17 / 90 blocks).}
\label{tab:s2_stream_length}
\centering
\footnotesize
\setlength{\tabcolsep}{5pt}
\begin{tabular}{llrrrrrr}
\toprule
stream & prefix & train & test & epoch & s per & scoring & s per \\
 & & edges & positives & time (s) & 100k edges & time (s) & 100k positives \\
\midrule
\multirow{3}{*}{LastFM} & 10\% & 74k & 19k & 12.6 & 17.1 & 1.3 & 6.8 \\
 & 30\% & 222k & 58k & 12.8 & 5.8 & 1.7 & 2.8 \\
 & 100\% & 723k & 194k & 21.7 & 3.0 & 7.1 & 3.7 \\
\midrule
\multirow{3}{*}{Flights} & 10\% & 104k & 24k & 0.5 & 0.5 & 0.5 & 1.9 \\
 & 30\% & 328k & 84k & 2.5 & 0.8 & 0.9 & 1.0 \\
 & 100\% & 1.10M & 288k & 13.8 & 1.2 & 2.9 & 1.0 \\
\bottomrule
\end{tabular}
\end{table}

\section{Datasets and Evaluation Protocols}
\label{app:datasets_protocols}

We evaluate on thirteen interaction streams from the dynamic link prediction benchmark \citep{poursafaei2022towards,yu2023dyglib}: Wikipedia, Reddit, MOOC, LastFM, Enron, Social Evolution, UCI, Flights, Canadian Parliament, US Legislation, UN Trade, UN Vote, and Contact. Table~\ref{tab:datasets} reports the properties of the released data files, including their sizes, available event features, bipartite structure, and temporal spans. We additionally report temporal statistics that directly characterize the information used by our model and the evaluation protocols.

\begin{table}[t]
\caption{The thirteen streams, measured on the released files. Feat. = number of event feature columns that are not identically zero (no stream has node features; the benchmark pads every feature table to 172 columns; -- = none); Bip. = bipartite; span = time between the first and the last event; steps = number of distinct timestamps, in the unit of one time step. Decades = $\log_{10}$ of the ratio between the span of the training events and their smallest positive inter-event gap, the range the time-scale grid covers; $K$ = the number of scales it yields (two per decade, at most 16). Tied = share of events that share their timestamp with another event, the events the strict-before rule keeps out of a query's view. Rep. = share of test-period events whose node pair already occurred earlier in the stream. The spans of MOOC and LastFM differ from the durations given in the benchmark's own table (17 months and 1 month): the timestamps of the released files, in seconds, cover 30 days and 4.3 years.}
\label{tab:datasets}
\centering
\scriptsize
\setlength{\tabcolsep}{1.5pt}
\begin{tabular}{llrrcclrrrrr}
\toprule
Dataset & Domain & \#Nodes & \#Events & Feat. & Bip. & Span & Steps / unit & Decades & $K$ & Tied (\%) & Rep. (\%) \\
\midrule
Wikipedia & Social & 9,227 & 157,474 & 170 & yes & 31 d & 152,757 / s & 6.3 & 14 & 5.9 & 89.2 \\
Reddit & Social & 10,984 & 672,447 & 172 & yes & 31 d & 669,065 / s & 9.3 & 16 & 1.0 & 93.1 \\
MOOC & Interaction & 7,144 & 411,749 & 4 & yes & 30 d & 345,600 / s & 6.3 & 14 & 29.9 & 60.8 \\
LastFM & Interaction & 1,980 & 1,293,103 & -- & yes & 4.3 yr & 1,283,614 / s & 8.0 & 16 & 1.4 & 90.7 \\
Enron & Social & 184 & 125,235 & -- & no & 3.6 yr & 22,632 / s & 7.9 & 16 & 96.4 & 98.3 \\
Social Evo. & Proximity & 74 & 2,099,519 & 2 & no & 242 d & 565,932 / s & 7.2 & 16 & 94.7 & 100.0 \\
UCI & Social & 1,899 & 59,835 & -- & no & 194 d & 58,911 / s & 6.6 & 15 & 2.8 & 82.1 \\
Flights & Transport & 13,169 & 1,927,145 & 1 & no & 121 d & 122 / day & 2.0 & 5 & 100.0 & 88.1 \\
Can. Parl. & Politics & 734 & 74,478 & 1 & no & 13.0 yr & 14 / yr & 1.0 & 3 & 100.0 & 52.4 \\
US Legis. & Politics & 225 & 60,396 & 1 & no & 11 congr. & 12 / congr. & 0.9 & 3 & 100.0 & 77.2 \\
UN Trade & Economics & 255 & 507,497 & 1 & no & 31.0 yr & 32 / yr & 1.4 & 4 & 100.0 & 99.1 \\
UN Vote & Politics & 201 & 1,035,742 & 1 & no & 71.0 yr & 72 / yr & 1.8 & 5 & 100.0 & 100.0 \\
Contact & Proximity & 692 & 2,426,279 & 1 & no & 28 d & 8,064 / 5 min & 3.7 & 9 & 100.0 & 98.1 \\
\bottomrule
\end{tabular}
\end{table}

\subsection{Dataset Statistics and Temporal Structure}
\label{app:dataset_statistics}

The \emph{Decades}, $K$, \emph{Tied}, and \emph{Rep.} columns in Table~\ref{tab:datasets} quantify three aspects of the data that are particularly relevant to our method.

\paragraph{Covered timescale range and the resulting $K$.}
For each dataset, we use training timestamps alone to obtain the smallest positive inter-event gap $\Delta t_{\min}$ and the training span $L$. The \emph{Decades} column reports
\[
\log_{10}\frac{L}{\Delta t_{\min}},
\]
which measures the number of orders of magnitude covered by the multi-timescale state. The number of exponential kernels, $K$, is then determined by the uniform construction in Appendix~\ref{app:timescale_configuration}. Thus, $K$ is the model's structural quantity determined directly by the dataset's temporal range, rather than a hyperparameter tuned separately for each dataset. Reporting both columns also specifies the timescale configurations used across all thirteen datasets.

\paragraph{Fraction of events with tied timestamps.}
The \emph{Tied} column reports the fraction of events whose timestamp is shared by at least one other event. We include this statistic because the history available to a query $(u,v,t)$ is defined strictly by
\[
\mathcal H_{t^-}=\{e:t_e<t\},
\]
not by the position of an event in the data file. Events occurring at $t_e=t$ are excluded even if they precede the current record in file order. For coarsely timestamped streams such as Flights, Canadian Parliament, US Legislation, UN Trade, UN Vote, and Contact, virtually all events have tied timestamps. On these datasets, treating earlier records with the same timestamp as observed history would change the information available to a large fraction of queries. The \emph{Tied} column quantifies precisely how often this distinction arises.

\paragraph{Fraction of repeated test pairs.}
The \emph{Rep.} column is the fraction of positive test events whose node pair $(u,v)$ has occurred at least once before that event. It measures how much of the test stream consists of recurring relationships. For example, the proportion is approximately $100\%$ for Social Evolution and UN Vote, but considerably lower for Canadian Parliament. This statistic helps characterize historical negative sampling: when many positive test events and the sampled negatives both involve previously observed pairs, memorizing whether a pair has ever appeared cannot distinguish them. It also provides a data-level context for the pair-history features introduced in CRM and the corresponding ablation studies.

All temporal spans and timestamp statistics in Table~\ref{tab:datasets} are measured directly from the released files. For MOOC and LastFM, we obtain approximately $30$ days and $4.3$ years, respectively, whereas the benchmark table of \citet{poursafaei2022towards} reports $17$ months and $1$ month. All temporal statistics and timescale grids in our experiments use the timestamps of the released files.

\subsection{Temporal Splits, Inductive Setting, and History Visibility}
\label{app:data_splits_visibility}

Each interaction stream is divided chronologically into $70\%$ training, $15\%$ validation, and $15\%$ test interactions, without randomly shuffling event order \citep{poursafaei2022towards,yu2023dyglib}. We evaluate both transductive and inductive link prediction using the same data partitions across the compared models.

For inductive evaluation, we construct the unseen-node split used by DyGLib. We sample nodes that occur after the training cutoff, with the sampled set comprising $10\%$ of all nodes, and exclude interactions involving these nodes from training. Validation and test interactions involving nodes absent from the resulting training graph form the corresponding inductive evaluation subsets. All compared models use the same node split and evaluation examples.

\begin{table}[t]
\caption{Inductive link prediction, AP (\%), mean{\scriptsize$\pm$std} over seeds (five; DyGFormer$^\dagger$ three, LastFM one) under random (rnd), historical (hist) and inductive (ind) negative sampling. \textbf{Bold}: best; \underline{underline}: second best. $\dagger$: retrained by us with the corrected released code (released configuration).}
\label{tab:ind_ap}
\centering
\resizebox{\textwidth}{!}{%
\renewcommand{\arraystretch}{1.0}\setlength{\tabcolsep}{4pt}%
\begin{tabular}{llccccccccccccc}
\toprule
NSS & Dataset & JODIE & TGN & CAWN & TGAT & GraphMixer & NAT & PINT & DyGFormer & TPNet & GRN$^\dagger$ & DSRD$^\dagger$ & Ours & Ours+$\lambda$ \\
\midrule
\multirow{14}{*}{\rotatebox{90}{rnd}} & Wikipedia & 94.82{\scriptsize$\pm$0.20} & 97.83{\scriptsize$\pm$0.04} & 98.24{\scriptsize$\pm$0.03} & 96.22{\scriptsize$\pm$0.07} & 96.65{\scriptsize$\pm$0.02} & 96.30{\scriptsize$\pm$0.08} & 98.03{\scriptsize$\pm$0.04} & \underline{98.60{\scriptsize$\pm$0.02}}$^\dagger$ & \textbf{98.91{\scriptsize$\pm$0.01}} & 93.93{\scriptsize$\pm$0.26}$^\dagger$ & 96.86{\scriptsize$\pm$0.08}$^\dagger$ & 98.43{\scriptsize$\pm$0.04} & 98.43{\scriptsize$\pm$0.04} \\
 & Reddit & 96.50{\scriptsize$\pm$0.13} & 97.50{\scriptsize$\pm$0.07} & 98.62{\scriptsize$\pm$0.01} & 97.09{\scriptsize$\pm$0.04} & 95.26{\scriptsize$\pm$0.02} & 98.24{\scriptsize$\pm$0.04} & 98.56{\scriptsize$\pm$0.05} & \underline{98.83{\scriptsize$\pm$0.00}}$^\dagger$ & \textbf{98.86{\scriptsize$\pm$0.01}} & 91.13{\scriptsize$\pm$0.31}$^\dagger$ & 98.75{\scriptsize$\pm$0.02}$^\dagger$ & 98.22{\scriptsize$\pm$0.02} & 98.22{\scriptsize$\pm$0.02} \\
 & MOOC & 79.63{\scriptsize$\pm$1.92} & 89.04{\scriptsize$\pm$1.17} & 81.42{\scriptsize$\pm$0.24} & 85.50{\scriptsize$\pm$0.19} & 81.41{\scriptsize$\pm$0.21} & 83.62{\scriptsize$\pm$1.19} & 87.90{\scriptsize$\pm$0.98} & 86.96{\scriptsize$\pm$0.43} & \textbf{95.07{\scriptsize$\pm$0.26}} & 76.35{\scriptsize$\pm$1.41}$^\dagger$ & 81.49{\scriptsize$\pm$0.12}$^\dagger$ & \underline{89.12{\scriptsize$\pm$0.59}} & \underline{89.12{\scriptsize$\pm$0.59}} \\
 & LastFM & 81.61{\scriptsize$\pm$3.82} & 81.45{\scriptsize$\pm$4.29} & 89.42{\scriptsize$\pm$0.07} & 78.63{\scriptsize$\pm$0.31} & 82.11{\scriptsize$\pm$0.42} & 92.24{\scriptsize$\pm$0.93} & 92.42{\scriptsize$\pm$0.64} & \underline{94.20}$^\dagger$ & \textbf{95.36{\scriptsize$\pm$0.11}} & 85.20{\scriptsize$\pm$0.38}$^\dagger$ & 94.14{\scriptsize$\pm$0.41}$^\dagger$ & 92.23{\scriptsize$\pm$2.25} & 92.23{\scriptsize$\pm$2.25} \\
 & Enron & 80.72{\scriptsize$\pm$1.39} & 77.94{\scriptsize$\pm$1.02} & 86.35{\scriptsize$\pm$0.51} & 67.05{\scriptsize$\pm$1.51} & 75.88{\scriptsize$\pm$0.48} & 87.18{\scriptsize$\pm$1.24} & 88.12{\scriptsize$\pm$0.30} & \underline{89.61{\scriptsize$\pm$0.31}}$^\dagger$ & \textbf{90.34{\scriptsize$\pm$0.28}} & 78.45{\scriptsize$\pm$0.90}$^\dagger$ & 87.85{\scriptsize$\pm$0.34}$^\dagger$ & 89.50{\scriptsize$\pm$0.11} & 89.50{\scriptsize$\pm$0.11} \\
 & Social Evo. & 91.96{\scriptsize$\pm$0.48} & 90.77{\scriptsize$\pm$0.86} & 79.94{\scriptsize$\pm$0.18} & 91.41{\scriptsize$\pm$0.16} & 91.86{\scriptsize$\pm$0.06} & 87.44{\scriptsize$\pm$5.48} & 92.40{\scriptsize$\pm$0.60} & \underline{93.14{\scriptsize$\pm$0.04}} & \textbf{93.24{\scriptsize$\pm$0.07}} & 92.35{\scriptsize$\pm$0.42}$^\dagger$ & 90.74{\scriptsize$\pm$0.13}$^\dagger$ & 92.09{\scriptsize$\pm$0.11} & 92.09{\scriptsize$\pm$0.11} \\
 & UCI & 79.86{\scriptsize$\pm$1.48} & 88.12{\scriptsize$\pm$2.05} & 92.73{\scriptsize$\pm$0.06} & 79.54{\scriptsize$\pm$0.48} & 91.19{\scriptsize$\pm$0.42} & 87.31{\scriptsize$\pm$0.28} & 94.72{\scriptsize$\pm$0.15} & 94.56{\scriptsize$\pm$0.03}$^\dagger$ & \textbf{95.74{\scriptsize$\pm$0.05}} & 85.05{\scriptsize$\pm$0.50}$^\dagger$ & 91.60{\scriptsize$\pm$0.46}$^\dagger$ & \underline{94.81{\scriptsize$\pm$0.13}} & \underline{94.81{\scriptsize$\pm$0.13}} \\
 & Flights & 94.74{\scriptsize$\pm$0.37} & 95.03{\scriptsize$\pm$0.60} & 97.06{\scriptsize$\pm$0.02} & 88.73{\scriptsize$\pm$0.33} & 83.03{\scriptsize$\pm$0.05} & 96.74{\scriptsize$\pm$0.22} & 97.54{\scriptsize$\pm$0.06} & \underline{97.79{\scriptsize$\pm$0.02}} & \textbf{97.97{\scriptsize$\pm$0.04}} & 75.05{\scriptsize$\pm$3.15}$^\dagger$ & 82.50{\scriptsize$\pm$0.09}$^\dagger$ & 97.33{\scriptsize$\pm$0.03} & 97.33{\scriptsize$\pm$0.03} \\
 & Can. Parl. & 53.92{\scriptsize$\pm$0.94} & 54.10{\scriptsize$\pm$0.93} & 55.80{\scriptsize$\pm$0.69} & 55.18{\scriptsize$\pm$0.79} & 55.91{\scriptsize$\pm$0.82} & \textbf{61.90{\scriptsize$\pm$2.52}} & 50.32{\scriptsize$\pm$0.86} & 54.46{\scriptsize$\pm$0.65}$^\dagger$ & 57.30{\scriptsize$\pm$0.80}$^\dagger$ & 56.26{\scriptsize$\pm$0.97}$^\dagger$ & 54.16{\scriptsize$\pm$0.48}$^\dagger$ & \underline{57.65{\scriptsize$\pm$1.72}} & \underline{57.65{\scriptsize$\pm$1.72}} \\
 & US Legis. & 54.93{\scriptsize$\pm$2.29} & 58.63{\scriptsize$\pm$0.37} & 53.17{\scriptsize$\pm$1.20} & 51.00{\scriptsize$\pm$3.11} & 50.71{\scriptsize$\pm$0.76} & \textbf{60.41{\scriptsize$\pm$0.74}} & \underline{59.71{\scriptsize$\pm$1.36}} & 54.28{\scriptsize$\pm$2.87} & 47.35{\scriptsize$\pm$1.01}$^\dagger$ & 50.17{\scriptsize$\pm$1.86}$^\dagger$ & 47.20{\scriptsize$\pm$1.86}$^\dagger$ & 48.09{\scriptsize$\pm$1.99} & 47.52{\scriptsize$\pm$1.50} \\
 & UN Trade & 59.65{\scriptsize$\pm$0.77} & 58.31{\scriptsize$\pm$3.15} & 65.24{\scriptsize$\pm$0.21} & 61.03{\scriptsize$\pm$0.18} & 62.17{\scriptsize$\pm$0.31} & \underline{69.57{\scriptsize$\pm$1.45}} & 60.37{\scriptsize$\pm$0.78} & 64.55{\scriptsize$\pm$0.62} & \textbf{73.04{\scriptsize$\pm$1.63}}$^\dagger$ & 54.82{\scriptsize$\pm$2.04}$^\dagger$ & 57.20{\scriptsize$\pm$0.17}$^\dagger$ & 68.28{\scriptsize$\pm$1.82} & 68.28{\scriptsize$\pm$1.82} \\
 & UN Vote & 56.64{\scriptsize$\pm$0.96} & \underline{58.85{\scriptsize$\pm$2.51}} & 49.94{\scriptsize$\pm$0.45} & 52.24{\scriptsize$\pm$1.46} & 50.68{\scriptsize$\pm$0.44} & \textbf{66.60{\scriptsize$\pm$0.98}} & 57.43{\scriptsize$\pm$1.24} & 55.93{\scriptsize$\pm$0.39} & 58.60{\scriptsize$\pm$3.88}$^\dagger$ & 57.00{\scriptsize$\pm$2.66}$^\dagger$ & 51.07{\scriptsize$\pm$1.90}$^\dagger$ & 57.27{\scriptsize$\pm$2.59} & 57.15{\scriptsize$\pm$2.49} \\
 & Contact & 94.34{\scriptsize$\pm$1.45} & 93.82{\scriptsize$\pm$0.99} & 89.55{\scriptsize$\pm$0.30} & 95.87{\scriptsize$\pm$0.11} & 90.59{\scriptsize$\pm$0.05} & 96.12{\scriptsize$\pm$0.08} & 97.41{\scriptsize$\pm$0.14} & \underline{98.03{\scriptsize$\pm$0.02}} & \textbf{98.39{\scriptsize$\pm$0.02}} & 89.27{\scriptsize$\pm$0.93}$^\dagger$ & 90.31{\scriptsize$\pm$0.08}$^\dagger$ & 97.47{\scriptsize$\pm$0.03} & 97.47{\scriptsize$\pm$0.03} \\
\cmidrule{2-15}
 & Avg. Rank & 9.62 & 8.15 & 7.92 & 9.69 & 9.54 & 5.85 & 5.23 & \underline{4.15} & \textbf{2.23} & 10.23 & 8.92 & 4.62 & 4.85 \\
\midrule
\multirow{14}{*}{\rotatebox{90}{hist}} & Wikipedia & 68.69{\scriptsize$\pm$0.39} & 81.76{\scriptsize$\pm$0.32} & 67.27{\scriptsize$\pm$1.63} & 84.17{\scriptsize$\pm$0.22} & 87.60{\scriptsize$\pm$0.30} & 47.37{\scriptsize$\pm$4.39} & 78.22{\scriptsize$\pm$2.90} & 63.40{\scriptsize$\pm$1.23}$^\dagger$ & 71.28{\scriptsize$\pm$4.33} & \underline{88.27{\scriptsize$\pm$0.24}}$^\dagger$ & 86.57{\scriptsize$\pm$0.51}$^\dagger$ & 78.58{\scriptsize$\pm$1.80} & \textbf{88.34{\scriptsize$\pm$0.12}} \\
 & Reddit & 62.34{\scriptsize$\pm$0.54} & 64.85{\scriptsize$\pm$0.85} & 63.67{\scriptsize$\pm$0.41} & 63.47{\scriptsize$\pm$0.36} & 64.50{\scriptsize$\pm$0.26} & 61.51{\scriptsize$\pm$0.73} & 67.56{\scriptsize$\pm$0.83} & 64.83{\scriptsize$\pm$2.14}$^\dagger$ & 62.15{\scriptsize$\pm$1.72} & 61.26{\scriptsize$\pm$0.61}$^\dagger$ & \textbf{80.28{\scriptsize$\pm$1.04}}$^\dagger$ & 66.49{\scriptsize$\pm$2.49} & \underline{79.06{\scriptsize$\pm$0.32}} \\
 & MOOC & 63.22{\scriptsize$\pm$1.55} & 77.07{\scriptsize$\pm$3.41} & 74.68{\scriptsize$\pm$0.68} & 76.73{\scriptsize$\pm$0.29} & 74.00{\scriptsize$\pm$0.97} & 69.30{\scriptsize$\pm$0.95} & 73.74{\scriptsize$\pm$1.66} & 80.82{\scriptsize$\pm$0.30} & \underline{81.85{\scriptsize$\pm$1.60}} & 63.90{\scriptsize$\pm$0.95}$^\dagger$ & 74.35{\scriptsize$\pm$0.80}$^\dagger$ & 79.69{\scriptsize$\pm$1.56} & \textbf{95.54{\scriptsize$\pm$0.33}} \\
 & LastFM & 70.39{\scriptsize$\pm$4.31} & 66.65{\scriptsize$\pm$6.11} & 71.33{\scriptsize$\pm$0.47} & 76.27{\scriptsize$\pm$0.25} & 76.42{\scriptsize$\pm$0.22} & 70.21{\scriptsize$\pm$0.78} & 77.96{\scriptsize$\pm$1.55} & 76.57$^\dagger$ & 82.27{\scriptsize$\pm$1.22} & 76.17{\scriptsize$\pm$0.28}$^\dagger$ & \underline{86.40{\scriptsize$\pm$2.42}}$^\dagger$ & 78.58{\scriptsize$\pm$0.60} & \textbf{89.84{\scriptsize$\pm$0.63}} \\
 & Enron & 65.86{\scriptsize$\pm$3.71} & 62.91{\scriptsize$\pm$1.16} & 60.70{\scriptsize$\pm$0.36} & 61.40{\scriptsize$\pm$1.31} & 72.37{\scriptsize$\pm$1.37} & 62.69{\scriptsize$\pm$1.02} & 80.47{\scriptsize$\pm$1.52} & 66.13{\scriptsize$\pm$0.49}$^\dagger$ & 74.60{\scriptsize$\pm$1.35} & 69.86{\scriptsize$\pm$0.87}$^\dagger$ & \underline{83.75{\scriptsize$\pm$0.48}}$^\dagger$ & 78.68{\scriptsize$\pm$2.27} & \textbf{85.55{\scriptsize$\pm$0.72}} \\
 & Social Evo. & 88.51{\scriptsize$\pm$0.87} & 90.66{\scriptsize$\pm$1.62} & 79.83{\scriptsize$\pm$0.38} & 93.97{\scriptsize$\pm$0.54} & 94.01{\scriptsize$\pm$0.47} & 84.89{\scriptsize$\pm$4.62} & 95.75{\scriptsize$\pm$1.06} & \underline{96.82{\scriptsize$\pm$0.16}} & 96.38{\scriptsize$\pm$0.18} & 85.86{\scriptsize$\pm$0.47}$^\dagger$ & 92.84{\scriptsize$\pm$0.39}$^\dagger$ & 96.32{\scriptsize$\pm$0.15} & \textbf{98.44{\scriptsize$\pm$0.04}} \\
 & UCI & 63.11{\scriptsize$\pm$2.27} & 70.78{\scriptsize$\pm$0.78} & 64.54{\scriptsize$\pm$0.47} & 70.52{\scriptsize$\pm$0.93} & 81.66{\scriptsize$\pm$0.49} & 51.56{\scriptsize$\pm$0.75} & \underline{85.49{\scriptsize$\pm$0.22}} & 71.73{\scriptsize$\pm$0.77}$^\dagger$ & 78.48{\scriptsize$\pm$1.18} & 78.16{\scriptsize$\pm$0.33}$^\dagger$ & 82.59{\scriptsize$\pm$0.30}$^\dagger$ & 84.10{\scriptsize$\pm$0.38} & \textbf{89.05{\scriptsize$\pm$0.18}} \\
 & Flights & 61.01{\scriptsize$\pm$1.65} & 59.31{\scriptsize$\pm$1.43} & 56.82{\scriptsize$\pm$0.57} & 64.72{\scriptsize$\pm$0.36} & \underline{65.28{\scriptsize$\pm$0.24}} & 51.63{\scriptsize$\pm$1.10} & 53.46{\scriptsize$\pm$0.61} & 57.11{\scriptsize$\pm$0.21} & 54.67{\scriptsize$\pm$0.76} & 57.34{\scriptsize$\pm$1.91}$^\dagger$ & 65.26{\scriptsize$\pm$0.21}$^\dagger$ & 59.12{\scriptsize$\pm$1.21} & \textbf{70.31{\scriptsize$\pm$0.38}} \\
 & Can. Parl. & 52.60{\scriptsize$\pm$0.88} & 54.42{\scriptsize$\pm$0.77} & 57.14{\scriptsize$\pm$0.07} & 56.72{\scriptsize$\pm$0.47} & 55.84{\scriptsize$\pm$0.73} & \textbf{61.56{\scriptsize$\pm$2.68}} & 50.61{\scriptsize$\pm$1.76} & 55.47{\scriptsize$\pm$0.68}$^\dagger$ & 57.31{\scriptsize$\pm$0.52}$^\dagger$ & 55.83{\scriptsize$\pm$1.66}$^\dagger$ & 53.76{\scriptsize$\pm$0.64}$^\dagger$ & \underline{57.99{\scriptsize$\pm$1.98}} & 57.54{\scriptsize$\pm$1.96} \\
 & US Legis. & 52.94{\scriptsize$\pm$2.11} & \underline{61.18{\scriptsize$\pm$1.10}} & 55.56{\scriptsize$\pm$1.71} & 51.83{\scriptsize$\pm$3.95} & 52.03{\scriptsize$\pm$1.02} & \textbf{64.61{\scriptsize$\pm$3.02}} & 59.37{\scriptsize$\pm$1.84} & 56.31{\scriptsize$\pm$3.46} & 48.39{\scriptsize$\pm$1.07}$^\dagger$ & 50.05{\scriptsize$\pm$2.13}$^\dagger$ & 45.69{\scriptsize$\pm$2.67}$^\dagger$ & 49.93{\scriptsize$\pm$3.71} & 49.93{\scriptsize$\pm$3.71} \\
 & UN Trade & 55.46{\scriptsize$\pm$1.19} & 52.80{\scriptsize$\pm$3.19} & 55.00{\scriptsize$\pm$0.38} & 55.28{\scriptsize$\pm$0.71} & 54.94{\scriptsize$\pm$0.97} & \underline{70.04{\scriptsize$\pm$2.07}} & 57.35{\scriptsize$\pm$1.65} & 53.20{\scriptsize$\pm$1.07} & 53.42{\scriptsize$\pm$0.88}$^\dagger$ & 51.79{\scriptsize$\pm$1.48}$^\dagger$ & 54.00{\scriptsize$\pm$0.69}$^\dagger$ & 57.66{\scriptsize$\pm$2.35} & \textbf{73.07{\scriptsize$\pm$1.77}} \\
 & UN Vote & 61.04{\scriptsize$\pm$1.30} & 63.74{\scriptsize$\pm$3.00} & 47.98{\scriptsize$\pm$0.84} & 53.05{\scriptsize$\pm$3.10} & 48.09{\scriptsize$\pm$0.43} & \underline{64.91{\scriptsize$\pm$1.58}} & \textbf{65.75{\scriptsize$\pm$2.86}} & 52.63{\scriptsize$\pm$1.26} & 51.20{\scriptsize$\pm$1.44}$^\dagger$ & 56.52{\scriptsize$\pm$3.62}$^\dagger$ & 48.83{\scriptsize$\pm$3.56}$^\dagger$ & 57.68{\scriptsize$\pm$1.71} & 57.66{\scriptsize$\pm$1.73} \\
 & Contact & 90.42{\scriptsize$\pm$2.34} & 88.13{\scriptsize$\pm$1.50} & 74.20{\scriptsize$\pm$0.80} & \underline{94.15{\scriptsize$\pm$0.45}} & 89.91{\scriptsize$\pm$0.36} & 84.13{\scriptsize$\pm$1.78} & 90.68{\scriptsize$\pm$0.46} & 93.56{\scriptsize$\pm$0.52} & 93.56{\scriptsize$\pm$0.57} & 74.69{\scriptsize$\pm$1.23}$^\dagger$ & 89.22{\scriptsize$\pm$0.56}$^\dagger$ & 93.18{\scriptsize$\pm$0.55} & \textbf{96.62{\scriptsize$\pm$0.17}} \\
\cmidrule{2-15}
 & Avg. Rank & 8.77 & 7.69 & 9.69 & 7.00 & 6.69 & 8.77 & 5.77 & 6.96 & 6.88 & 9.08 & 6.54 & \underline{4.81} & \textbf{2.35} \\
\midrule
\multirow{14}{*}{\rotatebox{90}{ind}} & Wikipedia & 68.70{\scriptsize$\pm$0.39} & 81.77{\scriptsize$\pm$0.32} & 67.24{\scriptsize$\pm$1.63} & 84.17{\scriptsize$\pm$0.22} & 87.60{\scriptsize$\pm$0.29} & 47.37{\scriptsize$\pm$4.39} & 78.23{\scriptsize$\pm$2.89} & 63.40{\scriptsize$\pm$1.23}$^\dagger$ & 71.29{\scriptsize$\pm$4.33} & \underline{88.28{\scriptsize$\pm$0.24}}$^\dagger$ & 86.58{\scriptsize$\pm$0.51}$^\dagger$ & 78.60{\scriptsize$\pm$1.80} & \textbf{88.35{\scriptsize$\pm$0.12}} \\
 & Reddit & 62.32{\scriptsize$\pm$0.54} & 64.84{\scriptsize$\pm$0.84} & 63.65{\scriptsize$\pm$0.41} & 63.40{\scriptsize$\pm$0.36} & 64.49{\scriptsize$\pm$0.25} & 61.51{\scriptsize$\pm$0.73} & 67.56{\scriptsize$\pm$0.83} & 64.83{\scriptsize$\pm$2.14}$^\dagger$ & 62.14{\scriptsize$\pm$1.72} & 61.26{\scriptsize$\pm$0.61}$^\dagger$ & \textbf{80.27{\scriptsize$\pm$1.04}}$^\dagger$ & 66.49{\scriptsize$\pm$2.49} & \underline{79.06{\scriptsize$\pm$0.32}} \\
 & MOOC & 63.22{\scriptsize$\pm$1.55} & 77.07{\scriptsize$\pm$3.40} & 74.69{\scriptsize$\pm$0.68} & 76.72{\scriptsize$\pm$0.30} & 73.99{\scriptsize$\pm$0.97} & 69.30{\scriptsize$\pm$0.95} & 73.74{\scriptsize$\pm$1.66} & 80.82{\scriptsize$\pm$0.30} & \underline{81.85{\scriptsize$\pm$1.60}} & 63.90{\scriptsize$\pm$0.95}$^\dagger$ & 74.35{\scriptsize$\pm$0.80}$^\dagger$ & 79.69{\scriptsize$\pm$1.56} & \textbf{95.54{\scriptsize$\pm$0.33}} \\
 & LastFM & 70.39{\scriptsize$\pm$4.31} & 69.46{\scriptsize$\pm$4.65} & 71.33{\scriptsize$\pm$0.47} & 76.28{\scriptsize$\pm$0.25} & 76.42{\scriptsize$\pm$0.22} & 70.21{\scriptsize$\pm$0.78} & 77.96{\scriptsize$\pm$1.55} & 76.57$^\dagger$ & 82.27{\scriptsize$\pm$1.22} & 76.18{\scriptsize$\pm$0.28}$^\dagger$ & \underline{86.40{\scriptsize$\pm$2.42}}$^\dagger$ & 78.58{\scriptsize$\pm$0.60} & \textbf{89.84{\scriptsize$\pm$0.63}} \\
 & Enron & 65.86{\scriptsize$\pm$3.71} & 62.90{\scriptsize$\pm$1.16} & 60.72{\scriptsize$\pm$0.36} & 61.40{\scriptsize$\pm$1.30} & 72.37{\scriptsize$\pm$1.38} & 62.69{\scriptsize$\pm$1.02} & 80.47{\scriptsize$\pm$1.52} & 66.13{\scriptsize$\pm$0.49}$^\dagger$ & 74.60{\scriptsize$\pm$1.35} & 69.85{\scriptsize$\pm$0.87}$^\dagger$ & \underline{83.75{\scriptsize$\pm$0.48}}$^\dagger$ & 78.65{\scriptsize$\pm$2.27} & \textbf{85.55{\scriptsize$\pm$0.72}} \\
 & Social Evo. & 88.51{\scriptsize$\pm$0.87} & 90.65{\scriptsize$\pm$1.62} & 79.83{\scriptsize$\pm$0.39} & 93.97{\scriptsize$\pm$0.54} & 94.01{\scriptsize$\pm$0.47} & 84.89{\scriptsize$\pm$4.62} & 95.75{\scriptsize$\pm$1.06} & \underline{96.82{\scriptsize$\pm$0.17}} & 96.38{\scriptsize$\pm$0.18} & 85.87{\scriptsize$\pm$0.47}$^\dagger$ & 92.84{\scriptsize$\pm$0.39}$^\dagger$ & 96.31{\scriptsize$\pm$0.15} & \textbf{98.44{\scriptsize$\pm$0.04}} \\
 & UCI & 63.16{\scriptsize$\pm$2.27} & 70.73{\scriptsize$\pm$0.79} & 64.54{\scriptsize$\pm$0.47} & 70.49{\scriptsize$\pm$0.93} & 81.64{\scriptsize$\pm$0.49} & 51.58{\scriptsize$\pm$0.76} & \underline{85.50{\scriptsize$\pm$0.22}} & 71.76{\scriptsize$\pm$0.75}$^\dagger$ & 78.50{\scriptsize$\pm$1.18} & 78.25{\scriptsize$\pm$0.34}$^\dagger$ & 82.61{\scriptsize$\pm$0.30}$^\dagger$ & 84.10{\scriptsize$\pm$0.39} & \textbf{89.04{\scriptsize$\pm$0.18}} \\
 & Flights & 61.01{\scriptsize$\pm$1.66} & 59.32{\scriptsize$\pm$1.45} & 56.82{\scriptsize$\pm$0.56} & 64.72{\scriptsize$\pm$0.37} & \underline{65.29{\scriptsize$\pm$0.24}} & 51.60{\scriptsize$\pm$1.11} & 53.43{\scriptsize$\pm$0.61} & 57.11{\scriptsize$\pm$0.20} & 54.63{\scriptsize$\pm$0.75} & 57.35{\scriptsize$\pm$1.92}$^\dagger$ & 65.25{\scriptsize$\pm$0.20}$^\dagger$ & 59.07{\scriptsize$\pm$1.21} & \textbf{70.29{\scriptsize$\pm$0.38}} \\
 & Can. Parl. & 52.58{\scriptsize$\pm$0.86} & 54.18{\scriptsize$\pm$0.73} & 57.06{\scriptsize$\pm$0.08} & 56.46{\scriptsize$\pm$0.50} & 55.76{\scriptsize$\pm$0.65} & \textbf{61.71{\scriptsize$\pm$2.77}} & 50.66{\scriptsize$\pm$1.67} & 55.57{\scriptsize$\pm$0.77}$^\dagger$ & 57.37{\scriptsize$\pm$0.56}$^\dagger$ & 55.68{\scriptsize$\pm$1.77}$^\dagger$ & 54.03{\scriptsize$\pm$0.63}$^\dagger$ & \underline{57.92{\scriptsize$\pm$1.84}} & 57.50{\scriptsize$\pm$1.80} \\
 & US Legis. & 52.94{\scriptsize$\pm$2.11} & \underline{61.18{\scriptsize$\pm$1.10}} & 55.56{\scriptsize$\pm$1.71} & 51.83{\scriptsize$\pm$3.95} & 52.03{\scriptsize$\pm$1.02} & \textbf{64.61{\scriptsize$\pm$3.02}} & 59.37{\scriptsize$\pm$1.84} & 56.31{\scriptsize$\pm$3.46} & 48.39{\scriptsize$\pm$1.07}$^\dagger$ & 50.05{\scriptsize$\pm$2.13}$^\dagger$ & 45.69{\scriptsize$\pm$2.67}$^\dagger$ & 49.93{\scriptsize$\pm$3.71} & 49.93{\scriptsize$\pm$3.71} \\
 & UN Trade & 55.43{\scriptsize$\pm$1.20} & 52.80{\scriptsize$\pm$3.24} & 54.97{\scriptsize$\pm$0.38} & 55.58{\scriptsize$\pm$0.68} & 54.88{\scriptsize$\pm$1.01} & \underline{69.96{\scriptsize$\pm$2.08}} & 57.26{\scriptsize$\pm$1.68} & 52.56{\scriptsize$\pm$1.70} & 53.48{\scriptsize$\pm$0.89}$^\dagger$ & 51.82{\scriptsize$\pm$1.48}$^\dagger$ & 54.02{\scriptsize$\pm$0.69}$^\dagger$ & 57.77{\scriptsize$\pm$2.36} & \textbf{73.12{\scriptsize$\pm$1.75}} \\
 & UN Vote & 61.17{\scriptsize$\pm$1.33} & 63.71{\scriptsize$\pm$2.97} & 48.01{\scriptsize$\pm$0.82} & 53.08{\scriptsize$\pm$3.10} & 48.10{\scriptsize$\pm$0.40} & \underline{64.81{\scriptsize$\pm$1.54}} & \textbf{65.70{\scriptsize$\pm$2.82}} & 52.61{\scriptsize$\pm$1.25} & 51.19{\scriptsize$\pm$1.43}$^\dagger$ & 56.55{\scriptsize$\pm$3.65}$^\dagger$ & 48.90{\scriptsize$\pm$3.57}$^\dagger$ & 57.65{\scriptsize$\pm$1.69} & 57.62{\scriptsize$\pm$1.72} \\
 & Contact & 90.43{\scriptsize$\pm$2.33} & 88.12{\scriptsize$\pm$1.50} & 74.19{\scriptsize$\pm$0.81} & \underline{94.14{\scriptsize$\pm$0.45}} & 89.91{\scriptsize$\pm$0.36} & 84.13{\scriptsize$\pm$1.77} & 90.68{\scriptsize$\pm$0.46} & 93.55{\scriptsize$\pm$0.52} & 93.57{\scriptsize$\pm$0.57} & 74.70{\scriptsize$\pm$1.22}$^\dagger$ & 89.22{\scriptsize$\pm$0.56}$^\dagger$ & 93.19{\scriptsize$\pm$0.55} & \textbf{96.63{\scriptsize$\pm$0.17}} \\
\cmidrule{2-15}
 & Avg. Rank & 8.85 & 7.62 & 9.69 & 6.92 & 6.69 & 8.77 & 5.77 & 7.08 & 6.85 & 9.08 & 6.54 & \underline{4.81} & \textbf{2.35} \\
\midrule
\multicolumn{2}{l}{Overall Avg. Rank} & 9.08 & 7.82 & 9.10 & 7.87 & 7.64 & 7.79 & 5.59 & 6.06 & 5.32 & 9.46 & 7.33 & \underline{4.74} & \textbf{3.18} \\
\bottomrule
\end{tabular}}
\end{table}

Evaluation follows a streaming protocol: an interaction observed earlier in validation or test may enter the history of a subsequent query. History visibility is governed by event time rather than the training, validation, or test label. For a query at time $t$, interactions with $t_e<t$ are available, while all interactions with $t_e=t$ remain unavailable, irrespective of their file order. We enforce this rule for MTHR adjacency matrices and counts, CRM pair histories and recent interactions, and SNAM activity statistics. The \emph{Tied} column in Table~\ref{tab:datasets} measures the prevalence of precisely those timestamp ties for which file-order updates would differ from this rule.

\subsection{Negative Sampling and Batch-Level Evaluation}
\label{app:negative_sampling_protocols}

During training, each observed interaction $(u,v,t)$ is paired with one random destination negative. During evaluation, we consider the random, historical, and inductive negative-sampling protocols of \citet{poursafaei2022towards}, using the DyGLib samplers \citep{yu2023dyglib}. The latter two protocols replace easy, arbitrarily sampled non-edges with previously observed relationships and test whether a model can identify when such relationships occur or recur. Their distinction from random destination sampling is also central to the statistical analysis in Appendix~\ref{app:activity_motivation}.

Let $[b_i,e_i]$ be the timestamp range of evaluation batch $i$. Let $\mathcal P_i$ contain the pairs occurring in the sampler's interaction stream within this interval, and let $\mathcal P_{\leq b_i}$ contain the pairs with timestamps at or before $b_i$. Finally, let $\mathcal P_{\leq t_0}$ contain the pairs observed by the cutoff $t_0$ preceding the current evaluation period. The sampler operates on the complete evaluation stream in the transductive setting and on the corresponding new-node subset in the inductive setting.

\paragraph{Random negative sampling.}
For each positive interaction $(u,v,t)$, the random sampler preserves its source and timestamp and replaces only the destination:
\[
(u,v,t)\longmapsto (u,v^-,t).
\]
Each positive interaction and its matched random negative therefore share the same $(u,t)$.

\paragraph{Historical negative sampling.}
Historical negatives are drawn from node pairs that have appeared by the start of the current evaluation batch but are absent from its positive-pair set. Their candidate set is
\[
\mathcal C_i^{\mathrm{hist}}
=
\mathcal P_{\leq b_i}\setminus\mathcal P_i.
\]
Sampling is performed without replacement when sufficient candidates are available. The sampled negative pairs are assigned the query times of the corresponding positive examples. The task is consequently to determine whether a relationship that genuinely occurred in the observed history recurs at the current query, rather than merely detecting whether the pair has ever existed. If fewer historical candidates than required are available, the remaining negatives are drawn randomly from node pairs excluded from the current positive batch.

\paragraph{Inductive negative sampling.}
The inductive sampler selects pairs that have appeared before the current batch but were not observed before the current evaluation period. For validation, $t_0$ is the end of training; for testing, it is the end of validation. After excluding the current batch's positive pairs, the candidate set is
\[
\mathcal C_i^{\mathrm{ind}}
=
\mathcal P_{\leq b_i}
\setminus
\bigl(\mathcal P_{\leq t_0}\cup\mathcal P_i\bigr).
\]
Thus, inductive negatives represent relationships first observed during the current evaluation period that can subsequently be sampled at queries where they are not positive. This protocol tests when newly encountered relationships recur, rather than simply distinguishing pairs seen in training from pairs absent from training. As in historical sampling, any shortfall in the candidate pool is filled by random negative pairs that do not occur in the current positive batch.

The two harder protocols sample complete node pairs, whereas random destination sampling retains each positive example's source. Historical and inductive negatives may therefore have different source nodes from their matched positives. This changes the negative distribution in precisely the way analyzed in Appendix~\ref{app:activity_motivation}: source activity cancels for a matched random positive-negative pair, but it can contribute when comparing complete pairs with different source nodes.

We report Average Precision (AP) and ROC-AUC. Each evaluation batch contains at most $200$ positive interactions and an equal number of negatives. Both metrics are computed over all positive and negative examples in each batch and then averaged across batches, rather than being computed once from scores pooled over the entire test split \citep{yu2023dyglib}.

\begin{table}[t]
\caption{Transductive link prediction, AUC (\%), mean{\scriptsize$\pm$std} over seeds (five; DyGFormer$^\dagger$ three, LastFM one) under random (rnd), historical (hist) and inductive (ind) negative sampling. \textbf{Bold}: best; \underline{underline}: second best. $\dagger$: retrained by us with the corrected released code (released configuration).}
\label{tab:trans_auc}
\centering
\resizebox{\textwidth}{!}{%
\renewcommand{\arraystretch}{1.0}\setlength{\tabcolsep}{4pt}%
\begin{tabular}{llccccccccccccc}
\toprule
NSS & Dataset & JODIE & TGN & CAWN & EdgeBank & GraphMixer & NAT & PINT & DyGFormer & TPNet & GRN$^\dagger$ & DSRD$^\dagger$ & Ours & Ours+$\lambda$ \\
\midrule
\multirow{14}{*}{\rotatebox{90}{rnd}} & Wikipedia & 96.33{\scriptsize$\pm$0.07} & 98.37{\scriptsize$\pm$0.07} & 98.54{\scriptsize$\pm$0.04} & 90.78{\scriptsize$\pm$0.00} & 96.92{\scriptsize$\pm$0.03} & 97.75{\scriptsize$\pm$0.11} & 98.16{\scriptsize$\pm$0.06} & \underline{98.94{\scriptsize$\pm$0.00}}$^\dagger$ & \textbf{99.30{\scriptsize$\pm$0.02}} & 93.77{\scriptsize$\pm$0.23}$^\dagger$ & 96.92{\scriptsize$\pm$0.03}$^\dagger$ & 98.80{\scriptsize$\pm$0.01} & 98.80{\scriptsize$\pm$0.01} \\
 & Reddit & 98.31{\scriptsize$\pm$0.05} & 98.60{\scriptsize$\pm$0.06} & 99.01{\scriptsize$\pm$0.01} & 95.37{\scriptsize$\pm$0.00} & 97.17{\scriptsize$\pm$0.02} & 99.09{\scriptsize$\pm$0.10} & 99.09{\scriptsize$\pm$0.03} & 99.15{\scriptsize$\pm$0.00}$^\dagger$ & \underline{99.22{\scriptsize$\pm$0.00}} & 92.96{\scriptsize$\pm$0.22}$^\dagger$ & \textbf{99.29{\scriptsize$\pm$0.01}}$^\dagger$ & 99.05{\scriptsize$\pm$0.00} & 99.05{\scriptsize$\pm$0.00} \\
 & MOOC & 83.81{\scriptsize$\pm$2.09} & \underline{91.21{\scriptsize$\pm$1.15}} & 80.38{\scriptsize$\pm$0.26} & 60.86{\scriptsize$\pm$0.00} & 84.01{\scriptsize$\pm$0.17} & 87.42{\scriptsize$\pm$0.58} & 90.55{\scriptsize$\pm$0.43} & 87.91{\scriptsize$\pm$0.58} & \textbf{97.17{\scriptsize$\pm$0.08}} & 76.49{\scriptsize$\pm$1.06}$^\dagger$ & 84.28{\scriptsize$\pm$0.08}$^\dagger$ & 90.68{\scriptsize$\pm$0.62} & 90.68{\scriptsize$\pm$0.62} \\
 & LastFM & 70.49{\scriptsize$\pm$1.66} & 78.47{\scriptsize$\pm$2.94} & 85.92{\scriptsize$\pm$0.10} & 83.77{\scriptsize$\pm$0.00} & 73.53{\scriptsize$\pm$0.12} & 86.92{\scriptsize$\pm$2.72} & 89.28{\scriptsize$\pm$1.63} & \underline{93.03}$^\dagger$ & \textbf{94.39{\scriptsize$\pm$0.04}} & 74.71{\scriptsize$\pm$0.35}$^\dagger$ & 91.48{\scriptsize$\pm$0.26}$^\dagger$ & 92.77{\scriptsize$\pm$0.94} & 92.77{\scriptsize$\pm$0.94} \\
 & Enron & 87.96{\scriptsize$\pm$0.52} & 88.32{\scriptsize$\pm$0.99} & 90.45{\scriptsize$\pm$0.14} & 87.05{\scriptsize$\pm$0.00} & 84.38{\scriptsize$\pm$0.21} & 91.68{\scriptsize$\pm$0.83} & 92.87{\scriptsize$\pm$0.34} & 93.26{\scriptsize$\pm$0.05}$^\dagger$ & \underline{93.98{\scriptsize$\pm$0.26}} & 81.76{\scriptsize$\pm$0.43}$^\dagger$ & 92.97{\scriptsize$\pm$0.09}$^\dagger$ & \textbf{94.13{\scriptsize$\pm$0.35}} & \textbf{94.13{\scriptsize$\pm$0.35}} \\
 & Social Evo. & 92.05{\scriptsize$\pm$0.46} & 95.39{\scriptsize$\pm$0.17} & 87.34{\scriptsize$\pm$0.08} & 81.60{\scriptsize$\pm$0.00} & 95.23{\scriptsize$\pm$0.07} & 90.84{\scriptsize$\pm$3.72} & 96.16{\scriptsize$\pm$0.02} & \underline{96.30{\scriptsize$\pm$0.01}} & \textbf{96.43{\scriptsize$\pm$0.02}} & 85.89{\scriptsize$\pm$0.90}$^\dagger$ & 94.59{\scriptsize$\pm$0.03}$^\dagger$ & \underline{96.30{\scriptsize$\pm$0.02}} & \underline{96.30{\scriptsize$\pm$0.02}} \\
 & UCI & 90.44{\scriptsize$\pm$0.49} & 92.03{\scriptsize$\pm$1.13} & 93.87{\scriptsize$\pm$0.08} & 77.30{\scriptsize$\pm$0.00} & 91.81{\scriptsize$\pm$0.67} & 92.31{\scriptsize$\pm$0.37} & \underline{95.57{\scriptsize$\pm$0.16}} & 94.28{\scriptsize$\pm$0.31}$^\dagger$ & \textbf{96.79{\scriptsize$\pm$0.05}} & 84.79{\scriptsize$\pm$0.51}$^\dagger$ & 92.86{\scriptsize$\pm$0.75}$^\dagger$ & 93.99{\scriptsize$\pm$0.61} & 93.99{\scriptsize$\pm$0.61} \\
 & Flights & 96.21{\scriptsize$\pm$1.42} & 98.22{\scriptsize$\pm$0.13} & 98.45{\scriptsize$\pm$0.01} & 90.23{\scriptsize$\pm$0.00} & 91.13{\scriptsize$\pm$0.01} & 98.69{\scriptsize$\pm$0.10} & 98.89{\scriptsize$\pm$0.02} & \underline{98.93{\scriptsize$\pm$0.01}} & \textbf{99.00{\scriptsize$\pm$0.02}} & 86.57{\scriptsize$\pm$1.18}$^\dagger$ & 90.99{\scriptsize$\pm$0.01}$^\dagger$ & \underline{98.93{\scriptsize$\pm$0.02}} & \underline{98.93{\scriptsize$\pm$0.02}} \\
 & Can. Parl. & 78.21{\scriptsize$\pm$0.23} & 76.99{\scriptsize$\pm$1.80} & 75.70{\scriptsize$\pm$3.27} & 64.14{\scriptsize$\pm$0.00} & 83.17{\scriptsize$\pm$0.53} & 84.04{\scriptsize$\pm$1.13} & 77.96{\scriptsize$\pm$1.46} & 62.81{\scriptsize$\pm$3.23}$^\dagger$ & \textbf{86.54{\scriptsize$\pm$0.38}}$^\dagger$ & 75.42{\scriptsize$\pm$2.00}$^\dagger$ & 80.75{\scriptsize$\pm$0.26}$^\dagger$ & \underline{84.46{\scriptsize$\pm$2.46}} & 84.44{\scriptsize$\pm$2.45} \\
 & US Legis. & 82.85{\scriptsize$\pm$1.07} & \underline{83.34{\scriptsize$\pm$0.43}} & 77.16{\scriptsize$\pm$0.39} & 62.57{\scriptsize$\pm$0.00} & 76.96{\scriptsize$\pm$0.79} & \textbf{85.36{\scriptsize$\pm$0.52}} & 82.10{\scriptsize$\pm$0.85} & 77.90{\scriptsize$\pm$0.58} & 75.33{\scriptsize$\pm$0.49}$^\dagger$ & 73.81{\scriptsize$\pm$1.09}$^\dagger$ & 73.62{\scriptsize$\pm$1.10}$^\dagger$ & 75.13{\scriptsize$\pm$0.68} & 75.13{\scriptsize$\pm$0.69} \\
 & UN Trade & 69.62{\scriptsize$\pm$0.44} & 69.10{\scriptsize$\pm$1.67} & 68.54{\scriptsize$\pm$0.18} & 66.75{\scriptsize$\pm$0.00} & 65.52{\scriptsize$\pm$0.51} & \textbf{77.61{\scriptsize$\pm$1.36}} & \underline{74.87{\scriptsize$\pm$0.53}} & 70.20{\scriptsize$\pm$1.44} & 71.05{\scriptsize$\pm$0.34}$^\dagger$ & 60.36{\scriptsize$\pm$0.94}$^\dagger$ & 62.38{\scriptsize$\pm$0.25}$^\dagger$ & 72.52{\scriptsize$\pm$0.16} & 72.52{\scriptsize$\pm$0.16} \\
 & UN Vote & 68.53{\scriptsize$\pm$0.95} & 69.71{\scriptsize$\pm$2.65} & 53.09{\scriptsize$\pm$0.22} & 62.97{\scriptsize$\pm$0.00} & 52.46{\scriptsize$\pm$0.27} & \textbf{75.32{\scriptsize$\pm$0.63}} & \underline{70.69{\scriptsize$\pm$1.02}} & 57.12{\scriptsize$\pm$0.62} & 61.63{\scriptsize$\pm$0.05}$^\dagger$ & 54.47{\scriptsize$\pm$0.85}$^\dagger$ & 51.98{\scriptsize$\pm$0.97}$^\dagger$ & 63.55{\scriptsize$\pm$0.24} & 63.55{\scriptsize$\pm$0.24} \\
 & Contact & 96.66{\scriptsize$\pm$0.89} & 97.54{\scriptsize$\pm$0.35} & 89.99{\scriptsize$\pm$0.34} & 94.34{\scriptsize$\pm$0.00} & 93.94{\scriptsize$\pm$0.02} & 97.79{\scriptsize$\pm$0.16} & \underline{98.90{\scriptsize$\pm$0.02}} & 98.53{\scriptsize$\pm$0.01} & \textbf{98.91{\scriptsize$\pm$0.01}} & 87.86{\scriptsize$\pm$0.59}$^\dagger$ & 93.84{\scriptsize$\pm$0.05}$^\dagger$ & 98.46{\scriptsize$\pm$0.02} & 98.46{\scriptsize$\pm$0.02} \\
\cmidrule{2-15}
 & Avg. Rank & 8.62 & 6.77 & 8.62 & 11.31 & 9.62 & 5.42 & 4.50 & 4.77 & \textbf{2.62} & 12.00 & 8.38 & \underline{4.12} & 4.27 \\
\midrule
\multirow{14}{*}{\rotatebox{90}{hist}} & Wikipedia & 80.77{\scriptsize$\pm$0.73} & 82.74{\scriptsize$\pm$0.32} & 67.84{\scriptsize$\pm$0.64} & 77.27{\scriptsize$\pm$0.00} & 87.68{\scriptsize$\pm$0.17} & 69.32{\scriptsize$\pm$2.78} & \underline{89.25{\scriptsize$\pm$0.49}} & 75.46{\scriptsize$\pm$0.82}$^\dagger$ & 79.89{\scriptsize$\pm$2.47} & 88.64{\scriptsize$\pm$0.34}$^\dagger$ & 87.57{\scriptsize$\pm$0.20}$^\dagger$ & 86.52{\scriptsize$\pm$0.62} & \textbf{90.21{\scriptsize$\pm$0.20}} \\
 & Reddit & 80.52{\scriptsize$\pm$0.32} & 81.11{\scriptsize$\pm$0.19} & 80.27{\scriptsize$\pm$0.30} & 78.58{\scriptsize$\pm$0.00} & 77.80{\scriptsize$\pm$0.12} & 79.36{\scriptsize$\pm$0.31} & 82.98{\scriptsize$\pm$0.63} & 80.79{\scriptsize$\pm$0.33}$^\dagger$ & 81.87{\scriptsize$\pm$0.49} & 72.79{\scriptsize$\pm$0.55}$^\dagger$ & \underline{87.66{\scriptsize$\pm$0.25}}$^\dagger$ & 81.69{\scriptsize$\pm$0.91} & \textbf{88.05{\scriptsize$\pm$0.20}} \\
 & MOOC & 82.75{\scriptsize$\pm$0.83} & 88.00{\scriptsize$\pm$1.80} & 71.57{\scriptsize$\pm$1.07} & 61.90{\scriptsize$\pm$0.00} & 76.68{\scriptsize$\pm$1.40} & 84.93{\scriptsize$\pm$2.89} & 87.44{\scriptsize$\pm$1.74} & 87.04{\scriptsize$\pm$0.35} & \underline{93.45{\scriptsize$\pm$0.67}} & 73.89{\scriptsize$\pm$1.42}$^\dagger$ & 76.34{\scriptsize$\pm$1.35}$^\dagger$ & 88.21{\scriptsize$\pm$0.99} & \textbf{98.26{\scriptsize$\pm$0.18}} \\
 & LastFM & 75.22{\scriptsize$\pm$2.36} & 77.97{\scriptsize$\pm$3.04} & 67.88{\scriptsize$\pm$0.24} & 78.09{\scriptsize$\pm$0.00} & 64.21{\scriptsize$\pm$0.73} & 75.89{\scriptsize$\pm$2.21} & 81.89{\scriptsize$\pm$0.96} & 78.83$^\dagger$ & \underline{84.64{\scriptsize$\pm$0.45}} & 69.06{\scriptsize$\pm$0.22}$^\dagger$ & 78.63{\scriptsize$\pm$2.09}$^\dagger$ & 81.38{\scriptsize$\pm$0.83} & \textbf{94.74{\scriptsize$\pm$0.13}} \\
 & Enron & 75.39{\scriptsize$\pm$2.37} & 77.09{\scriptsize$\pm$2.22} & 65.10{\scriptsize$\pm$0.34} & 79.59{\scriptsize$\pm$0.00} & 75.27{\scriptsize$\pm$1.14} & 73.22{\scriptsize$\pm$2.18} & 83.80{\scriptsize$\pm$0.68} & 77.14{\scriptsize$\pm$0.67}$^\dagger$ & 81.16{\scriptsize$\pm$1.28} & 76.15{\scriptsize$\pm$0.52}$^\dagger$ & \underline{85.15{\scriptsize$\pm$0.62}}$^\dagger$ & 80.62{\scriptsize$\pm$1.95} & \textbf{89.86{\scriptsize$\pm$0.80}} \\
 & Social Evo. & 90.06{\scriptsize$\pm$3.15} & 94.71{\scriptsize$\pm$0.53} & 87.43{\scriptsize$\pm$0.15} & 85.81{\scriptsize$\pm$0.00} & 94.39{\scriptsize$\pm$0.31} & 91.87{\scriptsize$\pm$4.52} & 96.82{\scriptsize$\pm$0.24} & 97.28{\scriptsize$\pm$0.07} & 97.22{\scriptsize$\pm$0.30} & 85.99{\scriptsize$\pm$1.26}$^\dagger$ & 93.00{\scriptsize$\pm$0.46}$^\dagger$ & \underline{97.37{\scriptsize$\pm$0.21}} & \textbf{99.24{\scriptsize$\pm$0.02}} \\
 & UCI & 78.64{\scriptsize$\pm$3.50} & 77.25{\scriptsize$\pm$2.68} & 57.86{\scriptsize$\pm$0.15} & 69.56{\scriptsize$\pm$0.00} & 77.54{\scriptsize$\pm$2.02} & 71.52{\scriptsize$\pm$1.61} & \underline{92.05{\scriptsize$\pm$0.36}} & 76.32{\scriptsize$\pm$0.45}$^\dagger$ & 80.42{\scriptsize$\pm$0.64} & 77.86{\scriptsize$\pm$0.98}$^\dagger$ & 81.18{\scriptsize$\pm$1.15}$^\dagger$ & 86.72{\scriptsize$\pm$1.32} & \textbf{94.96{\scriptsize$\pm$0.25}} \\
 & Flights & 68.97{\scriptsize$\pm$1.87} & 68.39{\scriptsize$\pm$0.95} & 66.11{\scriptsize$\pm$0.71} & \underline{74.64{\scriptsize$\pm$0.00}} & 70.37{\scriptsize$\pm$0.23} & 67.11{\scriptsize$\pm$2.16} & 69.52{\scriptsize$\pm$0.90} & 68.09{\scriptsize$\pm$0.43} & 71.82{\scriptsize$\pm$0.82} & 64.37{\scriptsize$\pm$2.87}$^\dagger$ & 70.68{\scriptsize$\pm$0.14}$^\dagger$ & 71.79{\scriptsize$\pm$0.77} & \textbf{79.85{\scriptsize$\pm$0.20}} \\
 & Can. Parl. & 62.44{\scriptsize$\pm$1.11} & 73.23{\scriptsize$\pm$3.08} & 72.06{\scriptsize$\pm$3.94} & 63.04{\scriptsize$\pm$0.00} & 79.03{\scriptsize$\pm$1.01} & \underline{81.47{\scriptsize$\pm$2.58}} & 73.89{\scriptsize$\pm$4.50} & 63.27{\scriptsize$\pm$1.69}$^\dagger$ & 73.21{\scriptsize$\pm$2.22}$^\dagger$ & 74.29{\scriptsize$\pm$5.21}$^\dagger$ & 72.81{\scriptsize$\pm$2.04}$^\dagger$ & \textbf{84.16{\scriptsize$\pm$4.02}} & \textbf{84.16{\scriptsize$\pm$4.02}} \\
 & US Legis. & 67.47{\scriptsize$\pm$6.40} & 83.53{\scriptsize$\pm$4.53} & 78.62{\scriptsize$\pm$7.46} & 67.41{\scriptsize$\pm$0.00} & 85.17{\scriptsize$\pm$0.70} & \textbf{94.63{\scriptsize$\pm$1.16}} & 83.45{\scriptsize$\pm$2.85} & \underline{90.77{\scriptsize$\pm$1.96}} & 79.78{\scriptsize$\pm$1.11}$^\dagger$ & 76.65{\scriptsize$\pm$1.74}$^\dagger$ & 67.38{\scriptsize$\pm$5.83}$^\dagger$ & 78.84{\scriptsize$\pm$6.61} & 81.26{\scriptsize$\pm$3.81} \\
 & UN Trade & 68.92{\scriptsize$\pm$1.40} & 63.93{\scriptsize$\pm$5.41} & 63.09{\scriptsize$\pm$0.74} & \underline{86.61{\scriptsize$\pm$0.00}} & 63.20{\scriptsize$\pm$1.54} & 80.68{\scriptsize$\pm$0.98} & 76.95{\scriptsize$\pm$1.92} & 73.86{\scriptsize$\pm$1.13} & 72.28{\scriptsize$\pm$0.39}$^\dagger$ & 59.54{\scriptsize$\pm$0.93}$^\dagger$ & 61.38{\scriptsize$\pm$2.11}$^\dagger$ & 78.85{\scriptsize$\pm$3.82} & \textbf{91.46{\scriptsize$\pm$0.68}} \\
 & UN Vote & 76.84{\scriptsize$\pm$1.01} & 73.40{\scriptsize$\pm$5.20} & 51.27{\scriptsize$\pm$0.33} & \textbf{89.62{\scriptsize$\pm$0.00}} & 52.61{\scriptsize$\pm$1.44} & 76.47{\scriptsize$\pm$2.15} & 76.61{\scriptsize$\pm$3.05} & 64.27{\scriptsize$\pm$1.78} & 72.58{\scriptsize$\pm$1.12}$^\dagger$ & 61.08{\scriptsize$\pm$3.33}$^\dagger$ & 53.86{\scriptsize$\pm$1.50}$^\dagger$ & 76.74{\scriptsize$\pm$1.95} & \underline{81.00{\scriptsize$\pm$1.12}} \\
 & Contact & 96.35{\scriptsize$\pm$0.92} & 93.76{\scriptsize$\pm$1.29} & 83.06{\scriptsize$\pm$0.32} & 92.17{\scriptsize$\pm$0.00} & 93.14{\scriptsize$\pm$0.34} & 96.79{\scriptsize$\pm$0.50} & 97.34{\scriptsize$\pm$0.16} & 97.17{\scriptsize$\pm$0.05} & \underline{97.73{\scriptsize$\pm$0.11}} & 82.31{\scriptsize$\pm$1.19}$^\dagger$ & 92.92{\scriptsize$\pm$0.35}$^\dagger$ & 95.99{\scriptsize$\pm$0.37} & \textbf{98.63{\scriptsize$\pm$0.04}} \\
\cmidrule{2-15}
 & Avg. Rank & 8.38 & 7.08 & 11.54 & 8.62 & 8.31 & 7.62 & \underline{4.15} & 7.00 & 5.00 & 10.00 & 7.46 & 4.35 & \textbf{1.50} \\
\midrule
\multirow{14}{*}{\rotatebox{90}{ind}} & Wikipedia & 70.96{\scriptsize$\pm$0.78} & 80.97{\scriptsize$\pm$0.31} & 70.95{\scriptsize$\pm$0.95} & 81.73{\scriptsize$\pm$0.00} & \underline{84.28{\scriptsize$\pm$0.30}} & 52.11{\scriptsize$\pm$5.83} & 77.03{\scriptsize$\pm$1.93} & 69.86{\scriptsize$\pm$0.36}$^\dagger$ & 75.36{\scriptsize$\pm$3.41} & 83.53{\scriptsize$\pm$0.60}$^\dagger$ & 83.54{\scriptsize$\pm$0.40}$^\dagger$ & 81.22{\scriptsize$\pm$1.04} & \textbf{85.97{\scriptsize$\pm$0.19}} \\
 & Reddit & 83.51{\scriptsize$\pm$0.15} & 84.56{\scriptsize$\pm$0.24} & \underline{88.04{\scriptsize$\pm$0.29}} & 85.93{\scriptsize$\pm$0.00} & 82.21{\scriptsize$\pm$0.13} & 76.01{\scriptsize$\pm$1.04} & 81.02{\scriptsize$\pm$0.68} & 86.71{\scriptsize$\pm$0.62}$^\dagger$ & 81.64{\scriptsize$\pm$0.42} & 76.80{\scriptsize$\pm$0.50}$^\dagger$ & \textbf{91.24{\scriptsize$\pm$0.11}}$^\dagger$ & 85.00{\scriptsize$\pm$0.42} & 87.05{\scriptsize$\pm$0.22} \\
 & MOOC & 66.63{\scriptsize$\pm$2.30} & 77.44{\scriptsize$\pm$2.86} & 70.32{\scriptsize$\pm$1.43} & 48.18{\scriptsize$\pm$0.00} & 72.45{\scriptsize$\pm$0.72} & 75.56{\scriptsize$\pm$2.39} & 77.00{\scriptsize$\pm$2.24} & 80.76{\scriptsize$\pm$0.76} & \underline{89.07{\scriptsize$\pm$0.63}} & 62.07{\scriptsize$\pm$0.24}$^\dagger$ & 72.03{\scriptsize$\pm$0.63}$^\dagger$ & 80.14{\scriptsize$\pm$1.90} & \textbf{94.67{\scriptsize$\pm$0.35}} \\
 & LastFM & 61.32{\scriptsize$\pm$3.49} & 65.46{\scriptsize$\pm$4.27} & 67.92{\scriptsize$\pm$0.44} & \underline{77.37{\scriptsize$\pm$0.00}} & 60.22{\scriptsize$\pm$0.32} & 58.54{\scriptsize$\pm$2.37} & 69.51{\scriptsize$\pm$0.61} & 68.79$^\dagger$ & 72.48{\scriptsize$\pm$0.90} & 65.11{\scriptsize$\pm$0.36}$^\dagger$ & 76.10{\scriptsize$\pm$2.01}$^\dagger$ & 69.18{\scriptsize$\pm$0.76} & \textbf{88.13{\scriptsize$\pm$0.14}} \\
 & Enron & 70.92{\scriptsize$\pm$1.05} & 71.34{\scriptsize$\pm$2.46} & 75.17{\scriptsize$\pm$0.50} & 75.00{\scriptsize$\pm$0.00} & 71.53{\scriptsize$\pm$0.85} & 69.62{\scriptsize$\pm$1.12} & 78.91{\scriptsize$\pm$0.96} & 74.10{\scriptsize$\pm$0.46}$^\dagger$ & 75.44{\scriptsize$\pm$1.38} & 70.93{\scriptsize$\pm$0.73}$^\dagger$ & \underline{83.35{\scriptsize$\pm$0.48}}$^\dagger$ & 81.85{\scriptsize$\pm$0.93} & \textbf{87.56{\scriptsize$\pm$0.65}} \\
 & Social Evo. & 90.01{\scriptsize$\pm$3.19} & 95.24{\scriptsize$\pm$0.56} & 89.93{\scriptsize$\pm$0.15} & 87.88{\scriptsize$\pm$0.00} & 94.22{\scriptsize$\pm$0.32} & 92.53{\scriptsize$\pm$4.20} & 97.10{\scriptsize$\pm$0.22} & 97.51{\scriptsize$\pm$0.06} & 97.59{\scriptsize$\pm$0.25} & 86.13{\scriptsize$\pm$1.03}$^\dagger$ & 92.85{\scriptsize$\pm$0.44}$^\dagger$ & \underline{97.76{\scriptsize$\pm$0.18}} & \textbf{99.36{\scriptsize$\pm$0.02}} \\
 & UCI & 64.14{\scriptsize$\pm$1.26} & 64.11{\scriptsize$\pm$1.04} & 58.06{\scriptsize$\pm$0.26} & 58.03{\scriptsize$\pm$0.00} & 74.59{\scriptsize$\pm$0.74} & 49.73{\scriptsize$\pm$1.20} & \underline{79.42{\scriptsize$\pm$0.31}} & 65.45{\scriptsize$\pm$0.25}$^\dagger$ & 70.85{\scriptsize$\pm$0.96} & 74.17{\scriptsize$\pm$0.55}$^\dagger$ & 76.20{\scriptsize$\pm$0.35}$^\dagger$ & 76.06{\scriptsize$\pm$0.45} & \textbf{84.75{\scriptsize$\pm$0.15}} \\
 & Flights & 69.99{\scriptsize$\pm$3.10} & 71.63{\scriptsize$\pm$1.72} & 69.70{\scriptsize$\pm$0.75} & \textbf{81.10{\scriptsize$\pm$0.00}} & 72.21{\scriptsize$\pm$0.21} & 59.73{\scriptsize$\pm$1.61} & 59.61{\scriptsize$\pm$1.65} & 69.53{\scriptsize$\pm$1.17} & 64.21{\scriptsize$\pm$1.30} & 67.74{\scriptsize$\pm$3.92}$^\dagger$ & 72.48{\scriptsize$\pm$0.16}$^\dagger$ & 69.96{\scriptsize$\pm$1.37} & \underline{81.00{\scriptsize$\pm$0.36}} \\
 & Can. Parl. & 52.88{\scriptsize$\pm$0.80} & 69.57{\scriptsize$\pm$2.81} & 72.93{\scriptsize$\pm$1.78} & 61.41{\scriptsize$\pm$0.00} & 70.52{\scriptsize$\pm$0.94} & \textbf{80.03{\scriptsize$\pm$1.75}} & 65.30{\scriptsize$\pm$6.98} & 61.05{\scriptsize$\pm$1.65}$^\dagger$ & 72.34{\scriptsize$\pm$2.18}$^\dagger$ & 62.45{\scriptsize$\pm$7.55}$^\dagger$ & 64.21{\scriptsize$\pm$2.08}$^\dagger$ & 77.14{\scriptsize$\pm$3.31} & \underline{77.37{\scriptsize$\pm$3.51}} \\
 & US Legis. & 59.05{\scriptsize$\pm$5.52} & 78.12{\scriptsize$\pm$4.46} & 76.45{\scriptsize$\pm$7.02} & 68.66{\scriptsize$\pm$0.00} & 84.22{\scriptsize$\pm$0.91} & \textbf{93.04{\scriptsize$\pm$1.35}} & 79.09{\scriptsize$\pm$3.45} & \underline{87.96{\scriptsize$\pm$1.80}} & 79.58{\scriptsize$\pm$1.50}$^\dagger$ & 73.13{\scriptsize$\pm$1.85}$^\dagger$ & 65.75{\scriptsize$\pm$8.06}$^\dagger$ & 79.79{\scriptsize$\pm$7.64} & 83.74{\scriptsize$\pm$3.42} \\
 & UN Trade & 66.82{\scriptsize$\pm$1.27} & 66.37{\scriptsize$\pm$5.39} & 71.73{\scriptsize$\pm$0.74} & 74.20{\scriptsize$\pm$0.00} & 66.53{\scriptsize$\pm$1.22} & \underline{80.77{\scriptsize$\pm$1.31}} & 75.76{\scriptsize$\pm$1.54} & 62.56{\scriptsize$\pm$1.51} & 66.57{\scriptsize$\pm$0.62}$^\dagger$ & 52.84{\scriptsize$\pm$0.92}$^\dagger$ & 62.52{\scriptsize$\pm$1.80}$^\dagger$ & 74.04{\scriptsize$\pm$4.28} & \textbf{89.43{\scriptsize$\pm$0.49}} \\
 & UN Vote & 73.73{\scriptsize$\pm$1.61} & 72.69{\scriptsize$\pm$3.72} & 52.75{\scriptsize$\pm$0.90} & 72.85{\scriptsize$\pm$0.00} & 51.89{\scriptsize$\pm$0.74} & \textbf{77.36{\scriptsize$\pm$1.52}} & 74.11{\scriptsize$\pm$2.51} & 53.37{\scriptsize$\pm$1.26} & 61.96{\scriptsize$\pm$0.90}$^\dagger$ & 55.83{\scriptsize$\pm$4.27}$^\dagger$ & 50.33{\scriptsize$\pm$2.46}$^\dagger$ & 69.34{\scriptsize$\pm$2.40} & \underline{74.92{\scriptsize$\pm$1.54}} \\
 & Contact & 94.47{\scriptsize$\pm$1.08} & 91.64{\scriptsize$\pm$1.72} & 87.68{\scriptsize$\pm$0.24} & 85.87{\scriptsize$\pm$0.00} & 90.96{\scriptsize$\pm$0.27} & 92.51{\scriptsize$\pm$1.16} & 94.71{\scriptsize$\pm$0.33} & 95.01{\scriptsize$\pm$0.15} & \underline{95.93{\scriptsize$\pm$0.23}} & 78.48{\scriptsize$\pm$1.54}$^\dagger$ & 90.50{\scriptsize$\pm$0.37}$^\dagger$ & 95.83{\scriptsize$\pm$0.26} & \textbf{97.54{\scriptsize$\pm$0.06}} \\
\cmidrule{2-15}
 & Avg. Rank & 9.23 & 7.54 & 8.31 & 7.69 & 7.31 & 8.08 & 6.15 & 7.23 & 6.15 & 10.23 & 6.77 & \underline{4.69} & \textbf{1.62} \\
\midrule
\multicolumn{2}{l}{Overall Avg. Rank} & 8.74 & 7.13 & 9.49 & 9.21 & 8.41 & 7.04 & 4.94 & 6.33 & 4.59 & 10.74 & 7.54 & \underline{4.38} & \textbf{2.46} \\
\bottomrule
\end{tabular}}
\end{table}

This batch-level definition is important for interpreting source-activity correction. Under random negative sampling, $(u,v,t)$ and its matched negative $(u,v^-,t)$ receive the same additive term $\beta\log\lambda_u(t)$, leaving their within-pair ordering unchanged. However, AP and AUC jointly compare examples associated with different sources and query times within the same batch. The activity term can therefore alter the batch-level metrics through cross-query reordering. Our results across different values of $\beta$ in Figure~\ref{fig:beta_appendix} illustrate this distinction between same-source candidate ranking and mixed-source benchmark evaluation.

\section{Implementation Details and Reproducibility}
\label{app:implementation}

This appendix describes the training configurations of our model, the sources of the baseline results, corrections to information leakage and evaluation inconsistencies in released baseline implementations, and the procedure for transferring SNAM to independently trained backbones. We also provide the accompanying code and verification procedures needed to reproduce the reported results.

\subsection{Training Configuration}
\label{app:our_training}

\paragraph{Architecture and temporal partitioning.}

We use the same architecture across all thirteen datasets. The node embedding dimension is $d=128$, the rank of the bilinear matching features is $r=16$, and the fixed random projections have dimension $D=256$, with initialization seed $7$. The candidate-conditioned recent-interaction encoder processes up to $M=20$ historical events and uses a $64$-dimensional hidden representation. These components are detailed in Appendix~\ref{app:random_projection_features} and Appendix~\ref{app:recent_interaction_encoding}. The number of exponential timescales, $K$, is determined from training timestamps using the construction in Appendix~\ref{app:timescale_configuration}; its value for each dataset is reported in Table~\ref{tab:datasets}. Historical edge-feature channels are enabled when the dataset provides valid event features.

The training stream is partitioned chronologically into blocks. On eight datasets with fine-grained timestamps, blocks contain approximately equal numbers of events, with the number of blocks specified in Table~\ref{tab:training_settings}. For Flights, Canadian Parliament, US Legislation, UN Trade, and UN Vote, events are grouped by discrete time step so that a block boundary does not separate events with the same timestamp. When a split contains more than $96$ time steps, consecutive steps are merged into at most $96$ blocks. Validation and test streams use the corresponding partitioning procedure.

\begin{table}[t]
\caption{Per-stream training settings of our model. Blocks: the training period is processed in $C$ blocks of equal event counts (equal) or in one block per time step (step). Table: the node embedding table $E$ is trained (yes) or holds fixed random codes (no); weight decay applies to the table only and is therefore listed only where the table is trained. Negatives: training negatives are drawn from all destinations of the training period (all) or from the destinations observed before the current block (seen). Partner updates: within-block events also update the partner-sum state (Appendix~\ref{app:query_complexity}); they always update the counts. All other settings are shared by every stream: $d=128$, $r=16$, $256$ random projection dimensions, $M=20$, learning rate $10^{-3}$, at most $500$ epochs, five seeds; $K$ is given in Table~\ref{tab:datasets}.}
\label{tab:training_settings}
\centering
\footnotesize
\setlength{\tabcolsep}{5pt}
\begin{tabular}{lcccccc}
\toprule
Dataset & Blocks $C$ & Table & Dropout & Weight decay & Negatives & Partner updates \\
\midrule
Wikipedia   & 96 equal  & no  & 0   & --   & seen & no  \\
Reddit      & 48 equal  & no  & 0   & --   & seen & no  \\
MOOC        & 192 equal & no  & 0   & --   & seen & no  \\
LastFM      & 192 equal & no  & 0   & --   & seen & no  \\
Enron       & 192 equal & no  & 0   & --   & seen & yes \\
Social Evo. & 96 equal  & no  & 0   & --   & seen & no  \\
UCI         & 192 equal & yes & 0.2 & 0    & all  & no  \\
Flights     & step      & yes & 0   & 0.01 & all  & no  \\
Can. Parl.  & step      & no  & 0   & --   & seen & no  \\
US Legis.   & step      & yes & 0.2 & 0    & all  & no  \\
UN Trade    & step      & no  & 0   & --   & seen & no  \\
UN Vote     & step      & no  & 0   & --   & seen & no  \\
Contact     & 192 equal & no  & 0   & --   & seen & yes \\
\bottomrule
\end{tabular}
\end{table}

\paragraph{Optimization.}

We use AdamW \citep{loshchilov2019decoupled} and perform one optimizer update per training block. The relation branch has a base learning rate of $10^{-3}$; the zero-initialized output projections of the scalar pairwise features and recent-interaction encoder use ten times this rate. SNAM is optimized independently with learning rate $0.05$ and no weight decay.

The node embedding table is either trainable or initialized with fixed random codes, as specified in Table~\ref{tab:training_settings}. Weight decay is applied only to the node embedding table when it is trainable. Where enabled, dropout is applied to the concatenated node representations and pairwise historical features at the input of the relation scoring network.

Each positive training interaction is paired with one random destination negative, resampled every epoch. Depending on the dataset, negative destinations are sampled either from all training destinations or from destinations observed before the current block. The dataset-specific choices, block counts, dropout rates, and within-block partner-state updates are given in Table~\ref{tab:training_settings}. In particular, the MOOC configuration uses $192$ equal-event blocks, fixed random node codes, and zero dropout.

For Enron and Contact, within-block interactions additionally update the first-order partner-aggregation state. The remaining datasets use the default block-start node-vector states. Activity counts, historical edge-feature aggregates, and random-projection states incorporate strictly earlier within-block interactions in all datasets.

\paragraph{Model selection and evaluation.}

Each configuration is trained for at most $500$ epochs. After every epoch, we evaluate the validation split under random negative sampling. A checkpoint is retained when both validation AP and AUC are at least as high as their previously recorded values. Early stopping occurs after $100$ epochs without a checkpoint update, or $50$ epochs for LastFM and Flights. Main-model experiments use five seeds, $0$--$4$.

For the activity-corrected score in \eqref{eq:app-a10}, we select
\[
\beta\in\{0,0.5,1,1.5,2\}
\]
on the validation split independently for each prediction setting and negative-sampling protocol. We maximize validation AP and choose the smaller coefficient in the event of a tie. The selected value is then applied to the corresponding test split. Results labeled \emph{Ours} use the base relation score alone, with $\beta=0$.

\subsection{Baseline Results and Retraining}
\label{app:baseline_sources}

Table~\ref{tab:baseline_sources} identifies the source of every baseline result in the main comparison tables and distinguishes published results from our corrected retraining runs.

\begin{table}[t]
\caption{Baseline numbers that we produced ourselves. The models listed here were retrained with their released code, configuration and early-stopping rule after the corrections of Appendix~\ref{app:baseline_corrections}; the retrained cells are marked $\dagger$ in Tables~\ref{tab:trans_ap}, \ref{tab:trans_auc}, \ref{tab:ind_ap} and \ref{tab:ind_auc}, and the number of seeds is given in parentheses. All other cells, and all cells of JODIE, TGN, CAWN, EdgeBank, GraphMixer, NAT and PINT, are the values reported by \citet{lu2024tpnet} under the same protocol.}
\label{tab:baseline_sources}
\centering
\footnotesize
\setlength{\tabcolsep}{5pt}
\begin{tabular}{@{}l>{\raggedright\arraybackslash}p{6.0cm}>{\raggedright\arraybackslash}p{5.3cm}@{}}
\toprule
Model & Retrained on (seeds) & Correction \\
\midrule
DyGFormer & Can.\ Parl.\ (5); Wikipedia, Reddit, UCI, Enron (3); LastFM (1) & fixed-length padding \\
\addlinespace[2pt]
TPNet & Can.\ Parl., US Legis., UN Trade, UN Vote (5) & strictly earlier commit of events \\
\addlinespace[2pt]
GRN & all thirteen streams (5) & six corrections, Appendix~\ref{app:baseline_corrections} \\
\addlinespace[2pt]
DSRD & all thirteen streams (5) & batch row mapping; loader alignment \\
\bottomrule
\end{tabular}
\end{table}

\paragraph{Published and retrained results.}

The published values for JODIE, TGN, CAWN, EdgeBank, GraphMixer, NAT, PINT, and the unaffected DyGFormer and TPNet configurations are taken from \citet{lu2024tpnet}, which reports these methods under the same benchmark and evaluation protocols. Where our inspection of released implementations identified information leakage or protocol inconsistencies affecting the results, we corrected the implementation, retrained the corresponding model, and marked our results with $\dagger$.

GRN \citep{chang2026grn} and DSRD \citep{chang2026dsrd} are retrained on all thirteen datasets with five seeds using their released architectures and configurations, together with the corrections described in Appendix~\ref{app:baseline_corrections}.

For DyGFormer \citep{yu2023dyglib}, we retrain Canadian Parliament with five seeds after correcting its sequence-padding behavior. We also train DyGFormer on Wikipedia, Reddit, UCI, Enron, and LastFM to obtain the trained backbone checkpoints required by our transfer experiments, because the corresponding pretrained checkpoints were not available in the released materials. The first four datasets use three seeds, and LastFM one, since a DyGFormer run on LastFM is by far the most expensive of these trainings; its LastFM cells are therefore indicative rather than averaged results, with the seed spread seen on the other four streams (below 0.2 AP under random negatives, about two points under historical and inductive negatives) as a guide to their uncertainty. Single-run cells are identified by the seed counts in Table~\ref{tab:baseline_sources}.
 These runs produce the independently trained backbones to which SNAM is subsequently applied; SNAM transfer itself requires no backbone retraining. The main comparison tables report our retrained results on these six datasets and published values elsewhere.

For TPNet \citep{lu2024tpnet}, we first evaluate the corrected implementation with seed $0$ on all thirteen datasets. The tolerance for a dataset is $\max(1.0,2\sigma)$ percentage points on each of its six AP measurements---two prediction settings and three negative-sampling protocols---where $\sigma$ is the published standard deviation. Where the seed-0 run falls outside this tolerance we train five seeds (five seeds also exist on Flights, Contact, Enron, MOOC and UCI, and three on the transfer streams), and a dataset keeps its published values if the mean over its runs lies within the tolerance on all six cells. Canadian Parliament, US Legislation, UN Trade and UN Vote fail and are reported with their retrained means; the remaining nine datasets retain the published values.

All corrected retraining runs use the respective models' released architectures, dataset configurations, and early-stopping rules, without introducing a new baseline hyperparameter search. The corrected implementations and the runs producing the reported results are included in our accompanying code.

\subsection{Corrections to Released Baseline Implementations}
\label{app:baseline_corrections}

Our evaluation uses the strictly preceding history
\[
\mathcal H_{t^-}=\{e:t_e<t\},
\]
as defined in Appendix~\ref{app:data_splits_visibility}. We identified several ways in which released baseline implementations could expose information from the query's own timestamp or label, or allow one query's computation to depend on other examples in the same batch. We corrected these information paths while retaining the original model architectures, objectives, and training configurations. Corrections affecting training are applied in both training and evaluation, followed by retraining from initialization. All modifications are marked in the accompanying source code to facilitate independent inspection.

\paragraph{TPNet: delayed commitment of simultaneous events.}

TPNet maintains temporal random-projection states and updates them after scoring each batch. In the released implementation, an entire scored batch is immediately incorporated into the states used by the next batch. When two consecutive batches contain events with the same timestamp, a query in the second batch can therefore access simultaneous events from the first batch.

We replace immediate state updates with delayed commitment. Scored events are placed in a pending buffer. Before scoring a new batch whose first timestamp is $t$, only buffered events satisfying $t_e<t$ are committed through TPNet's original update operator. Events with $t_e\geq t$ remain pending until the query time advances beyond them.

This preserves TPNet's state representation and update equations while preventing information from simultaneous events from crossing batch boundaries. Remaining buffered events are committed when the training epoch or evaluation stream ends.

\paragraph{DyGFormer: eliminating label-correlated padding.}

DyGFormer encodes recent interaction sequences using temporal patches and pooled sequence representations. In the released implementation, each model call pads its sequences to the maximum length appearing in that call. Positive and negative examples are evaluated in separate calls, whose historical sequence-length distributions may differ.

Because padded positions participate in the subsequent unmasked pooling operation, the padding length can influence the resulting representations. Separate positive and negative calls therefore introduce a label-correlated feature unrelated to the individual query's history.

We set the padding length of every call to the maximum sequence length $L$ specified by the released configuration. The historical sequence truncation, padding values, network architecture, and pooling operation remain unchanged. Every query is consequently encoded using the same input length, independently of the other examples in its model call.

\paragraph{GRN: consistent supervision and causal state updates.}

We corrected six issues in the released GRN training and evaluation pipeline, organized into three parts.

First, we align predictions with their supervision. The released training loop used the sampled negative destination in the positive branch, rather than the true interaction destination. We restore the true pair $(u,v)$ as the positive example and use $(u,v^-)$ only as the negative example. Features of the interaction being predicted are also removed from the scoring inputs; observed event features enter the model only when their events are committed to the historical state.

Second, we enforce causal history and consistent state access. The released within-chunk mask was based on record order and could expose earlier rows with the same timestamp. Moreover, updates performed by the positive branch could influence the negative branch, while scoring a negative pair could overwrite the sampled destination's last-interaction time. We replace the row-order mask with a strict temporal condition, score positive and negative candidates against the same committed history, and prohibit negative examples from modifying historical states. Newly observed positive events become available only to queries with strictly later timestamps.

Third, we correct state restoration and dataset loading. The original state backups retained references to tensors subsequently modified in place, so restoring them did not reliably recover the intended historical state. We create independent state copies and reconstruct the training-end history by replaying training interactions under the selected checkpoint before validation or test evaluation. We also retain all appropriate transductive validation and test events, including those involving nodes held out for inductive evaluation, and provide historical and inductive negative-sampling evaluation using the DyGLib protocols.

These changes leave GRN's retention architecture and loss formulation intact while ensuring that its predictions are based on the intended historical information.

\paragraph{DSRD: separating the local subgraphs of different queries.}

DSRD constructs a joint subgraph from the center nodes and sampled temporal neighbors of a batch. The released construction concatenated center-node IDs and neighbor occurrences, then mapped global node IDs to subgraph rows. When a center node also appeared as another query's neighbor, repeated IDs could be mapped to the same row, causing historical information from different queries to be combined unintentionally.

Because positive and negative examples were scored in separate calls, the resulting collisions could also depend on the composition and label of the batch.

We assign separate local subgraph rows to every center node and every sampled neighbor occurrence, and connect each historical edge to the representation belonging to its own query. The resulting subgraph of a candidate depends on that candidate's sampled history rather than on the other candidates in the batch.

We additionally align the unseen-node sampling order with DyGLib and retain the complete transductive validation and test event sets.

\paragraph{Memory-based models: causal message commitment.}

The released memory-update implementations used by JODIE \citep{kumar2019predicting}, DyRep \citep{trivedi2019dyrep}, and TGN \citep{rossi2020temporal} may aggregate pending messages from the query's own timestamp when consecutive batches share that timestamp. We apply the same strict temporal commitment principle used for TPNet: messages are incorporated into readable memory only when their timestamps precede the current batch's first query time.

The main comparison tables retain the attributed published results for models that were not retrained in this study. For SNAM transfer, we use TGN trained with the corrected memory implementation. The corresponding corrected source is included with the other baseline modifications.

\subsection{Transferring SNAM to Other Models}
\label{app:transfer_details}

We evaluate SNAM as a reusable activity component by applying it to TPNet, TGN, DyGFormer, and DSRD on Wikipedia, Reddit, UCI, Enron, and LastFM. The results are reported in Table~\ref{tab:transfer}.

We first obtain independently trained backbone checkpoints using each model's released architecture, dataset configuration, and early-stopping procedure, with the relevant corrections from Appendix~\ref{app:baseline_corrections}. These training runs provide the backbone checkpoints needed for evaluation where pretrained checkpoints were unavailable. After training, every backbone is frozen: adding SNAM does not require another backbone training run or any change to the backbone's parameters.

For each dataset, we reuse the SNAM parameters
\[
\vartheta=(\theta_\mu,\theta_1,\ldots,\theta_K)
\]
from the seed-$0$ checkpoint of our own model on that dataset. An independent historical-counting procedure computes the source activity of every evaluated positive and negative pair, using the same time units, exponential timescales, and strictly preceding history as our model. In particular, a historical or inductive negative pair receives the activity rate of its own source node.

For each backbone, we record the link predictor's output before the sigmoid, together with the source node and query time of every evaluated example. The transferred score is
\[
\boxed{
s_\beta(u,v,t)
=
f_{\mathrm{base}}(u,v,t)
+
\beta\log\lambda_u(t).
}
\]

Here, $f_{\mathrm{base}}$ is the frozen backbone's original logit. Neither the pretrained SNAM parameters nor the backbone parameters are updated during transfer.

We select
\[
\beta\in\{0,0.5,1,1.5,2,3,4\}
\]
by maximizing validation AP separately for each dataset, backbone, prediction setting, and negative-sampling protocol. The selected coefficient is applied unchanged to the corresponding test split. AP and AUC are computed using the batch-level evaluation procedure of Appendix~\ref{app:negative_sampling_protocols}.

\paragraph{Historical state at the validation boundary.}

For backbones that maintain persistent historical states, selecting $\beta$ requires validation predictions generated from the state at the end of training, before any validation events have been consumed.

DyGFormer has no persistent cross-batch memory. For TGN, we restore the training-end memory saved during training. For TPNet, we replay training events through its parameter-free historical update to reconstruct the training-end state. For DSRD, we restore the corresponding saved node state. Each backbone then follows its original streaming evaluation procedure.

As a consistency check, setting $\beta=0$ in the transfer pipeline reproduces the backbone-only evaluation results. This separates the effect of the additional source-activity score from changes in the backbone or its evaluation procedure.

\subsection{Code Release and Reproducibility Checks}
\label{app:reproducibility}

Our accompanying code includes the complete model, all corrected baseline implementations, their annotated modifications, experiment configurations and seeds, and the result files used to generate the paper's tables and figures. We provide scripts for reproducing individual runs and regenerating the reported results from the saved experimental records.

The dataset files are obtained from the public benchmark \citep{poursafaei2022towards}. We provide the data preparation procedure and a script that recomputes Table~\ref{tab:datasets}. Data splits, evaluation negative samplers, and metric calculations follow the corresponding DyGLib implementation \citep{yu2023dyglib}. Table~\ref{tab:baseline_sources} identifies which baseline results are reproduced by our corrected implementations and which are attributed to published experiments.

We verified the reproducibility of the experimental pipeline in three ways. First, reevaluating a saved checkpoint on another RTX 5090 GPU changes the selected test AP by at most $3\times10^{-5}$. Second, independently retraining the seed-$0$ UCI configuration on another machine reproduces its six test AP measurements---two prediction settings under three negative-sampling protocols---within $0.35$ percentage points, with identical selected $\beta$ values. Third, the SNAM transfer pipeline reproduces the independently evaluated backbone results when $\beta=0$.

Together with the annotated corrections and complete run configurations, these checks make the reported retraining and transfer results independently reproducible and allow the corrected implementations to be audited against the released originals.

\section{Additional Experimental Results and Analysis}

This appendix supplements the main experiments with complete results, cross-model transfer, module ablations, and computational cost analyses. Experimental settings and implementation details are provided in Appendices~\ref{app:datasets_protocols} and~\ref{app:implementation}, respectively. Below, the ``inductive setting'' refers to link prediction involving unseen nodes, whereas ``inductive negative sampling'' refers to the negative-sample construction protocol; we discuss these two concepts separately. All performance gains are reported in percentage points.

\subsection{Complete Main Results}
\label{app:complete_results}
Tables~\ref{tab:ind_ap}, \ref{tab:trans_auc}, and~\ref{tab:ind_auc} supplement the results in the main-text Table~\ref{tab:trans_ap}. \textbf{The advantage under hard negative sampling extends to both unseen-node prediction and different evaluation metrics.} In the inductive setting, Ours using only the relation score attains an average rank of 4.81 under both historical and inductive negative sampling, outperforming all comparison baselines. Adding SNAM improves both average ranks to 2.35 and achieves the highest AP on 9 of the 13 datasets under each protocol. For example, under historical negative sampling, its AP on MOOC reaches 95.54\%, exceeding the strongest baseline, TPNet, by 13.69 points. TPNet remains stronger under random negative sampling; our competitiveness also includes the efficiency gains from shared historical computation, as discussed in Section~\ref{app:efficiency_results}.

The AUC results exhibit the same overall trend. In the transductive setting, the full model attains average ranks of 1.50 and 1.62 under historical and inductive negative sampling, respectively. Under the historical protocol, it achieves the highest AUC on 11 of the 13 datasets; on LastFM, it reaches 94.74\%, exceeding the strongest baseline by 10.10 points (Table~\ref{tab:trans_auc}). In the inductive setting, its average rank is 2.81 under both hard negative-sampling protocols (Table~\ref{tab:ind_auc}). Across all three protocols, the full model achieves overall average ranks of 2.44, 3.18, 2.46, and 3.33 for transductive AP, inductive AP, transductive AUC, and inductive AUC, respectively, ranking first in every table.

\begin{table}[t]
\caption{Inductive link prediction, AUC (\%), mean{\scriptsize$\pm$std} over seeds (five; DyGFormer$^\dagger$ three, LastFM one) under random (rnd), historical (hist) and inductive (ind) negative sampling. \textbf{Bold}: best; \underline{underline}: second best. $\dagger$: retrained by us with the corrected released code (released configuration).}
\label{tab:ind_auc}
\centering
\resizebox{\textwidth}{!}{%
\renewcommand{\arraystretch}{1.0}\setlength{\tabcolsep}{4pt}%
\begin{tabular}{llccccccccccccc}
\toprule
NSS & Dataset & JODIE & TGN & CAWN & TGAT & GraphMixer & NAT & PINT & DyGFormer & TPNet & GRN$^\dagger$ & DSRD$^\dagger$ & Ours & Ours+$\lambda$ \\
\midrule
\multirow{14}{*}{\rotatebox{90}{rnd}} & Wikipedia & 94.33{\scriptsize$\pm$0.27} & 97.72{\scriptsize$\pm$0.03} & 98.03{\scriptsize$\pm$0.04} & 95.90{\scriptsize$\pm$0.09} & 96.30{\scriptsize$\pm$0.04} & 95.82{\scriptsize$\pm$0.18} & 97.76{\scriptsize$\pm$0.05} & \underline{98.52{\scriptsize$\pm$0.01}}$^\dagger$ & \textbf{98.90{\scriptsize$\pm$0.01}} & 92.59{\scriptsize$\pm$0.31}$^\dagger$ & 96.47{\scriptsize$\pm$0.07}$^\dagger$ & 98.25{\scriptsize$\pm$0.04} & 98.25{\scriptsize$\pm$0.04} \\
 & Reddit & 96.52{\scriptsize$\pm$0.13} & 97.39{\scriptsize$\pm$0.07} & 98.42{\scriptsize$\pm$0.02} & 96.98{\scriptsize$\pm$0.04} & 94.97{\scriptsize$\pm$0.05} & 98.00{\scriptsize$\pm$0.04} & 98.38{\scriptsize$\pm$0.07} & 98.69{\scriptsize$\pm$0.00}$^\dagger$ & \underline{98.73{\scriptsize$\pm$0.02}} & 91.04{\scriptsize$\pm$0.31}$^\dagger$ & \textbf{98.74{\scriptsize$\pm$0.01}}$^\dagger$ & 97.90{\scriptsize$\pm$0.01} & 97.90{\scriptsize$\pm$0.01} \\
 & MOOC & 83.16{\scriptsize$\pm$1.30} & \underline{91.24{\scriptsize$\pm$0.99}} & 81.86{\scriptsize$\pm$0.25} & 86.84{\scriptsize$\pm$0.17} & 82.77{\scriptsize$\pm$0.24} & 84.72{\scriptsize$\pm$1.31} & 90.27{\scriptsize$\pm$0.96} & 87.62{\scriptsize$\pm$0.51} & \textbf{95.55{\scriptsize$\pm$0.25}} & 76.92{\scriptsize$\pm$2.01}$^\dagger$ & 82.92{\scriptsize$\pm$0.11}$^\dagger$ & 90.55{\scriptsize$\pm$0.50} & 90.55{\scriptsize$\pm$0.50} \\
 & LastFM & 81.13{\scriptsize$\pm$3.39} & 82.61{\scriptsize$\pm$3.15} & 87.82{\scriptsize$\pm$0.12} & 76.99{\scriptsize$\pm$0.29} & 80.37{\scriptsize$\pm$0.18} & 91.60{\scriptsize$\pm$1.31} & 92.15{\scriptsize$\pm$0.68} & \underline{94.04}$^\dagger$ & \textbf{95.36{\scriptsize$\pm$0.06}} & 84.26{\scriptsize$\pm$0.56}$^\dagger$ & 93.54{\scriptsize$\pm$0.38}$^\dagger$ & 92.70{\scriptsize$\pm$1.95} & 92.70{\scriptsize$\pm$1.95} \\
 & Enron & 81.96{\scriptsize$\pm$1.34} & 78.83{\scriptsize$\pm$1.11} & 87.02{\scriptsize$\pm$0.50} & 64.63{\scriptsize$\pm$1.74} & 76.51{\scriptsize$\pm$0.71} & 87.95{\scriptsize$\pm$0.58} & 87.97{\scriptsize$\pm$0.61} & \textbf{90.26{\scriptsize$\pm$0.13}}$^\dagger$ & \underline{90.21{\scriptsize$\pm$0.49}} & 79.76{\scriptsize$\pm$1.02}$^\dagger$ & 89.11{\scriptsize$\pm$0.21}$^\dagger$ & 90.03{\scriptsize$\pm$0.17} & 90.03{\scriptsize$\pm$0.17} \\
 & Social Evo. & 93.70{\scriptsize$\pm$0.29} & 93.43{\scriptsize$\pm$0.59} & 84.73{\scriptsize$\pm$0.27} & 93.41{\scriptsize$\pm$0.19} & 94.09{\scriptsize$\pm$0.07} & 88.15{\scriptsize$\pm$6.36} & 94.78{\scriptsize$\pm$0.37} & \underline{95.29{\scriptsize$\pm$0.03}} & \textbf{95.47{\scriptsize$\pm$0.04}} & 94.79{\scriptsize$\pm$0.32}$^\dagger$ & 93.17{\scriptsize$\pm$0.08}$^\dagger$ & 94.91{\scriptsize$\pm$0.05} & 94.91{\scriptsize$\pm$0.05} \\
 & UCI & 78.80{\scriptsize$\pm$0.94} & 86.68{\scriptsize$\pm$2.29} & 90.40{\scriptsize$\pm$0.11} & 77.64{\scriptsize$\pm$0.38} & 89.30{\scriptsize$\pm$0.57} & 83.78{\scriptsize$\pm$0.37} & \underline{93.18{\scriptsize$\pm$0.17}} & 92.59{\scriptsize$\pm$0.02}$^\dagger$ & \textbf{94.40{\scriptsize$\pm$0.03}} & 82.22{\scriptsize$\pm$0.49}$^\dagger$ & 89.80{\scriptsize$\pm$0.60}$^\dagger$ & 92.83{\scriptsize$\pm$0.28} & 92.83{\scriptsize$\pm$0.28} \\
 & Flights & 95.21{\scriptsize$\pm$0.32} & 95.92{\scriptsize$\pm$0.43} & 96.86{\scriptsize$\pm$0.02} & 88.64{\scriptsize$\pm$0.35} & 82.27{\scriptsize$\pm$0.06} & 96.97{\scriptsize$\pm$0.20} & 97.69{\scriptsize$\pm$0.05} & \underline{97.80{\scriptsize$\pm$0.02}} & \textbf{98.05{\scriptsize$\pm$0.04}} & 76.72{\scriptsize$\pm$2.70}$^\dagger$ & 81.92{\scriptsize$\pm$0.06}$^\dagger$ & 97.48{\scriptsize$\pm$0.02} & 97.48{\scriptsize$\pm$0.02} \\
 & Can. Parl. & 53.81{\scriptsize$\pm$1.14} & 55.86{\scriptsize$\pm$0.75} & 58.83{\scriptsize$\pm$1.13} & 56.51{\scriptsize$\pm$0.75} & 58.32{\scriptsize$\pm$1.08} & \textbf{62.70{\scriptsize$\pm$2.91}} & 49.64{\scriptsize$\pm$0.69} & 54.27{\scriptsize$\pm$1.01}$^\dagger$ & 59.85{\scriptsize$\pm$1.57}$^\dagger$ & 55.55{\scriptsize$\pm$1.55}$^\dagger$ & 56.40{\scriptsize$\pm$0.63}$^\dagger$ & \underline{59.92{\scriptsize$\pm$2.60}} & \underline{59.92{\scriptsize$\pm$2.60}} \\
 & US Legis. & 58.12{\scriptsize$\pm$2.35} & \underline{62.38{\scriptsize$\pm$0.48}} & 51.49{\scriptsize$\pm$1.13} & 48.27{\scriptsize$\pm$3.50} & 47.20{\scriptsize$\pm$0.89} & \textbf{64.22{\scriptsize$\pm$0.65}} & 61.89{\scriptsize$\pm$1.52} & 53.21{\scriptsize$\pm$3.04} & 44.50{\scriptsize$\pm$0.99}$^\dagger$ & 49.33{\scriptsize$\pm$2.04}$^\dagger$ & 43.80{\scriptsize$\pm$1.90}$^\dagger$ & 44.69{\scriptsize$\pm$2.09} & 45.06{\scriptsize$\pm$1.55} \\
 & UN Trade & 62.28{\scriptsize$\pm$0.50} & 59.99{\scriptsize$\pm$3.50} & 67.05{\scriptsize$\pm$0.21} & 62.72{\scriptsize$\pm$0.12} & 63.48{\scriptsize$\pm$0.37} & 69.15{\scriptsize$\pm$2.33} & 64.05{\scriptsize$\pm$0.72} & 67.25{\scriptsize$\pm$1.05} & \textbf{73.18{\scriptsize$\pm$2.18}}$^\dagger$ & 55.02{\scriptsize$\pm$1.95}$^\dagger$ & 58.38{\scriptsize$\pm$0.21}$^\dagger$ & \underline{71.07{\scriptsize$\pm$1.61}} & \underline{71.07{\scriptsize$\pm$1.61}} \\
 & UN Vote & 58.13{\scriptsize$\pm$1.43} & 61.23{\scriptsize$\pm$2.71} & 48.34{\scriptsize$\pm$0.76} & 51.83{\scriptsize$\pm$1.35} & 50.04{\scriptsize$\pm$0.86} & \textbf{68.55{\scriptsize$\pm$0.90}} & 59.01{\scriptsize$\pm$1.88} & 56.73{\scriptsize$\pm$0.69} & \underline{63.59{\scriptsize$\pm$4.49}}$^\dagger$ & 56.34{\scriptsize$\pm$3.17}$^\dagger$ & 50.28{\scriptsize$\pm$2.11}$^\dagger$ & 61.21{\scriptsize$\pm$3.40} & 60.86{\scriptsize$\pm$3.14} \\
 & Contact & 95.37{\scriptsize$\pm$0.92} & 94.84{\scriptsize$\pm$0.75} & 89.07{\scriptsize$\pm$0.34} & 96.53{\scriptsize$\pm$0.10} & 92.83{\scriptsize$\pm$0.05} & 95.70{\scriptsize$\pm$0.06} & 97.66{\scriptsize$\pm$0.12} & \underline{98.30{\scriptsize$\pm$0.02}} & \textbf{98.51{\scriptsize$\pm$0.01}} & 92.65{\scriptsize$\pm$0.34}$^\dagger$ & 92.65{\scriptsize$\pm$0.05}$^\dagger$ & 97.89{\scriptsize$\pm$0.04} & 97.89{\scriptsize$\pm$0.04} \\
\cmidrule{2-15}
 & Avg. Rank & 9.38 & 7.62 & 8.15 & 9.69 & 9.77 & 6.23 & 5.62 & \underline{4.15} & \textbf{2.31} & 10.54 & 8.69 & 4.46 & 4.38 \\
\midrule
\multirow{14}{*}{\rotatebox{90}{hist}} & Wikipedia & 61.86{\scriptsize$\pm$0.53} & 75.75{\scriptsize$\pm$0.29} & 62.04{\scriptsize$\pm$0.65} & 78.38{\scriptsize$\pm$0.20} & \underline{82.87{\scriptsize$\pm$0.21}} & 40.95{\scriptsize$\pm$4.99} & 73.29{\scriptsize$\pm$2.33} & 63.45{\scriptsize$\pm$0.52}$^\dagger$ & 67.95{\scriptsize$\pm$2.77} & \textbf{84.00{\scriptsize$\pm$0.27}}$^\dagger$ & 82.11{\scriptsize$\pm$0.35}$^\dagger$ & 73.42{\scriptsize$\pm$0.93} & 81.21{\scriptsize$\pm$0.11} \\
 & Reddit & 61.69{\scriptsize$\pm$0.39} & 64.55{\scriptsize$\pm$0.50} & 64.94{\scriptsize$\pm$0.21} & 64.43{\scriptsize$\pm$0.27} & 64.27{\scriptsize$\pm$0.13} & 58.32{\scriptsize$\pm$0.63} & 64.02{\scriptsize$\pm$0.46} & 64.92{\scriptsize$\pm$0.64}$^\dagger$ & 62.37{\scriptsize$\pm$0.83} & 62.59{\scriptsize$\pm$0.51}$^\dagger$ & \textbf{80.14{\scriptsize$\pm$0.41}}$^\dagger$ & 63.89{\scriptsize$\pm$1.33} & \underline{72.57{\scriptsize$\pm$0.24}} \\
 & MOOC & 64.48{\scriptsize$\pm$1.64} & 77.69{\scriptsize$\pm$3.55} & 71.68{\scriptsize$\pm$0.94} & 74.08{\scriptsize$\pm$0.27} & 72.53{\scriptsize$\pm$0.84} & 67.99{\scriptsize$\pm$1.68} & 75.92{\scriptsize$\pm$2.48} & 80.77{\scriptsize$\pm$0.63} & \underline{84.46{\scriptsize$\pm$0.88}} & 61.26{\scriptsize$\pm$1.21}$^\dagger$ & 71.48{\scriptsize$\pm$0.51}$^\dagger$ & 79.57{\scriptsize$\pm$1.32} & \textbf{93.67{\scriptsize$\pm$0.47}} \\
 & LastFM & 68.44{\scriptsize$\pm$3.26} & 66.99{\scriptsize$\pm$5.62} & 67.69{\scriptsize$\pm$0.24} & 69.89{\scriptsize$\pm$0.28} & 70.07{\scriptsize$\pm$0.20} & 64.18{\scriptsize$\pm$0.65} & 75.02{\scriptsize$\pm$0.66} & 70.69$^\dagger$ & 77.10{\scriptsize$\pm$0.78} & 72.68{\scriptsize$\pm$0.23}$^\dagger$ & \underline{80.04{\scriptsize$\pm$3.87}}$^\dagger$ & 73.60{\scriptsize$\pm$1.24} & \textbf{88.38{\scriptsize$\pm$0.09}} \\
 & Enron & 65.32{\scriptsize$\pm$3.57} & 62.68{\scriptsize$\pm$1.09} & 62.25{\scriptsize$\pm$0.40} & 57.84{\scriptsize$\pm$2.18} & 68.20{\scriptsize$\pm$1.62} & 61.69{\scriptsize$\pm$0.68} & 78.90{\scriptsize$\pm$1.29} & 65.54{\scriptsize$\pm$0.51}$^\dagger$ & 74.50{\scriptsize$\pm$1.02} & 70.26{\scriptsize$\pm$1.25}$^\dagger$ & \underline{80.02{\scriptsize$\pm$0.80}}$^\dagger$ & 74.86{\scriptsize$\pm$2.02} & \textbf{82.89{\scriptsize$\pm$1.02}} \\
 & Social Evo. & 88.53{\scriptsize$\pm$0.55} & 92.10{\scriptsize$\pm$1.22} & 83.54{\scriptsize$\pm$0.24} & 91.87{\scriptsize$\pm$0.72} & 93.62{\scriptsize$\pm$0.35} & 84.89{\scriptsize$\pm$4.72} & 96.29{\scriptsize$\pm$0.56} & \underline{96.91{\scriptsize$\pm$0.09}} & 96.76{\scriptsize$\pm$0.14} & 88.17{\scriptsize$\pm$0.28}$^\dagger$ & 92.00{\scriptsize$\pm$0.30}$^\dagger$ & 96.74{\scriptsize$\pm$0.13} & \textbf{98.59{\scriptsize$\pm$0.03}} \\
 & UCI & 60.24{\scriptsize$\pm$1.94} & 62.69{\scriptsize$\pm$0.90} & 56.39{\scriptsize$\pm$0.10} & 62.32{\scriptsize$\pm$1.18} & 75.98{\scriptsize$\pm$0.84} & 42.59{\scriptsize$\pm$0.96} & \underline{81.34{\scriptsize$\pm$0.29}} & 64.93{\scriptsize$\pm$0.16}$^\dagger$ & 71.35{\scriptsize$\pm$0.84} & 76.29{\scriptsize$\pm$0.54}$^\dagger$ & 77.62{\scriptsize$\pm$0.32}$^\dagger$ & 76.49{\scriptsize$\pm$0.52} & \textbf{85.73{\scriptsize$\pm$0.20}} \\
 & Flights & 60.72{\scriptsize$\pm$1.29} & 59.66{\scriptsize$\pm$1.04} & 56.58{\scriptsize$\pm$0.44} & \underline{63.38{\scriptsize$\pm$0.26}} & 63.30{\scriptsize$\pm$0.19} & 48.39{\scriptsize$\pm$1.37} & 49.76{\scriptsize$\pm$0.89} & 56.05{\scriptsize$\pm$0.21} & 53.08{\scriptsize$\pm$0.87} & 58.16{\scriptsize$\pm$1.93}$^\dagger$ & 63.17{\scriptsize$\pm$0.14}$^\dagger$ & 56.99{\scriptsize$\pm$1.05} & \textbf{64.14{\scriptsize$\pm$0.19}} \\
 & Can. Parl. & 51.62{\scriptsize$\pm$1.00} & 55.64{\scriptsize$\pm$0.54} & \underline{60.11{\scriptsize$\pm$0.48}} & 58.30{\scriptsize$\pm$0.61} & 56.68{\scriptsize$\pm$1.20} & \textbf{61.72{\scriptsize$\pm$2.76}} & 48.93{\scriptsize$\pm$2.35} & 55.62{\scriptsize$\pm$0.26}$^\dagger$ & 59.29{\scriptsize$\pm$1.04}$^\dagger$ & 53.31{\scriptsize$\pm$2.49}$^\dagger$ & 57.31{\scriptsize$\pm$0.65}$^\dagger$ & 58.51{\scriptsize$\pm$3.11} & 58.10{\scriptsize$\pm$3.09} \\
 & US Legis. & 58.12{\scriptsize$\pm$2.94} & \underline{64.87{\scriptsize$\pm$1.65}} & 54.41{\scriptsize$\pm$1.31} & 49.99{\scriptsize$\pm$4.88} & 49.28{\scriptsize$\pm$0.86} & \textbf{66.95{\scriptsize$\pm$4.17}} & 62.82{\scriptsize$\pm$2.18} & 56.57{\scriptsize$\pm$3.22} & 45.36{\scriptsize$\pm$1.28}$^\dagger$ & 48.81{\scriptsize$\pm$2.33}$^\dagger$ & 42.10{\scriptsize$\pm$3.27}$^\dagger$ & 45.80{\scriptsize$\pm$3.30} & 45.80{\scriptsize$\pm$3.30} \\
 & UN Trade & 58.73{\scriptsize$\pm$1.19} & 55.61{\scriptsize$\pm$3.54} & 60.95{\scriptsize$\pm$0.80} & 59.74{\scriptsize$\pm$0.59} & 59.88{\scriptsize$\pm$1.17} & \underline{70.43{\scriptsize$\pm$2.07}} & 62.61{\scriptsize$\pm$1.70} & 58.46{\scriptsize$\pm$1.65} & 56.12{\scriptsize$\pm$1.13}$^\dagger$ & 51.28{\scriptsize$\pm$2.00}$^\dagger$ & 57.26{\scriptsize$\pm$0.87}$^\dagger$ & 62.72{\scriptsize$\pm$3.44} & \textbf{75.60{\scriptsize$\pm$1.77}} \\
 & UN Vote & 65.16{\scriptsize$\pm$1.28} & \underline{68.59{\scriptsize$\pm$3.11}} & 48.01{\scriptsize$\pm$1.77} & 51.73{\scriptsize$\pm$4.12} & 45.49{\scriptsize$\pm$0.42} & 67.69{\scriptsize$\pm$1.92} & \textbf{69.47{\scriptsize$\pm$3.01}} & 53.85{\scriptsize$\pm$2.02} & 52.73{\scriptsize$\pm$2.15}$^\dagger$ & 54.54{\scriptsize$\pm$4.13}$^\dagger$ & 46.02{\scriptsize$\pm$4.48}$^\dagger$ & 61.20{\scriptsize$\pm$2.50} & 60.95{\scriptsize$\pm$2.67} \\
 & Contact & 90.80{\scriptsize$\pm$1.18} & 88.84{\scriptsize$\pm$1.39} & 74.79{\scriptsize$\pm$0.37} & 93.76{\scriptsize$\pm$0.41} & 90.04{\scriptsize$\pm$0.29} & 84.74{\scriptsize$\pm$1.44} & 91.99{\scriptsize$\pm$0.28} & \underline{94.14{\scriptsize$\pm$0.26}} & 93.47{\scriptsize$\pm$0.43} & 80.59{\scriptsize$\pm$0.75}$^\dagger$ & 89.61{\scriptsize$\pm$0.38}$^\dagger$ & 92.83{\scriptsize$\pm$0.49} & \textbf{96.28{\scriptsize$\pm$0.17}} \\
\cmidrule{2-15}
 & Avg. Rank & 8.92 & 7.31 & 8.92 & 7.15 & 6.92 & 9.08 & 5.77 & 6.62 & 6.92 & 8.54 & 6.46 & \underline{5.58} & \textbf{2.81} \\
\midrule
\multirow{14}{*}{\rotatebox{90}{ind}} & Wikipedia & 61.87{\scriptsize$\pm$0.53} & 75.76{\scriptsize$\pm$0.29} & 62.02{\scriptsize$\pm$0.65} & 78.38{\scriptsize$\pm$0.20} & \underline{82.88{\scriptsize$\pm$0.21}} & 40.95{\scriptsize$\pm$4.99} & 73.30{\scriptsize$\pm$2.33} & 63.45{\scriptsize$\pm$0.52}$^\dagger$ & 67.96{\scriptsize$\pm$2.77} & \textbf{84.00{\scriptsize$\pm$0.27}}$^\dagger$ & 82.11{\scriptsize$\pm$0.34}$^\dagger$ & 73.42{\scriptsize$\pm$0.93} & 81.21{\scriptsize$\pm$0.11} \\
 & Reddit & 61.69{\scriptsize$\pm$0.39} & 64.55{\scriptsize$\pm$0.50} & 64.91{\scriptsize$\pm$0.21} & 64.39{\scriptsize$\pm$0.27} & 64.27{\scriptsize$\pm$0.13} & 58.31{\scriptsize$\pm$0.63} & 64.02{\scriptsize$\pm$0.46} & 64.91{\scriptsize$\pm$0.64}$^\dagger$ & 62.37{\scriptsize$\pm$0.83} & 62.59{\scriptsize$\pm$0.51}$^\dagger$ & \textbf{80.14{\scriptsize$\pm$0.41}}$^\dagger$ & 63.89{\scriptsize$\pm$1.33} & \underline{72.57{\scriptsize$\pm$0.24}} \\
 & MOOC & 64.48{\scriptsize$\pm$1.64} & 77.68{\scriptsize$\pm$3.55} & 71.69{\scriptsize$\pm$0.94} & 74.07{\scriptsize$\pm$0.27} & 72.52{\scriptsize$\pm$0.84} & 68.00{\scriptsize$\pm$1.68} & 75.93{\scriptsize$\pm$2.48} & 80.77{\scriptsize$\pm$0.63} & \underline{84.46{\scriptsize$\pm$0.88}} & 61.27{\scriptsize$\pm$1.21}$^\dagger$ & 71.48{\scriptsize$\pm$0.51}$^\dagger$ & 79.57{\scriptsize$\pm$1.32} & \textbf{93.67{\scriptsize$\pm$0.47}} \\
 & LastFM & 68.44{\scriptsize$\pm$3.26} & 66.99{\scriptsize$\pm$5.61} & 67.68{\scriptsize$\pm$0.24} & 69.89{\scriptsize$\pm$0.28} & 70.07{\scriptsize$\pm$0.20} & 64.18{\scriptsize$\pm$0.65} & 75.02{\scriptsize$\pm$0.66} & 70.69$^\dagger$ & 77.10{\scriptsize$\pm$0.78} & 72.69{\scriptsize$\pm$0.24}$^\dagger$ & \underline{80.04{\scriptsize$\pm$3.87}}$^\dagger$ & 73.60{\scriptsize$\pm$1.24} & \textbf{88.38{\scriptsize$\pm$0.09}} \\
 & Enron & 65.32{\scriptsize$\pm$3.57} & 62.68{\scriptsize$\pm$1.09} & 62.27{\scriptsize$\pm$0.40} & 57.83{\scriptsize$\pm$2.18} & 68.19{\scriptsize$\pm$1.63} & 61.69{\scriptsize$\pm$0.68} & 78.90{\scriptsize$\pm$1.29} & 65.54{\scriptsize$\pm$0.50}$^\dagger$ & 74.50{\scriptsize$\pm$1.02} & 70.25{\scriptsize$\pm$1.24}$^\dagger$ & \underline{80.02{\scriptsize$\pm$0.80}}$^\dagger$ & 74.86{\scriptsize$\pm$2.02} & \textbf{82.89{\scriptsize$\pm$1.02}} \\
 & Social Evo. & 88.53{\scriptsize$\pm$0.55} & 92.10{\scriptsize$\pm$1.22} & 83.54{\scriptsize$\pm$0.24} & 91.88{\scriptsize$\pm$0.72} & 93.62{\scriptsize$\pm$0.35} & 84.89{\scriptsize$\pm$4.72} & 96.29{\scriptsize$\pm$0.56} & \underline{96.91{\scriptsize$\pm$0.09}} & 96.76{\scriptsize$\pm$0.14} & 88.17{\scriptsize$\pm$0.28}$^\dagger$ & 92.00{\scriptsize$\pm$0.30}$^\dagger$ & 96.74{\scriptsize$\pm$0.13} & \textbf{98.59{\scriptsize$\pm$0.04}} \\
 & UCI & 60.27{\scriptsize$\pm$1.94} & 62.66{\scriptsize$\pm$0.91} & 56.39{\scriptsize$\pm$0.11} & 62.29{\scriptsize$\pm$1.17} & 75.97{\scriptsize$\pm$0.85} & 42.63{\scriptsize$\pm$0.97} & \underline{81.35{\scriptsize$\pm$0.29}} & 64.98{\scriptsize$\pm$0.16}$^\dagger$ & 71.37{\scriptsize$\pm$0.84} & 76.35{\scriptsize$\pm$0.54}$^\dagger$ & 77.65{\scriptsize$\pm$0.31}$^\dagger$ & 76.51{\scriptsize$\pm$0.52} & \textbf{85.71{\scriptsize$\pm$0.20}} \\
 & Flights & 60.72{\scriptsize$\pm$1.29} & 59.66{\scriptsize$\pm$1.05} & 56.58{\scriptsize$\pm$0.44} & \underline{63.40{\scriptsize$\pm$0.26}} & 63.32{\scriptsize$\pm$0.19} & 48.36{\scriptsize$\pm$1.37} & 49.74{\scriptsize$\pm$0.90} & 56.05{\scriptsize$\pm$0.22} & 53.05{\scriptsize$\pm$0.87} & 58.17{\scriptsize$\pm$1.93}$^\dagger$ & 63.17{\scriptsize$\pm$0.14}$^\dagger$ & 56.96{\scriptsize$\pm$1.05} & \textbf{64.12{\scriptsize$\pm$0.19}} \\
 & Can. Parl. & 51.61{\scriptsize$\pm$0.98} & 55.43{\scriptsize$\pm$0.42} & \underline{60.01{\scriptsize$\pm$0.47}} & 58.15{\scriptsize$\pm$0.62} & 56.63{\scriptsize$\pm$1.09} & \textbf{61.83{\scriptsize$\pm$2.92}} & 48.93{\scriptsize$\pm$2.24} & 55.93{\scriptsize$\pm$0.23}$^\dagger$ & 59.19{\scriptsize$\pm$1.00}$^\dagger$ & 53.24{\scriptsize$\pm$2.67}$^\dagger$ & 57.04{\scriptsize$\pm$0.62}$^\dagger$ & 58.53{\scriptsize$\pm$3.00} & 58.11{\scriptsize$\pm$2.98} \\
 & US Legis. & 58.12{\scriptsize$\pm$2.94} & \underline{64.87{\scriptsize$\pm$1.65}} & 54.41{\scriptsize$\pm$1.31} & 49.99{\scriptsize$\pm$4.88} & 49.28{\scriptsize$\pm$0.86} & \textbf{66.95{\scriptsize$\pm$4.17}} & 62.82{\scriptsize$\pm$2.18} & 56.57{\scriptsize$\pm$3.22} & 45.36{\scriptsize$\pm$1.28}$^\dagger$ & 48.81{\scriptsize$\pm$2.33}$^\dagger$ & 42.10{\scriptsize$\pm$3.27}$^\dagger$ & 45.80{\scriptsize$\pm$3.30} & 45.80{\scriptsize$\pm$3.30} \\
 & UN Trade & 58.71{\scriptsize$\pm$1.20} & 55.62{\scriptsize$\pm$3.59} & 60.88{\scriptsize$\pm$0.79} & 59.98{\scriptsize$\pm$0.59} & 59.71{\scriptsize$\pm$1.17} & \underline{70.34{\scriptsize$\pm$2.04}} & 62.54{\scriptsize$\pm$1.73} & 57.28{\scriptsize$\pm$3.06} & 56.20{\scriptsize$\pm$1.14}$^\dagger$ & 51.29{\scriptsize$\pm$2.00}$^\dagger$ & 57.30{\scriptsize$\pm$0.86}$^\dagger$ & 62.82{\scriptsize$\pm$3.45} & \textbf{75.66{\scriptsize$\pm$1.75}} \\
 & UN Vote & 65.29{\scriptsize$\pm$1.30} & \underline{68.58{\scriptsize$\pm$3.08}} & 48.04{\scriptsize$\pm$1.76} & 51.78{\scriptsize$\pm$4.14} & 45.57{\scriptsize$\pm$0.41} & 67.61{\scriptsize$\pm$1.90} & \textbf{69.44{\scriptsize$\pm$2.98}} & 53.87{\scriptsize$\pm$2.01} & 52.70{\scriptsize$\pm$2.16}$^\dagger$ & 54.54{\scriptsize$\pm$4.15}$^\dagger$ & 46.07{\scriptsize$\pm$4.47}$^\dagger$ & 61.18{\scriptsize$\pm$2.50} & 60.92{\scriptsize$\pm$2.69} \\
 & Contact & 90.80{\scriptsize$\pm$1.18} & 88.85{\scriptsize$\pm$1.39} & 74.79{\scriptsize$\pm$0.38} & 93.76{\scriptsize$\pm$0.40} & 90.04{\scriptsize$\pm$0.29} & 84.74{\scriptsize$\pm$1.44} & 92.00{\scriptsize$\pm$0.28} & \underline{94.14{\scriptsize$\pm$0.26}} & 93.47{\scriptsize$\pm$0.43} & 80.60{\scriptsize$\pm$0.75}$^\dagger$ & 89.61{\scriptsize$\pm$0.38}$^\dagger$ & 92.83{\scriptsize$\pm$0.49} & \textbf{96.28{\scriptsize$\pm$0.17}} \\
\cmidrule{2-15}
 & Avg. Rank & 8.92 & 7.38 & 9.00 & 7.08 & 7.00 & 9.08 & 5.77 & 6.54 & 6.92 & 8.54 & 6.38 & \underline{5.58} & \textbf{2.81} \\
\midrule
\multicolumn{2}{l}{Overall Avg. Rank} & 9.08 & 7.44 & 8.69 & 7.97 & 7.90 & 8.13 & 5.72 & 5.77 & 5.38 & 9.21 & 7.18 & \underline{5.21} & \textbf{3.33} \\
\bottomrule
\end{tabular}}
\end{table}

\subsection{Complete Transfer Results}

Table~\ref{tab:transfer_appendix} extends the transfer study in the main-text Table~\ref{tab:transfer} to additional protocols. TPNet, TGN, DyGFormer, and DSRD represent different relation-modeling architectures, while SNAM contains only 15--17 learnable parameters on these five datasets. Transfer directly reuses the SNAM parameters already learned on the same dataset, freezes both the activity model and the target backbone, and selects only the fusion coefficient $\beta$ on the validation split; neither component is retrained. The detailed procedure is provided in Appendix~\ref{app:transfer_details}.

Across the two settings in the supplementary table, all 40 comparisons spanning four backbones and five datasets yield AP improvements. In the transductive setting under inductive negative sampling, the average gain across datasets ranges from 5.86 to 12.28 points across backbones; in the inductive setting under historical negative sampling, it ranges from 6.42 to 17.27 points. In the latter setting, DyGFormer improves from 63.40\% to 88.51\% on Wikipedia. These results show that source-activity information complements the relation representations learned by different backbones and that its benefit does not depend on our relation encoder.

Table~\ref{tab:transfer_tpnet_streams} further covers the remaining eight datasets. Among the 24 reported comparisons under hard negative sampling, 19 improve and 4 remain unchanged; the only decrease is 2.42 points on UN Vote in the inductive setting. On UN Trade, the gain reaches 25.75 points in the transductive setting under inductive negative sampling. This extended study uses one seed per dataset. Overall, the results support source activity as a modeling dimension complementary to relation encoding. They also suggest that future work should distinguish between the two prediction questions---whether a source is active and which destination it selects---when designing training negative sampling, loss functions, and evaluation protocols.

\subsection{Roles of the Individual Modules}

\textbf{Multi-timescale historical representation.} Table~\ref{tab:a1_scales_full} shows that multiple timescales generally provide substantial gains over a single timescale, whereas further densifying the timescale grid yields smaller improvements. For example, in the transductive setting under historical negative sampling, the AP of the full model on UCI increases from 79.48\% at $K=1$ to 95.43\% at $K=8$, and reaches 96.12\% at $K=28$; on Reddit, it remains between 90.01\% and 90.16\% for $K=11$--$32$. Covering different temporal ranges is therefore more important than continually increasing the number of timescales. This trend does not hold for every individual result: for example, in the inductive setting under historical negative sampling, Wikipedia performs better at $K=1$. The default timescales are determined uniformly from the temporal range of the training data, as described in Appendix~\ref{app:timescale_configuration}.

\textbf{Within-block updates.} Table~\ref{tab:a2_event_clock} shows that, when within-block updates are retained, the full model is generally robust to changes in the number of blocks $C$; disabling these updates substantially weakens the benefit of the activity term. Even at $C=192$, AP under the transductive historical protocol decreases by 2.13--5.30 points across the five datasets. Together with Appendix~\ref{app:query_complexity}, these results show the importance of retaining within-block events that have occurred before the query time, so that the activity counts are updated exactly to the current time. This exactness applies to the exponential counts and does not imply that all higher-order representations are recomputed at every query.

\textbf{Inputs to the relation branch.} Table~\ref{tab:a4_trunk_channels} shows that removing the recent-interaction encoding or the random-projection features produces only small changes on several datasets, indicating that the performance does not depend on any single auxiliary branch. The count features and the joint contribution of the different inputs are more consequential. For example, removing all three input groups on UCI reduces transductive random AP from 95.81\% to 83.09\%. The results therefore support the complementary roles of multi-timescale historical representations and candidate-specific relation features, rather than suggesting that all inputs can be omitted. In this table, ``no projections'' also removes the recent-interaction encoding that depends on those projections and should therefore be interpreted as a joint ablation.

\textbf{Form of the activity term.} Table~\ref{tab:a3_appendix} and the main-text Table~\ref{tab:a3_transductive_inductive} show that recency and the MLP also yield substantial gains, supporting the value of activity information itself. Under the transductive historical protocol, the MLP exceeds SNAM by 0.17--0.35 points, but it is not consistently better in the other settings. This MLP uses the same historical counts as SNAM, takes $\log(1+n_u)$ as input, and adopts a two-layer $K\!\rightarrow\!32\!\rightarrow\!1$ architecture with GELU activations. During training only, it treats the source nodes of real events as positives and source nodes sampled randomly from the training set at the same timestamps as negatives, fits a binary cross-entropy objective, and is then frozen and used in place of the activity score.

We adopt SNAM because it requires only $K+1$ parameters, admits a clear event-intensity interpretation, and shares exponential historical statistics with the backbone. Its point-process likelihood has a closed-form integral and requires neither additional negative sampling nor numerical integration (Appendices~\ref{app:shared_historical_foundation} and~\ref{app:activity_likelihood}). In the main-text Table~\ref{tab:i1_decomposition}, the activity score alone outperforms TPNet and DyGFormer on four of the five datasets, further demonstrating its independent predictive value.

\textbf{Activity weight and evaluation protocol.} Figure~\ref{fig:beta_appendix} supplements the main-text Figure~\ref{fig:beta_mooc}. Under hard negative sampling, a moderate activity correction generally improves AP, after which performance saturates or declines as $\beta$ continues to increase. The decline under the random protocol arises from the joint ranking of examples with different source nodes and query times within each batch. For a fixed source node and query time, every candidate receives the same additive term $\beta\log\lambda_u(t)$, so their relative ranking is strictly unchanged; this does not imply that AP or AUC computed over mixed queries is unchanged. The theoretical explanation and evaluation definition are provided in Appendices~\ref{app:activity_motivation} and~\ref{app:negative_sampling_protocols}, respectively.

\subsection{Efficiency and Cost Trade-offs}
\label{app:efficiency_results}

\textbf{Efficiency across datasets.} Figure~\ref{fig:speed_appendix} shows that the main-text result in Figure~\ref{fig:speed_lastfm} is not the largest relative speedup. On Flights, the per-epoch training times of TPNet and DyGFormer are approximately $22\times$ and $100\times$ ours, respectively, while their full-test-set scoring times are approximately $6\times$ and $84\times$ ours; Reddit also exhibits a clear advantage. On smaller event streams such as UCI, fixed per-block overhead means that our method is not the fastest, and faster baselines also exist on Enron. The advantage of shared computation is more pronounced on the larger event streams, where alleviating large-scale processing costs matters most. All timings use the same RTX~5090 and report the median of three runs. The vertical axis reports AP under historical negative sampling, whereas test scoring is timed under random negative sampling. For the baselines, we time a subset of batches after warm-up and extrapolate to the full workload; for our model, we time the complete epoch and the complete test split.

\textbf{Cost of the number of timescales.} In Figure~\ref{fig:cost_s3}(a), relative to the main configuration with $K=14$--$16$, a single timescale takes approximately $0.65\times$ the average training time, whereas increasing the number of timescales to $27$--$32$ raises it to approximately $1.3\times$. Together with Table~\ref{tab:a1_scales_full}, this result shows that a moderate multi-timescale configuration already captures most of the benefit, without continually increasing the number of timescales.

\textbf{Within-block updates and memory.} In Figure~\ref{fig:cost_s3}(b), the ratio of per-epoch training time with versus without within-block updates is as high as approximately $3\times$ at $C=48$ and approximately $1.0$--$1.35\times$ at $C=192$. Figure~\ref{fig:cost_s3}(c) shows that finer blocks also reduce peak inference memory: LastFM falls to approximately 30\% of its value at $C=48$, while Reddit decreases from approximately 3.8~GB to 2.6~GB. Together with Table~\ref{tab:a2_event_clock}, these results make block granularity a practical time--memory control, but not a cost-free one: increasing $C$ also increases the frequency of block-start reconstruction, so per-epoch time is not guaranteed to decrease monotonically. The resulting cost structure is analyzed in Appendix~\ref{app:overall_complexity}.

\textbf{Event-stream length.} In Table~\ref{tab:s2_stream_length}, a tenfold longer LastFM stream costs 1.7$\times$ the epoch time (12.6\,s to 21.7\,s) and 5.5$\times$ the scoring time for 10$\times$ the test positives, so the cost per event and per query falls with length; with $C$ fixed at 192, the short prefix pays a high per-example overhead. On Flights the number of blocks grows with the stream (5 to 90) and the epoch time grows with it (0.5\,s to 13.8\,s), while the scoring time stays at about 1\,s per $10^5$ positives. This matches the cost structure of Appendix~\ref{app:overall_complexity}: an epoch costs a fixed block-start term plus a term linear in the number of events, and a query reads a state of fixed size plus the events of its own block, so its cost depends on the block size rather than on the length of the history. While two streams cannot establish an asymptotic rate, the measured costs never grow faster than the number of events over a tenfold range, which is what matters for the stream sizes of the benchmark.

\end{document}